\documentclass[preprint,12pt,authoryear,5p,twocolum]{elsarticle}

\usepackage{amssymb}
\usepackage{amsmath}

\usepackage{url}
\usepackage[dvipsnames,table]{xcolor}

\usepackage{hyperref}

\usepackage[disable]{todonotes}
\usepackage{caption}
\usepackage{subcaption}
\usepackage{amsmath}
\usepackage{cleveref}
\usepackage{array}
\usepackage{booktabs}
\usepackage{rotating}
\definecolor{newcolor}{rgb}{.8,.349,.1}

\begin{document}

\begin{frontmatter}



\title{Comparative Benchmarking of Failure Detection Methods in Medical Image Segmentation: Unveiling the Role of Confidence Aggregation} 


\author[1,2]{Maximilian Zenk\corref{cor1}}
\cortext[cor1]{Corresponding author. E-mail: m.zenk@dkfz-heidelberg.de}
\author[1]{David Zimmerer}
\author[1,4]{Fabian Isensee}
\author[3,4]{Jeremias Traub}
\author[1]{Tobias Norajitra}
\author[3,4]{Paul F. Jäger}
\author[1,5]{Klaus Maier-Hein}

\affiliation[1]{organization={German Cancer Research Center (DKFZ) Heidelberg, Division of Medical Image Computing, Germany}}
\affiliation[2]{organization={Medical Faculty Heidelberg, Heidelberg University, Heidelberg, Germany}}
\affiliation[3]{organization={German Cancer Research Center (DKFZ) Heidelberg, Interactive Machine Learning Group, Germany}}
\affiliation[4]{organization={Helmholtz Imaging, German Cancer Research Center (DKFZ), Heidelberg, Germany}}
\affiliation[5]{organization={Pattern Analysis and Learning Group, Department of Radiation Oncology, Heidelberg University Hospital, 69120 Heidelberg, Germany}}

\begin{abstract}
Semantic segmentation is an essential component of medical image analysis research, with recent deep learning algorithms offering out-of-the-box applicability across diverse datasets.
Despite these advancements, segmentation failures remain a significant concern for real-world clinical applications, necessitating reliable detection mechanisms.
This paper introduces a comprehensive benchmarking framework aimed at evaluating failure detection methodologies within medical image segmentation.
Through our analysis, we identify the strengths and limitations of current failure detection metrics, advocating for the risk-coverage analysis as a holistic evaluation approach.
Utilizing a collective dataset comprising five public 3D medical image collections, we assess the efficacy of various failure detection strategies under realistic test-time distribution shifts.
Our findings highlight the importance of pixel confidence aggregation and we observe superior performance of the pairwise Dice score \citep{roy_bayesian_2019} between ensemble predictions, positioning it as a simple and robust baseline for failure detection in medical image segmentation.
To promote ongoing research, we make the benchmarking framework available to the community.
\end{abstract}


\end{frontmatter}


\section{Introduction}

Segmentation is one of the most extensively studied tasks in medical image analysis and some algorithms based on deep learning perform well across various datasets \citep{isensee_nnu-net_2021}.
However, especially when applied to real-world environments or datasets from unseen scanners or institutions, decreased performance of deep learning models has been observed, for segmentation as well as other image analysis tasks \citep{albadawy_deep_2018,zech_variable_2018,badgeley_deep_2019,beede_human-centered_2020,campello_multi-centre_2021}.
Consequently, predictions may sometimes be inaccurate and cannot be trusted blindly.
While problematic segmentations can be identified through manual inspection, this becomes increasingly time-consuming with larger image dimensions and complex segmented structures, especially with (radiological) 3D images.
This issue becomes worse when segmentation is just one step in an automated analysis pipeline for large-scale datasets, making manual inspection impractical and trustworthy segmentations crucial.
Improving segmentation models and their robustness is one possible solution, but this work focuses on a complementary approach, which augments segmentation models with failure detection methods.
Failure detection, as defined and evaluated in this paper, aims to automatically identify segmentations that require exclusion or manual correction before proceeding to downstream tasks (e.g., volumetrics, radiotherapy planning, large-scale analyses). This involves providing a scalar confidence score per segmentation (image-level) to indicate the likelihood of segmentation failure.
While class- or pixel-level failure detection are interesting alternative options, we focus on image-level failure detection as it is often the practically most relevant task within the context of the above downstream analyses:
If a segmentation failure occurs on any level, a decision has to be made whether the whole prediction (image-level) is retained for further analyses or rejected.
In applications where partial predictions are useful, i.e. for a subset of pixels or classes, pixel-/class-level failure detection may be beneficial.
From a methodological perspective, it's worth noting that pixel- or class-level methods remain relevant for the image-level failure detection task, but they require appropriate aggregation functions that add complexity.

Failure detection for medical image segmentation is the motivation for several lines of research, each approaching the task in different ways:
Uncertainty estimation methods \citep{mehrtash_confidence_2020} typically aim to provide calibrated probabilities for each pixel prediction's correctness. Some studies propose to use these scores for failure detection tasks by aggregating them to a class or image level \citep{roy_bayesian_2019,jungo_analyzing_2020,ng_estimating_2023}. 
Methods for out-of-distribution (OOD) detection \citep{gonzalez_distance-based_2022,graham_transformer-based_2022} are designed instead to identify data samples that deviate from the training set distribution, which are suspected to result in segmentation failures.
As a third strand, segmentation quality regression methods\citep{valindria_reverse_2017,robinson_real-time_2018,li_towards_2022} attempt to directly predict segmentation metric values given an image without ground truth.
A comprehensive description of specific methods is provided in \cref{sec:related_work}.

Despite the practical relevance and the diversity of approaches, progress in segmentation failure detection is currently hindered by insufficient evaluation practices in existing works:
\begin{itemize}
    \item Different task definitions and evaluation metrics are used, although the approaches share the practical motivation of failure detection, making cross-work result comparison difficult. Often, proxy tasks like OOD detection, uncertainty calibration, or segmentation quality regression are evaluated instead of directly addressing failure detection \citep{mehrtash_confidence_2020,graham_transformer-based_2022,wang_efficient_2022,ouyang_improved_2022}. Moreover, the metrics used to measure failure detection lack standardization \citep{valindria_reverse_2017,wang_aleatoric_2019,jungo_analyzing_2020,wang_layer_2022,ng_estimating_2023}, with distinct characteristics and weaknesses rarely discussed.
    \item Evaluation typically focuses on a subset of relevant methods. Approaches to failure detection can be coarsely divided into pixel-level and image-level methods, but existing works usually concentrate on one of them, disregarding the potential for aggregating pixel-level uncertainty to image-level uncertainty. Some studies \citep{gonzalez_distance-based_2022,greenspan_segmentation_2023} compare both groups but limit aggregation methods to simple approaches like the mean uncertainty, which is biased toward object size \citep{jungo_analyzing_2020,kahl_values_2024}.
    \item Only a single dataset (anatomy) is used or no dataset shifts are considered \citep[for example]{jungo_analyzing_2020,ng_estimating_2023}.
    While focusing on a single segmentation task/dataset is valid for works targeting specific applications, this cannot answer questions about generalizability to other datasets and real-world applications, where distribution shifts are expected. Given segmentation methods like nnU-Net \citep{isensee_nnu-net_2021} that are easily trainable on various datasets, it is of high interest to determine which failure detection methods can complement them.
    \item Limited availability of publicly available implementations: Few papers release their code, often omitting baseline implementations (\ref{sec:code_availability}). This impedes reproducibility and leads to unreliable baseline performances.
\end{itemize}

To address these issues, we revisit the failure detection task definition and evaluation protocol to align them with the practical motivation of failure detection. This enables a comprehensive comparison of all relevant methods, leading us to construct a thorough benchmark for failure detection in medical image segmentation.

\begin{figure*}[t]
    \centering
    \includegraphics[width=0.8\textwidth]{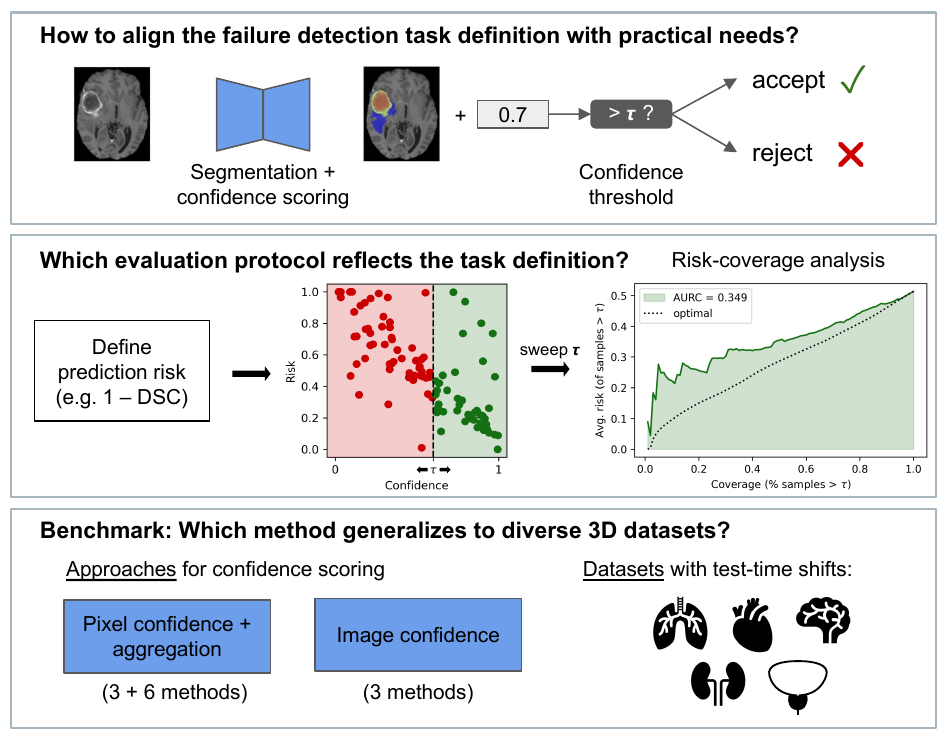}
    \caption{Overview of the research questions and contributions of this paper.
    Based on a formal definition of the image-level failure detection task, we formulate requirements for the evaluation protocol.
    Existing failure detection metrics are compared and the risk-coverage analysis is identified as a suitable evaluation protocol.
    We then propose a benchmarking framework for failure detection in medical image segmentation, which includes a diverse pool of 3D medical image datasets.
    A wide range of relevant methods are compared, including lines of research for image-level confidence and aggregated pixel confidence, which have been mostly studied in separation so far.
    }
    \label{fig:overview}
\end{figure*}

Our contributions (summarized in \cref{fig:overview}) include:
\begin{enumerate}
    \item Consolidating existing evaluation protocols by dissecting their pitfalls and suggesting a versatile and robust failure detection evaluation pipeline. This pipeline is grounded in a risk-coverage analysis adapted from the selective classification literature and mitigates the identified pitfalls.
    \item Introducing a benchmark that comprises multiple publicly available radiological 3D datasets to assess the generalization of failure detection methods beyond a single dataset setup. Our test datasets incorporate realistic distribution shifts, simulating potential sources of failure for a more comprehensive assessment.
    \item Under this proposed benchmark, we compare various methods that represent diverse approaches to failure detection, including image-level methods and pixel-level methods with subsequent aggregation. We find that the pairwise Dice \citep{roy_bayesian_2019} between ensemble predictions consistently performs best among all compared methods and recommend it as a strong baseline for future studies.
\end{enumerate}
The source code for all experiments, including dataset preparation, segmentation, failure detection method implementations, and evaluation scripts, is publicly available\footnote{Link will be added upon acceptance}.

\section{Realistic Evaluation of Failure Detection Methods}
\label{sec:metric_pitfalls}

\subsection{Task Definition}

We consider the task of detecting failures of a segmentation model $f: \mathcal{X} \rightarrow \mathcal{Y}$, which generates a segmentation $y = f(x)$ based on an image sample $x\in \mathbb{R}^{d_1\times d_2 \times d_3} $.
Complementing the segmentation model, a \emph{confidence scoring function} (CSF) provides a confidence score $\kappa$,
\begin{equation}
    g: \mathcal{X} \times \mathcal{Y} \times \mathcal{H} \rightarrow \mathbb{R} \, , \quad g(x, y, f) = \kappa\, ,
\end{equation}
where $\mathcal{H}$ is the space of segmentation models and higher scores imply higher confidence in the prediction.
Note that most concrete CSFs only use a subset of these inputs.
This definition could even be generalized to the case where $f$ and $g$ are integrated in the same model, but we do not consider this possibility here.
Failure detection consists of making a decision on whether to accept a prediction $y$ for downstream tasks.
This decision is based upon a threshold $\tau$, so $y$ is accepted if $g(x, y, f) \geq \tau$ and rejected otherwise.
The rejection threshold $\tau$ requires tuning before testing, which can for example be performed as in \citet{geifman_selective_2017}.

To evaluate failure detection methods, the risk associated with an accepted prediction has to be defined through a \emph{risk function} $R(y)$, where a higher risk indicates worse segmentation. 
Various risk functions are conceivable and the eventual choice depends on the specific application.
For instance, a domain expert could assign ordinal risk labels like ``high'', ``medium'', and ``low'' to images.
For this paper and the purpose of benchmarking failure detection methods, however, we assume the availability of ground truth masks $y_{gt}$ and employ segmentation metrics $m$ to construct the risk function. If higher values of $m$ correspond to better segmentation quality, for example,  $R(y, y_{gt}) = 1 - m(y, y_{gt})$.
Importantly, the assumption of having ground truth available for the test dataset is solely necessary for evaluating failure detection performance in this benchmark. However, none of the compared methods rely on this ground truth.

\subsection{Requirements on the Evaluation Protocol}

Based on the task definition above, we formulate requirements for the evaluation of methods and point out related pitfalls in current practice, to ensure progress is measured in a realistic failure detection setting.

\textbf{Requirement R1: Evaluate the failure detection task directly and allow comparison of all relevant solutions.} Similar to how \citet{jaeger_call_2022} argued for classification, a variety of proxy tasks for segmentation failure detection has been studied, each of them with their own metrics and restrictions, although failure detection is the commonly stated goal. To allow a comprehensive comparison and avoid excluding relevant methods, metrics are needed that summarize failure detection performance.
\\
\emph{Pitfalls in current practice:} A popular proxy task is OOD detection \citep{gonzalez_distance-based_2022,graham_transformer-based_2022}. While OOD detection certainly is useful, it is not identical to failure detection. For example, when applying a segmentation model to a new hospital, all samples are technically OOD, but only some of them might turn out to be failures. Vice versa, in-distribution samples can also result in failures.
Another commonly studied task is segmentation quality estimation, which phrases failure detection as a regression task of segmentation metric values \citep{kohlberger_evaluating_2012,valindria_reverse_2017,robinson_real-time_2018,liu_alarm_2019,li_towards_2022,greenspan_qcresunet_2023}. Although close to our task definition, it is slightly more restrictive, as confidence scores need to be on the same scale as the risk values.
This ``calibration'' can be desirable for some applications or to compute metrics like mean-absolute-error (MAE), but failure detection only requires a monotonous relationship between risk and confidence, and the evaluation should not be restricted to methods that output segmentation metric values directly.

\textbf{Requirement R2: Consider both segmentation performance and confidence ranking.}
Following \citet{jaeger_call_2022}, we argue that in practice the performance of the whole segmentation system matters, i.e. segmentation model and CSF. A desirable system has low remaining risk after rejection based on thresholding the confidence score, which can be achieved through (a) the CSF assigning lower confidence to samples with higher risk, i.e. better confidence ranking, or (b) avoiding high risks in the first place, i.e. better segmentation performance. These aspects cannot be easily disentangled, because the CSF might require architectural modifications that adversely impact segmentation performance, such as the introduction of dropout layers.
The evaluation metric should hence consider both aspects. 
Beyond the choice of metric, this requirement also implies that a fair comparison between failure detection methods uses the same segmentation model for different CSFs, if possible.
\\
\emph{Pitfalls in current practice:} Most related works use metrics that ignore the segmentation performance aspect and focus on confidence ranking \citep{robinson_real-time_2018,liu_alarm_2019,jungo_analyzing_2020,li_towards_2022}, such as AUROC of binary failure labels and Spearman correlation coefficients.
As a side-effect in the case of continuous risk definitions, exclusively considering confidence ranking while neglecting absolute risk differences can also lead to unexpected evaluation outcomes. Consider an example with four test samples and risk values of $\{0.1, 0.5, 0.7, 0.72\}$ and perfect confidence ranking, i.e. the first sample has the highest confidence and so on. Switching confidence ranks between the first two samples has the same effect on the Spearman correlation as switching the ranks of the last two, but the first switch is more problematic from a failure detection perspective. This issue is, for example, relevant in scenarios where there is a group of test samples with similar, low risks and a smaller number of samples with higher, more variable risks, which is likely to happen in a failure detection scenario where failures are rare.

\textbf{Requirement R3: Support flexible risk definitions.}
In contrast to image classification, there is no universal definition of what makes a segmentation faulty. The risk function depends ultimately on the specific application and can in particular be continuous. Therefore, a general evaluation protocol for failure detection, as required for our benchmark, should be flexible enough to support different choices.
\\
\emph{Pitfalls in current practice:} Several papers use a threshold on the Dice score to define failure \citep{devries_leveraging_2018,chen_cnn-based_2020,jungo_analyzing_2020,lin_novel_2022,ng_estimating_2023}, resulting in a binary risk function, which is reasonable if the specific application has a natural threshold. For many existing datasets, however, such a threshold cannot be determined easily, for instance when inter-annotator variability is unknown. In these cases, a continuous risk function like the Dice score can avoid information loss and discontinuity effects. Hence, a general-purpose evaluation metric should be applicable to both discrete and continuous risk functions, which is not given for some popular metrics like failure AUROC.

\textbf{Requirement R4: Consider realistic failure sources.}
CSFs should be primarily judged on how successful they are in detecting realistic failures. These can happen for numerous reasons, but distribution shifts in data from different scanners and populations are especially important, as they are likely to be encountered in real-world applications. The data used for evaluating CSFs should hence reflect these failure sources, ideally covering different types of dataset shifts.
\\
\emph{Pitfalls in current practice:} While earlier works focused on in-distribution testing \citep{devries_leveraging_2018,jungo_analyzing_2020,chen_cnn-based_2020}, there has been a development towards including test datasets from different centers or scanners in the evaluation \citep{mehrtash_confidence_2020,gonzalez_distance-based_2022,li_towards_2022,ng_estimating_2023}.
Some studies augment their test dataset with ``artificial'' predictions that are not produced by the actual segmentation model, for example by corrupting the segmentation masks or using auxiliary (weaker) segmentation models \citep{robinson_real-time_2018,li_towards_2022,greenspan_qcresunet_2023}.
While this practice has the benefit of testing the CSF on a wide range of segmentation qualities, we argue that it is not ideal for a benchmark on failure detection: Firstly, it contradicts R1, because only methods can be tested on the artificial test data that are independent of the segmentation model, excluding lines of work like ensemble uncertainty \citep{lakshminarayanan_simple_2017} or posthoc \citep{gonzalez_distance-based_2022} methods, although they are usually applicable in failure detection scenarios. Secondly, the additional samples might put more emphasis on failure cases that would never occur in a realistic setting, decreasing the influence of practically relevant cases in the evaluation.
It's important to note that realistic artificial images, unlike artificial predictions, can circumvent these drawbacks and meet requirement R4.

We present an evaluation protocol and dataset setup that meets the above requirements in \cref{sec:methods_eval,sec:methods_data}, respectively.

\section{Related Work}
\label{sec:related_work}

\subsection{Image-level Failure Detection Methods}
\label{sec:related_work_image_level}

Detecting when a model fails or identifying low-quality predictions has been used to motivate a wide range of approaches that directly output a confidence score on the image level. Two major lines of research are described below.

\subsubsection{Segmentation Quality Estimation}

One common approach in medical image analysis is segmentation quality control, which frames the task as a regression problem where the objective is to estimate one or more segmentation metrics' values for a given (image, segmentation) pair without access to the ground truth segmentation.

Initially, methods relied on hand-crafted features to train support vector machine regressors \citep{kohlberger_evaluating_2012}. However, recent advancements have seen the emergence of deep learning methods trained directly on raw images \citep{robinson_real-time_2018}. These deep learning approaches have been further enhanced by incorporating external uncertainty maps, if available, as additional input \citep{devries_leveraging_2018}, or by introducing secondary training objectives for pixel-level error detection \citep{greenspan_qcresunet_2023}.
An extension of this approach involves training a generative model to synthesize images from segmentations \citep{xia_synthesize_2020,li_towards_2022}.
Here, a Siamese network is trained in a second step to estimate both image-level and pixel-level segmentation quality based on the dissimilarity between the original and generated images.

Another notable method, known as reverse classification accuracy  \citep{valindria_reverse_2017}, utilizes each (image, segmentation) pair to train a new segmentation algorithm. This algorithm is then evaluated on a database of reference images with known ground truth. The estimated quality is determined by the best quality achieved within the reference set.

Finally, \citet{wang_deep_2020} train a Variational Autoencoder (VAE) \citep{kingma_auto-encoding_2013} on (image, ground truth segmentation) pairs and obtain a surrogate for the ground truth mask by iteratively optimizing the latent representation of the test sample. The desired segmentation metrics are then computed between the surrogate and the segmentation model prediction, which serve as an approximation to the true metrics.

\subsubsection{Distribution Shift Detection}

Another perspective in medical image segmentation focuses on detecting distribution shifts, which frequently lead to model failures. Such shifts occur when data samples in the testing set are not adequately represented in the training data. This line of research is closely related to OOD detection, a topic extensively explored in the broader machine learning community \citep{salehi_unified_2022}.

Various adaptations of OOD detection for medical image segmentation have been proposed. These methods typically attempt to fit a density estimation model to the training data distribution and utilize the likelihood of test samples as a confidence score. However, they vary in the choice of features used for density estimation and the probabilistic model.
For instance, \citet{liu_alarm_2019} train a VAE on ground truth segmentations and use the VAE loss directly as a confidence score. They also fit a linear model on the confidence scores and measured segmentation metrics on a validation set to convert this to a segmentation quality estimator.
Others utilize the latent representations of the training set generated by the segmentation network and fit a multivariate Gaussian, quantifying uncertainty as the Mahalanobis distance of a test sample \citep{gonzalez_distance-based_2022}.
Another variation involves utilizing a VQ-GAN \citep{esser_taming_2021} for feature extraction and a transformer network for density estimation \citep{graham_transformer-based_2022}.

\subsection{Pixel-level Uncertainty Methods}

Numerous methods for predictive uncertainty estimation in image classification can be adapted to segmentation, resulting in pixel-level uncertainty maps.

Bayesian methods attempt to model the posterior over model parameters instead of providing a point estimate based on the training data. One popular method with Bayesian interpretation is MC-Dropout \citep{gal_dropout_2016}, which utilizes dropout at test-time to approximate the posterior. This method has been adapted to segmentation by \citet{kendall_bayesian_2016} and frequently applied to tasks in the medical domain \citep{roy_bayesian_2019,jungo_analyzing_2020,hoebel_exploration_2020,kwon_uncertainty_2020,mehrtash_confidence_2020,nair_exploring_2020}.

Ensembles offer another approach to obtain uncertainty estimates \citep{lakshminarayanan_simple_2017}. For deep neural networks, it is common to train multiple models with different random initializations and average their predictions at test time. Pixel uncertainties can be computed from the softmax distribution through predictive entropy or mutual information, for example. Ensemble uncertainty has also found application in medical image segmentation \citep{mehrtash_confidence_2020,hoebel_i_2022}.

Ambiguity in the ground truth and inter-annotator differences pose significant challenges in medical image analysis. Some methods address this by producing a distribution of predictions that cover the different possible segmentations of an image \citep{kohl_probabilistic_2018,monteiro_stochastic_2020}. These methods focus on aleatoric uncertainty estimation, which concerns inherent data variability, rather than epistemic (model) uncertainty.

Another pixel-level uncertainty approach is based on test-time augmentation \citep{wang_aleatoric_2019}. This technique involves applying various transformations to the input data during inference to obtain multiple predictions. Merging these predictions, after applying corresponding inverse transforms, can improve segmentation performance and also estimate pixel-level uncertainty. Recent evidence indicates that this method primarily estimates epistemic uncertainty \citep{kahl_values_2024}.

\subsection{Aggregation of Pixel-level Uncertainties}
\label{sec:related_work_aggregation}

Aggregating pixel-level uncertainties to obtain image-level uncertainty is crucial for failure detection but has not been extensively studied to date.

Some studies have proposed aggregation methods that rely on a segmentation network outputting multiple predictions. Based on this predictive distribution, pairwise segmentation metrics like Dice score can be computed and used as a confidence score \citep{roy_bayesian_2019}, akin to regression methods, but also other quantities like the coefficient of variation of segmentation volumes.

A comparison of different aggregation methods on a brain tumor segmentation dataset, including simple mean and learned aggregation models based on hand-crafted or radiomics features, revealed improved failure detection performance with more sophisticated aggregation approaches \citep{jungo_analyzing_2020}.

Additionally, some methods try to circumvent the bias of mean confidence towards images with large foreground and aggregate by considering only image patches with the lowest confidence or by averaging only confidences above a tuned threshold \citep{kahl_values_2024}.

\subsection{Benchmarking Efforts for Segmentation Failure Detection}

While there are previous efforts to benchmark uncertainty methods for medical image segmentation, this paper stands out as the first to conduct a comprehensive benchmark on failure detection, as it compares a wide range of methods from the previous sections on multiple radiological datasets, which contain realistic distribution shifts at test time.

\citet{jungo_analyzing_2020} are limited to analyzing a single brain tumor dataset without distribution shifts in the test data.
\citet{mehrtash_confidence_2020} focused primarily on the calibration of pixel-level uncertainty and did not consider image-level methods. 
The benchmark in \citet{ng_estimating_2023} includes distribution shifts but concentrates on heart segmentation. Furthermore, it does not compare image-level methods.
Some recent works propose benchmarks with a similar motivation as failure detection \citep{adams_benchmarking_2023,vasiliuk_redesigning_2023}. However, the evaluation by \citet{vasiliuk_redesigning_2023} is closely related to OOD detection and still requires OOD labels, which precludes failure detection evaluation on in-distribution data. \citet{adams_benchmarking_2023} focus on two organ segmentation tasks, of which one has distribution shifts in the test set. They exclude image-level methods from the comparison and do not consider confidence aggregation methods in depth.
Lastly, a comprehensive study on uncertainty estimation for segmentation was performed by \citet{kahl_values_2024}, but failure detection was only investigated as one of several downstream tasks of uncertainty estimation. Therefore, they utilized only a single medical dataset and did not consider image-level methods.

In the field of international competitions (also known as challenges), there are two works related to segmentation failure detection. 
The BraTS challenge 2020 \citep{bakas_identifying_2019,mehta_uncertainty_2020} focused on comparing uncertainty methods for brain tumor segmentation. However, it did not consider dataset shifts or explicitly address failure detection tasks; instead, it concentrated on pixel-level uncertainty estimation.
The Shifts challenge 2022 \citep{malinin_shifts_2022} featured a task on white matter lesions segmentation in the context of Multiple Sclerosis, incorporating distribution shifts in the test data. Although this competition evaluated the robustness of methods to shifts and considered failure detection, it utilized only a single medical dataset, and no meta-analysis of submitted methods has been published to date.
\section{Materials and Methods}

\subsection{Evaluation}
\label{sec:methods_eval}

To benchmark failure detection methods, we need concise failure detection metrics that fulfill the requirements R1--R3 from \cref{sec:metric_pitfalls}. 
We compare common metric candidates in \cref{tab:metric_pitfalls} and choose to perform a risk-coverage analysis as the main evaluation, with the area under the risk-coverage curve (AURC) as a scalar failure detection performance metric, as it fulfills all requirements.
The risk-coverage analysis was originally proposed by \citet{el-yaniv_foundations_2010} and AURC was suggested as a comprehensive failure detection metric for image classification by \citet{jaeger_call_2022}.
\begin{table*}[t]
    \caption{Comparison of metric candidates for segmentation failure detection. Among those, AURC is the only metric that captures segmentation performance and confidence ranking, which we find necessary for the comprehensive evaluation of a failure detection system. A detailed discussion of the requirements (R1--R3) associated with each column is in \cref{sec:metric_pitfalls}. f-AUROC uses binary failure labels. MAE: mean absolute error. PC: Pearson correlation. SC: Spearman correlation. 
    }

    \centering
    \begin{tabular}{l >{\centering\arraybackslash}p{3.3cm} >{\centering\arraybackslash}p{3cm} >{\centering\arraybackslash}p{3cm} >{\centering\arraybackslash}p{3cm}}
    \toprule
    \textbf{Metric} & \textbf{Required confidence scale (R1)} & \textbf{Considers confidence ranking (R2)} & \textbf{Considers segmentation performance (R2)} & \textbf{Compatible with binary/continuous risk (R3)} \\
    \midrule
    f-AUROC        & \textcolor{Green}{ordinal/real-valued} & \textcolor{Green}{yes} & \textcolor{BrickRed}{no} &  \textcolor{Green}{yes}/\textcolor{BrickRed}{no} \\
    MAE            & \textcolor{BrickRed}{same as risk} & \textcolor{BrickRed}{no} & \textcolor{BrickRed}{no} &  \textcolor{BrickRed}{no}/\textcolor{Green}{yes} \\
    PC  & \textcolor{Orange}{real-valued} & \textcolor{Orange}{implicitly} & \textcolor{BrickRed}{no} &  \textcolor{Green}{yes}/\textcolor{Green}{yes} \\
    SC & \textcolor{Green}{ordinal/real-valued} & \textcolor{Green}{yes} & \textcolor{BrickRed}{no} &  \textcolor{Green}{yes}/\textcolor{Green}{yes} \\
    AURC           & \textcolor{Green}{ordinal/real-valued} & \textcolor{Green}{yes} & \textcolor{Green}{yes} &  \textcolor{Green}{yes}/\textcolor{Green}{yes} \\
    \bottomrule
\end{tabular}


    \label{tab:metric_pitfalls}
\end{table*}

In our experiments, we define the risk function using the Dice score (DSC) as $R(y, y_{gt}) = 1 - \mathrm{DSC}(y, y_{gt})$, taking the mean over all classes in cases of multi-class datasets. The normalized surface distance \citep[NSD]{nikolov_deep_2020} is used in an auxiliary analysis on the choice of the risk function.
Eventually, experiment results consist of risks $r_i$ and confidence scores $\kappa_i$ for each test case $x_i$ ($i = 1, \ldots, N$).
We first determine a risk-coverage curve by sweeping a confidence threshold $\tau$ and measuring the selective risk $R_s$ and the coverage $C$. $R_s$ is defined as the average risk of all samples that are above the threshold:
\begin{equation}
R_s (\tau) = \frac{\sum_{i=1}^N r_i \cdot \mathbb{I}(\kappa_i \geq \tau) } {\sum_{i=1}^N \mathbb{I}(\kappa_i \geq \tau)} \, ,
\end{equation}
where $\mathbb{I}$ denotes the indicator function. The coverage is defined as the fraction of samples that are above the threshold:
\begin{equation}
 C(\tau) = \sum_i \mathbb{I}(\kappa_i \geq \tau) / N    
\end{equation}
An example risk-coverage curve for artificial experiment results is depicted in \cref{fig:overview}. The AURC can then be obtained as the area under the curve.
We adapt the publicly available implementation for AURC from \citet{jaeger_call_2022} to segmentation tasks.

AURC can be interpreted as the average selective risk across confidence thresholds, i.e. the average 1 - DSC in the standard setting of our experiments, so lower values are better.
The AURC of random CSFs is identical to the average overall risk $\sum_i r_i / N$, while the optimal CSF sorts the risk values in descending order.
As discussed in \cref{sec:metric_pitfalls} (R2), care must be taken to compare CSFs based on the same underlying segmentation model. In our study there are three different models: predictions are obtained either from single networks with one forward pass or with multiple forward passes using test-time dropout or from ensembles. We highlight any cross-model comparisons in the text.

Alternative metrics from the literature include correlation coefficients between confidence scores and segmentation metric values \citep{liu_alarm_2019} such as Spearman's rank correlation coefficient (SC) and Pearson's correlation coefficient (PC). Further popular metrics are failure-AUROC and MAE. In \cref{tab:metric_pitfalls}, we show that none of these metrics fulfill all requirements from \cref{sec:metric_pitfalls}. To emphasize that our findings do not strongly depend on the choice of AURC as a metric, however, we report SC in \ref{sec:additional_results}, as it captures confidence ranking and fulfills all requirements except R2.

\subsection{Datasets}
\label{sec:methods_data}

Requirement R4 from \cref{sec:metric_pitfalls} refers to the datasets used for evaluating failure detection. Specifically, we considered radiological datasets with segmentations of 3D structures from computed tomography (CT) or magnetic resonance imaging (MRI), as these are the most common modalities in the medical image analysis community.    
We strived to include different distribution shifts, but also included one dataset for which enough failures occur in distribution.
Importantly, only publicly available datasets were considered, to guarantee reproducibility and usefulness to the community.
Based on these criteria and previous work \citep{gonzalez_distance-based_2022,wang_layer_2022}, we selected the datasets summarized in Table \ref{tab:dataset_summary}.

Each dataset was split into training and testing cases on a per-patient basis; the test cases were not used for training or tuning. From the training set, 20\% of cases are set aside for validation in a 5-fold cross-validation manner, so that there are five different training-validation folds. Examples from the test split of each dataset are shown in \ref{sec:dataset_details} and a detailed description of all datasets follows.

\begin{table*}[!t]
    \caption{Summary of datasets used in this study. The \#Testing column contains case numbers for each subset of the test set separated by a comma, starting with the in-distribution test split and followed by the shifted ``domains''. The number of classes includes one count for background.}
    \label{tab:dataset_summary}
    \centering
    \resizebox{\textwidth}{!}{
    \begin{tabular}{lccccl}
    \toprule
    \textbf{Dataset} & \textbf{\#Classes} & \textbf{\#Training} & \textbf{\#Testing} & \textbf{Modality} & \textbf{Shift in test set} \\
    \midrule
    Brain tumor (2D) & 2 & 939 & $313 \times 5$ & MRI & Artificial corruptions \\
    Brain tumor    & 4 & 235 & 50, 50 & MRI & Higher prevalence of low-grade tumors \\
    Heart          & 4 & 190 & 60, 190, 100, 100 & MRI & Unseen scanner vendors \\
    Prostate       & 2 & 26 & 6, 30, 19, 13, 12, 12 & MRI & Unseen institutions \\
    Covid          & 2 & 160 & 39, 50, 20 & CT & Unseen institutions \\
    Kidney tumor   & 4 & 367 & 122 & CT & - \\
    \bottomrule
\end{tabular}
    }
\end{table*}

\subsubsection*{Brain Tumor (2D)}
Despite being 2D, this simplified version of the FeTS 2022 dataset \citep{pati_federated_2021,bakas_advancing_2017,menze_multimodal_2015} is included in the benchmark, as it allows for quick experimentation.
For pre-processing, we cropped the original images around the brain, selected only the axial slice with the largest tumor extent for each case, and resized that slice to $64 \times 64$ pixels. Each case consists of four MR sequences (T1, T1-Gd, T2, T2-FLAIR).
All publicly available cases were split randomly into a training and a test set.
To introduce shifts in the test set, we applied artificial corruptions using the torchIO library \citep{perez-garcia_torchio_2021}. For each test case, four randomized image transformations were applied, producing four additional corrupted versions per test case: affine transforms, bias field, spike and ghosting artifacts.
Due to the low image resolution, only the whole tumor region was used as a label for this dataset.

\subsubsection*{Brain Tumor}
The BraTS 2019 dataset \citep{bakas_advancing_2017,menze_multimodal_2015,bakas_identifying_2019} contains information about the tumor grade (glioblastoma, HGG, or lower grade glioma, LGG) for each training case.
To simulate a population shift with more LGG cases during testing, we split all publicly available cases into a training and a test set, such that there are 167 HGG and 26 LGG cases in the training set and 50 cases for each grade in the testing set. Note that LGG cases are often harder to segment \citep{bakas_identifying_2019}. 
Each case consists of four MR sequences (T1, T1-Gd, T2, T2-FLAIR). The labels for this dataset are nested tumor regions: whole tumor, tumor core, and enhancing tumor.
A similar dataset is used in \citet{hoebel_i_2022}, but we include a small number of LGG cases during training to make the setup more realistic.

\subsubsection*{Heart}
We use the M\&Ms dataset \citep{campello_multi-centre_2021}, which provides short-axis MRI data from four scanner vendors. For the training set, we use only samples from vendor B, while the testing set contains 30 patients (60 images) of vendor B and data from the other three vendors. Note that each patient comprises two images at the end-diastolic and end-systolic phase, respectively.
The labels are the left ventricle, the right ventricle, as well as the left ventricular myocardium.
This dataset has been used in \citet{wang_layer_2022} as well, but we split it differently, as \citet{full_studying_2020} showed that generalization from vendor B to A is most difficult.

\subsubsection*{Prostate}
For training and in-distribution testing, we use the prostate dataset from the medical segmentation decathlon \citep{simpson_large_2019,antonelli_medical_2022}.
More testing data is added from \citet{liu_shape-aware_2020}, who prepared data from \citet{bloch_nci-isbi_2015,litjens_promise12_2023,lemaitre_computer-aided_2015}.
We use the T2-weighted MR sequence and evaluate only the whole prostate label.
This setup is similar to \citet{gonzalez_distance-based_2022}, with the difference that we use all institutions from \citet{liu_shape-aware_2020} except RUNMC, as it originates from the same institution as the training data.

\subsubsection*{Covid}
We use the COVID-19 CT segmentation challenge dataset \citep{roth_rapid_2022,an_ct_2020,clark_cancer_2013}, from which 39 cases are set aside for testing and the remaining patients used for training. Additional test cases come from datasets collected at other institutions \citep{morozov_mosmeddata_2020,jun_covid-19_2020}.
There is a single foreground label for lesions related to COVID-19.
This dataset follows the setup from \citet{gonzalez_distance-based_2022}.

\subsubsection*{Kidney Tumor}
The publicly available cases from the KiTS23 dataset \citep{heller_state_2021,heller_kits21_2023} are split randomly into a training and test set.
Although no explicit shift is present in the test set, we observed that there are enough difficult cases in it that can be used for failure detection evaluation.
The same three nested regions as in the challenge are used as labels: Kidney + cyst + tumor, cyst + tumor, and tumor.

\subsection{Segmentation algorithm}

For the 2D brain tumor dataset, we used a U-Net architecture  \citep{ronneberger_u-net:_2015} with 5 layers, residual units, and dropout with rate 0.3, using the implementation of the MONAI library \citep{cardoso_monai_2022}. The Dice loss was optimized with the AdamW algorithm \citep{loshchilov_decoupled_2019}. The network was trained on whole images using only mirroring augmentations, to make the network susceptible to test-time corruptions. Inference was performed on whole images, too.

All 3D datasets were pre-processed using the nnU-Net framework \cite{isensee_nnu-net_2021} and 3D U-Nets were trained with a combination of Dice and cross-entropy loss (binary cross-entropy for region-based datasets) and the Momentum-SGD optimizer using a polynomial learning rate decay.
The U-Net architecture was adapted dynamically to the dataset using the MONAI library.
One dropout layer with a dropout rate of 0.5 was used as the final layer of five U-Net levels centered around the bottleneck following \citet{kendall_bayesian_2016}, to allow for test-time dropout (\cref{sec:fd_methods}) while only mildly regularizing the network.
The networks were trained on image patches using nnU-Net's data loader and augmentations until convergence.
Sliding-window inference was employed for the 3D datasets with an overlap of 0.5 and combined using Hann window weighting \citep{perez-garcia_torchio_2021}.

For each dataset, we trained U-Nets with five different random seeds for each of the five cross-validation folds, resulting in 25 models per dataset.
Detailed hyperparameters are given in \ref{sec:appendix_hparams}.

\subsection{Failure detection methods}
\label{sec:fd_methods}

When selecting methods to compare in our benchmark, we considered the following criteria:
popularity in the literature, diversity of approaches, and simplicity or availability of a reference implementation. All methods were re-implemented using PyTorch \citep{paszke_pytorch_2019}. 
Each method is described in the following and detailed hyperparameters are given in \ref{sec:appendix_hparams}.

\subsubsection{Pixel-level Confidence Methods}

We consider three methods, which produce predictions and a confidence map, which serve as an intermediate step for image-level failure detection.

\textbf{Single network} This simple baseline utilizes the softmax predictions of a single U-Net and computes pixel-wise predictive entropy (PE) as the confidence map.

\textbf{MC-Dropout}, introduced by \citet{gal_dropout_2016}, activates dropout layers at test-time to produce 10 softmax maps (5 for the kidney tumor dataset due to resource constraints). These are averaged to obtain a prediction and the pixel-wise PE is computed as a confidence map.

\textbf{Deep ensemble}, as proposed by \citet{lakshminarayanan_simple_2017}, trains five networks with different experimental seeds. Similar to \citet{mehrtash_confidence_2020}, adversarial training is omitted for simplicity. The prediction is obtained from the mean softmax map and pixel confidence scores are derived by computing pixel-wise PE.

\subsubsection{Pixel Confidence Aggregation Methods}

Pixel confidence aggregation methods receive the discrete prediction and the confidence map from the pixel confidence methods and output a scalar confidence score for the entire image.
The aggregation methods below can be subdivided into two groups: those that do not require training (mean, non-boundary, patch-based) and those that do (simple, radiomics).

\textbf{Mean} aggregation method simply computes the mean of the confidence scores across all pixels.

\textbf{Non-boundary-weighted} aggregation is motivated by the observation that uncertainty at object boundaries is often high due to minor annotation ambiguities. As the boundary length is correlated with object size, mean aggregation may result in higher uncertainty simply because an object is large \citep{jungo_analyzing_2020,kahl_values_2024}. To mitigate this, boundary regions are masked out during confidence aggregation by computing a segmentation boundary mask (with a width of 4 pixels) and averaging the confidence only within the non-boundary region.

\textbf{Patch-based} aggregation, proposed by \citet{kahl_values_2024}, offers a different solution and computes patch-wise confidence scores in a sliding-window manner using a predefined patch size of $10^D$ for $D$-dimensional images. These are aggregated into an image-level score by considering the minimum patch confidence.

\textbf{Regression forest (RF) on radiomics features} follows  \citet{jungo_analyzing_2020} to fit a regression random forest (RF) to DSC scores based on radiomics features, which are computed from the pixel confidence map. The region of interest for feature extraction is defined by thresholding this confidence map, as in \citet{jungo_analyzing_2020}.
One challenge with this approach is generating a suitable training set. For simplicity, we utilize cross-validation predictions and confidence maps, obtained using the same inference procedure as during testing.
We used the scikit-learn implementation of the regression forest \citep{pedregosa_scikit-learn_2011} and pyradiomics for feature extraction \citep{van_griethuysen_computational_2017}.

\textbf{Regression forest (RF) on simple features} is a similar but simpler variant we introduce, which replaces radiomics features with five hand-crafted heuristic features. These features are: mean confidence in the predicted (1) foreground, (2) background, and (3) boundary region, along with (4) foreground size (fraction) and (5) the number of connected components in the prediction.

\subsubsection{Pairwise DSC Estimator}

\citet{roy_bayesian_2019} proposed several class-level uncertainty methods, including the \textbf{pairwise DSC} between prediction samples. It requires a set of $M$ discrete segmentation masks, which are produced by MC-Dropout or ensembles in our experiments.
Dice scores are then computed between each pair of masks and all $M \cdot (M - 1) / 2$ values (since DSC is symmetric) are averaged to obtain a scalar confidence score. For datasets with multiple classes, we use mean DSC in the pairwise computation.

\subsubsection{Quality regression}
\label{sec:methods_qr}

The idea behind this image-level method is to train a deep neural network to predict segmentation quality (i.e. metrics) directly for a given image and segmentation mask \citet{robinson_real-time_2018}.
We call this approach \textbf{quality regression} in the remainder, as this is essentially a regression task.

For the regression network, we use the same U-Net encoder architecture as for segmentation training and add a regression head, which consists of global 3D average pooling and a linear layer. The L2 loss also used in \citet{robinson_real-time_2018} is optimized with the AdamW optimizer \citep{loshchilov_decoupled_2019} and cosine annealing learning rate decay.

The training data for this method consists of (image, segmentation) pairs and we use the DSC scores for each class as the target values. It is important to include a wide distribution of target values (DSC scores for individual classes) in the dataset. For simplicity, we use the validation set predictions (CV) of the single segmentation networks for training the regression network. More sophisticated target balancing methods are possible but beyond the scope of this paper (see \cref{sec:discussion}).

Images and masks were first cropped to a dataset-specific bounding box around the foreground (which is given by the prediction during testing) and, if necessary to train on an 11GB GPU, resized by a factor of 0.5.
Data augmentations consisting of randomized Zoom, Gaussian noise, intensity scaling and mirroring were applied to the images and masks. Additionally, we applied affine transformations with probability $1/3$ to segmentation masks to simulate slight misalignments and cover a wider distribution of target quality scores.

\subsubsection{Mahalanobis-distance OOD Detector}

\citet{gonzalez_distance-based_2022} proposed this image-level method, which is trained by extracting feature maps from the pre-trained segmentation model for the training data and fitting a multi-variate Gaussian distribution on them.
This yields a probabilistic model that can be used at test time to estimate how far a test data point is from the training distribution, by computing the \textbf{Mahalanobis} distance, which is used as the confidence score.

We make use of the public method implementation from \citet{gonzalez_distance-based_2022}. As in the original publication, we use the U-Net bottleneck layer features and reduce their dimension by adaptive average pooling before fitting the Gaussian to the flattened features. For inference, we also use the original patch-based approach.

\subsubsection{Variational Autoencoder}

Following \citet{liu_alarm_2019}, 
we first train \textbf{VAE} \citep{kingma_auto-encoding_2013} on the ground truth segmentation masks of the training set to learn a model of ``correct'' segmentations.
During testing, we feed the predicted mask into the VAE and use its loss as a scalar confidence score, which is a lower bound of the likelihood of the predicted mask and hence a measure for the ``normality'' of a mask.

We use a symmetric encoder-decoder architecture with five pooling operations and a latent dimension of 256. 
The binary cross-entropy is used in the reconstruction loss term of segmentations and the KL divergence term is weighted by a factor of $\beta=0.001$. The Adam optimizer \citep{kingma_adam_2017} is used with a learning rate of 0.0001.
A similar data loading pipeline as for the quality regression method was used, the main difference being that the cropped region was smaller, excluding a larger part of the background region. Further, no misalignment augmentations were used.

\section{Results}

In the following sections, we first report the segmentation performances without failure detection in \cref{sec:results_segmentation}.
Then, we describe the main benchmark results, starting with a comparison of pixel confidence aggregation methods (\cref{sec:results_aggregation}) and extending the scope towards pixel- and image-level methods (\cref{sec:results_overall_comparison}).
In \cref{sec:results_risk_ablation}, we study the effect of alternative failure risk definitions.
Finally, we perform a qualitative analysis of the pairwise DSC method, to understand its strengths and weaknesses (\cref{sec:results_qualitative}).

\subsection{Segmentation results}
\label{sec:results_segmentation}

\begin{figure*}[!t]
    \centering
    \includegraphics[width=\textwidth]{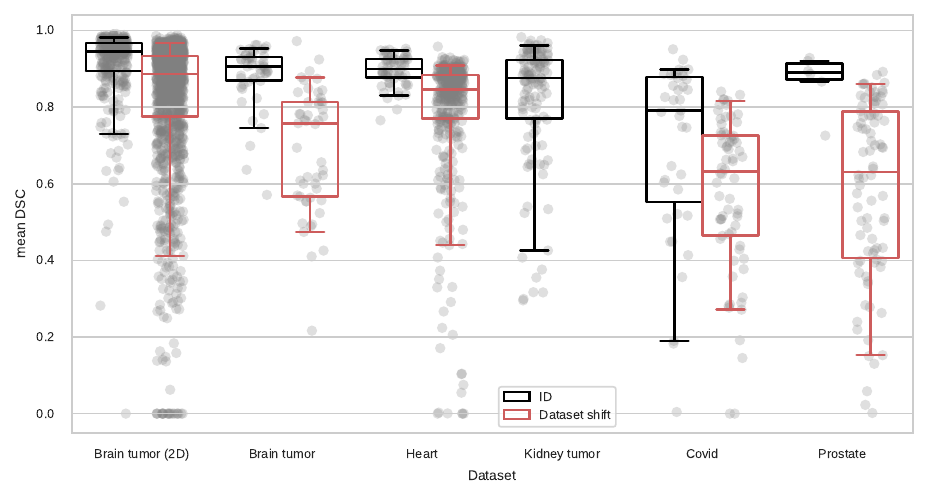}
    \caption{Segmentation performance of a single U-Net on the test sets.
    Boxes show the median and IQR, while whiskers extend to the 5th and 95th percentiles, respectively.
    Each dataset contains samples drawn from the same distribution as the training set (in-distribution, ID) and samples drawn from a different data distribution (dataset shift) with the same structures to be segmented.
    Usually, the performance on the in-distribution samples is higher than on the samples with distribution shift, but especially for the Kidney tumor (which lacks dataset shifts) and Covid datasets, there are also several in-distribution failure cases.
    }
    \label{fig:seg_performance}
\end{figure*}

To motivate the need for failure detection, we first show the segmentation performance of a single baseline network on all test sets  (\cref{fig:seg_performance}), distinguishing between test cases that are randomly sampled from the training distribution (in-distribution, ID) and test cases from parts of the dataset with distribution shift.
On in-distribution data, segmentation quality is usually high, with median DSC scores around 0.9. However, especially for the Covid and kidney tumor datasets, in-distribution test cases can still be hard and lead to failures.
As expected, dataset shifts lead on average to a clear drop in performance. Some test cases with distribution shifts are still segmented well by segmentation models, though, which stresses the point that the distinction between ID and OOD is often not sufficient for failure detection.

\subsection{Comparison of pixel confidence aggregation methods}
\label{sec:results_aggregation}

\begin{figure*}[!t]
    \centering
    \includegraphics[width=\textwidth]{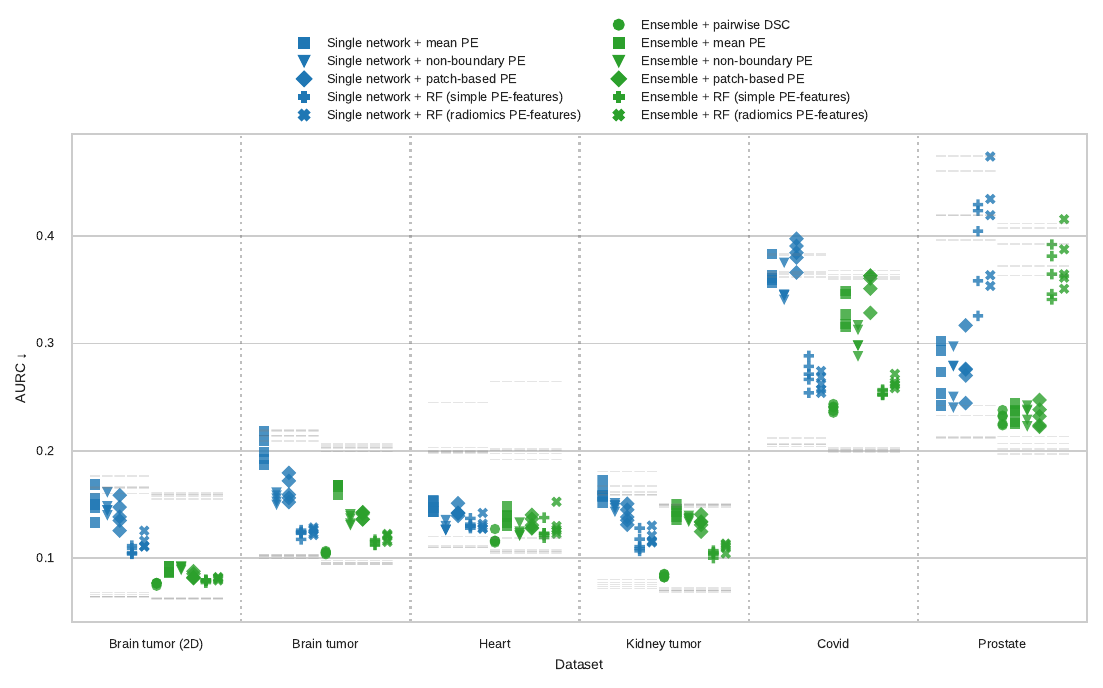}
    \caption{
        Comparison of aggregation methods in terms of AURC scores for all datasets (lower is better).  The experiments are named as ``prediction model + confidence method'' and each of them was repeated using 5 folds.
        Colored markers denote AURC values achieved by the methods, while gray marks above/below them are AURC values for random/optimal confidence rankings (which differ between the models trained on different folds; see \cref{sec:methods_eval}).
        Pairwise DSC scores consistently best, but does not apply to single network outputs.
        Aggregation methods based on regression forests (RF) also show performance gains compared to the mean PE baseline, but fail catastrophically on the prostate dataset, possibly due to the small training set size. PE: predictive entropy. RF: regression forest.
    }
    \label{fig:overview_aurc_aggregation}
\end{figure*}

Given that aggregation of pixel confidence methods has not been studied in a multi-dataset setup thus far (\cref{sec:related_work_aggregation}), we first examine the effect of various aggregation methods \cref{fig:overview_aurc_aggregation} for different pixel predictors.
By evaluating all methods in our multi-dataset setup, we can investigate how general they are and in which conditions they break.
We use two pixel-confidence methods, single network and deep ensemble, because they cover the whole range of performances observed in the benchmark. MC-Dropout is excluded to avoid cluttering \cref{fig:overview_aurc_aggregation}.

The ensemble + pairwise DSC method performs best across the board. While other aggregation methods operate on the pixel-wise predictive entropy (PE) of ensemble softmax distributions, pairwise DSC uses the set of discrete ensemble predictions. As it also requires a distribution of pixel-level predictions, we included it in this comparison of aggregation methods, but note that it is not applicable to single network outputs.

Comparing aggregation methods for the same pixel CSF (different marker styles in the same color), our multi-dataset setup reveals considerable variation between datasets. 
The mean confidence baseline is usually among the worst-scoring methods, but we could not identify a single aggregation method that performs best across all tested datasets and predictors (ensemble + pairwise DSC works best but does not apply to a single network prediction).
Among the untrained aggregation methods, the non-boundary and patch-based aggregations can provide minor improvements over mean, and it depends on the dataset which works best.
The trained aggregation methods boost failure detection performance on most datasets but also display a severe performance drop on the prostate dataset. This could be due to the small training/validation set size of 21/5 samples. Notably, the simple features consistently perform on par or better than the radiomics features, which suggests that features of the prediction mask can help detect failures and that the complexity of radiomics features is not required.

Comparing the same aggregation method for different pixel predictors (same marker style for different colors), the AURC scores of the deep ensemble are in most cases better than those of a single network. This effect is less pronounced for trained aggregation methods. As described in requirement R2 from \cref{sec:metric_pitfalls}), a difference in absolute AURC values can be due to better segmentation performance or confidence ranking. While the ensemble has better DSC scores on average, auxiliary metrics such as SC (\cref{fig:overview_agg_spearman}) indicate that on all but the Kidney tumor dataset, also the confidence ranking of the ensemble is better and, hence, their confidence maps may be more informative of failures.

\subsection{Comparison of all failure detection methods}
\label{sec:results_overall_comparison}

\begin{figure*}[!t]
    \centering
    \includegraphics[width=\textwidth]{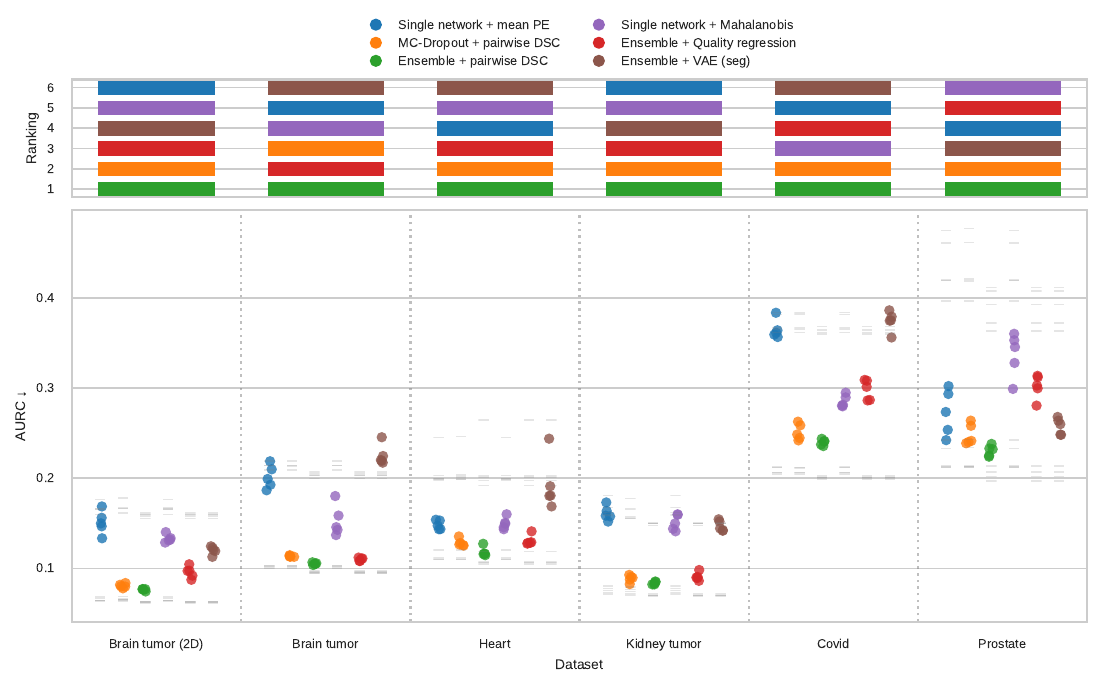}
    \caption{
        Rankings by average AURC (top, lower ranks are better) and the underlying AURC scores (bottom; lower is better) for all datasets and methods. The experiments are named as ``prediction model + confidence method'' and each of them was repeated using 5 folds. In the lower diagram, colored dots denote AURC values achieved by the methods, while gray marks above/below them are AURC values for random/optimal confidence rankings (which differ between the models trained on different folds; see \cref{sec:methods_eval}).
        Most of the aggregation methods from \cref{fig:overview_aurc_aggregation} were excluded for clarity, as they perform worse than pairwise DSC.
        Ensemble + pairwise DSC is the best method overall, often achieving close to optimal AURC scores. The ranking on the prostate dataset is an outlier, which could be due to the small training set size. PE: predictive entropy.
        \todo[inline]{@reviewers: 
        (1) Should I replace mean PE with a better aggregation method? In the previous section, random forest (simple features) performed usually better, but not always. The mean PE might also just be a good lower baseline here.
        (2) What do you think about the small grey dots? Keep them or not?}
    }
    \label{fig:overview_aurc}
\end{figure*}

\Cref{sec:results_aggregation} focused on pixel-confidence aggregation methods, but failure detection can also be solved directly by image-level methods.
To perform a comprehensive comparison that fulfills all requirements from \cref{sec:metric_pitfalls}, we include also image-level methods from \cref{sec:fd_methods} and present the results in \cref{fig:overview_aurc}.
From the aggregation methods, only pairwise DSC is included in this overview, as it performed best in \cref{sec:results_aggregation}.
The single network + mean PE baseline is additionally included as it is a naive but commonly used baseline.

As a high-level overview of the results, we ranked each failure detection method on each dataset (top of \cref{fig:overview_aurc}), after averaging AURCs across training folds. This shows that ensemble + pairwise DSC is consistently the best method overall. MC-Dropout + pairwise DSC is a close second.
These two methods are ranked consistently across datasets, which could indicate they are not specialized to a particular dataset, but generally applicable, an insight that could only be gained through our multi-dataset benchmark setup.
In contrast, the other methods exhibit more variability in their relative rank. Quality regression is often in the third place but degrades on the Covid and Prostate datasets. For the latter, this result could be due to the small training set size.
The ranking of the remaining three methods (mean PE, Mahalanobis, VAE) depends strongly on the dataset.

The AURC scores in the lower part of \cref{fig:overview_aurc} provide a more nuanced view of the results.
In general, we can see that the choice of failure detection method can help to avoid significant risks: The AURC difference between the naive baseline (Single network + mean PE) and the best method (Ensemble + pairwise Dice) can exceed 0.1 (Kidney tumor and Covid), which can be interpreted as an expected increase of mean DSC scores by 0.1 if rejecting low-confidence samples (averaged over all confidence thresholds; can be improved by threshold calibration).
While the mean PE only shows a clear benefit over random AURC for the Heart and Prostate datasets, the ensemble + pairwise DSC is always close to the optimal AURC and shows considerably less variance between the training folds.
Pairwise DSC also yields low AURCs when used in conjunction with MC-Dropout. Even though the latter produces different predictions than an ensemble, our evaluation setup using AURC allows us to fairly compare the overall failure detection performance of both methods, because it considers both segmentation performance and confidence ranking (requirement R2).
We report results for alternative failure detection metrics in \ref{sec:additional_results}, which are overall consistent with the AURC results.

As an interesting side-observation, we note that the recently proposed Mahalanobis method is not among the best methods in our benchmark. In contrast, when evaluated on the OOD detection task, it attains the first place on most datasets (\cref{fig:ood-auc}), which is plausible given that the method was designed for out-of-distribution detection. This confirms that the failure detection task is different from OOD detection and suggests that distinct methods may be required for each task.

\subsection{Alternative Risk Definition}
\label{sec:results_risk_ablation}

In the previous sections, pairwise DSC turned out to be best at detecting failures. As the risk function was based on DSC scores in our experiments, the question arises whether this finding changes for different risk functions.
Therefore, we investigate how the method ranking changes when we use an alternative risk function based on the mean normalized surface distance (NSD) \citep{nikolov_deep_2020}, which is a popular segmentation metric that focuses on the distance of the predicted segmentation boundary to the reference instead of the overlap between the two masks.

\begin{figure*}[!t]
\centering
     \begin{subfigure}[b]{0.48\textwidth}
         \centering
         \includegraphics[width=\textwidth]{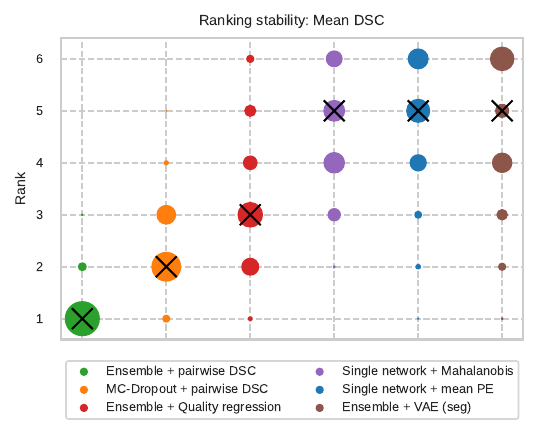}
     \end{subfigure}
     \hfill
     \begin{subfigure}[b]{0.48\textwidth}
         \centering
         \includegraphics[width=\textwidth]{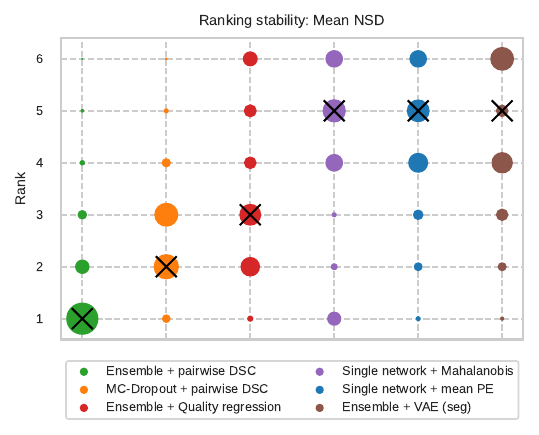}
    \end{subfigure}
    \caption{Impact of the choice of segmentation metric as a risk function on the ranking stability, comparing mean DSC (left) and NSD (right). Bootstrapping ($N=500$) was used to obtain a distribution of ranks for the results of each fold and the ranking distributions of all folds were accumulated. All ranks across datasets are combined in this figure, where the circle area is proportional to the rank count and the black x-markers indicate median ranks, which were also used to sort the methods.
    Overall, the ranking distributions are similar for mean DSC and NSD. The variance in the ranking distributions largely originates from combining the rankings across datasets, so for each dataset individually the ranking is more stable (see for example the Covid dataset in \cref{fig:compare_surface_dice_520}).
    }
    \label{fig:compare_surface_dice}
\end{figure*}

We find in \cref{fig:compare_surface_dice} that the ranking for the NSD risk is slightly less stable but overall very similar to the ranking with DSC risk, which indicates that our results are robust to a moderate change in the risk function.
The most significant change in rankings is visible in the rank-1 placements of the Mahalanobis method when using mean NSD as the risk function (\cref{fig:compare_surface_dice}, right). Interestingly, these outliers occur only for the Covid dataset (\cref{fig:compare_surface_dice_520}). As the Mahalanobis method excels particularly in OOD detection for this dataset, it is possible that the ``OOD-ness'' is more informative of the NSD score than of the DSC score in this special case.

\subsection{Qualitative analysis of ensemble predictions}
\label{sec:results_qualitative}

\begin{figure*}
    \begin{subfigure}{\textwidth}
        \centering
        \includegraphics[width=\textwidth]{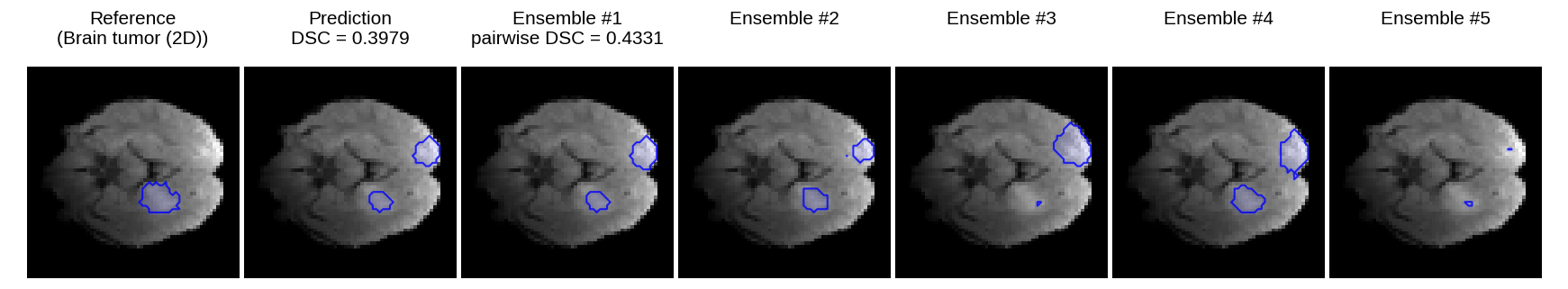}
    \end{subfigure}
    \hfill
    \begin{subfigure}{\textwidth}
        \centering
        \includegraphics[width=\textwidth]{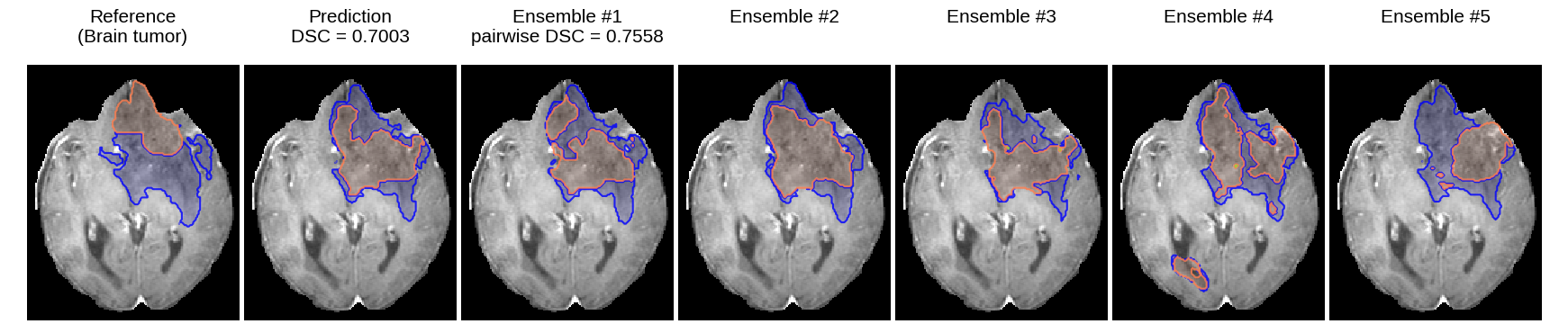}
    \end{subfigure}
    \hfill
    \begin{subfigure}{\textwidth}
        \centering
        \includegraphics[width=\textwidth]{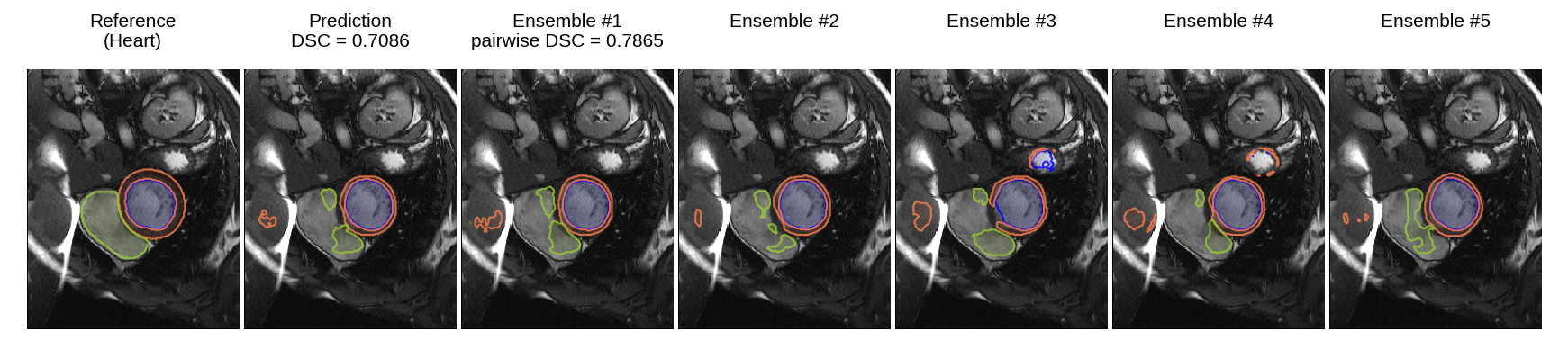}
    \end{subfigure}
    \hfill
    \begin{subfigure}{\textwidth}
        \centering
        \includegraphics[width=\textwidth]{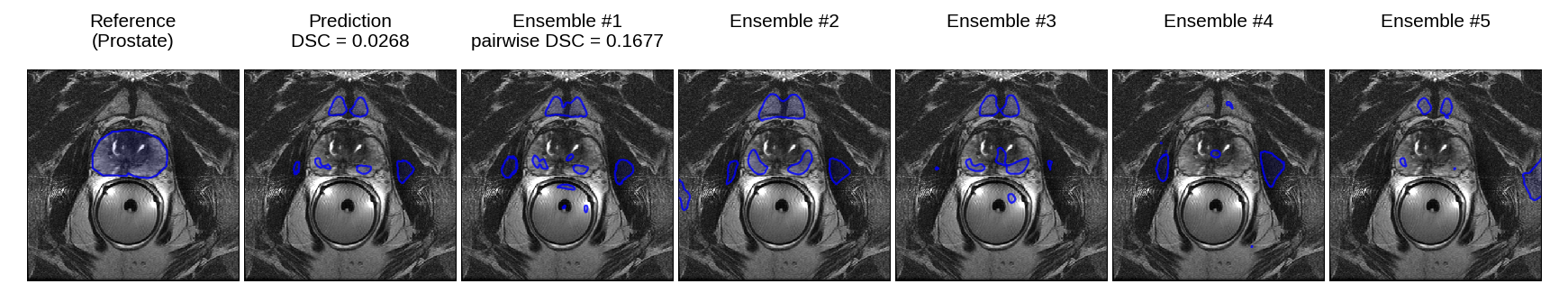}
    \end{subfigure}
    \hfill
    \begin{subfigure}{\textwidth}
        \centering
        \includegraphics[width=\textwidth]{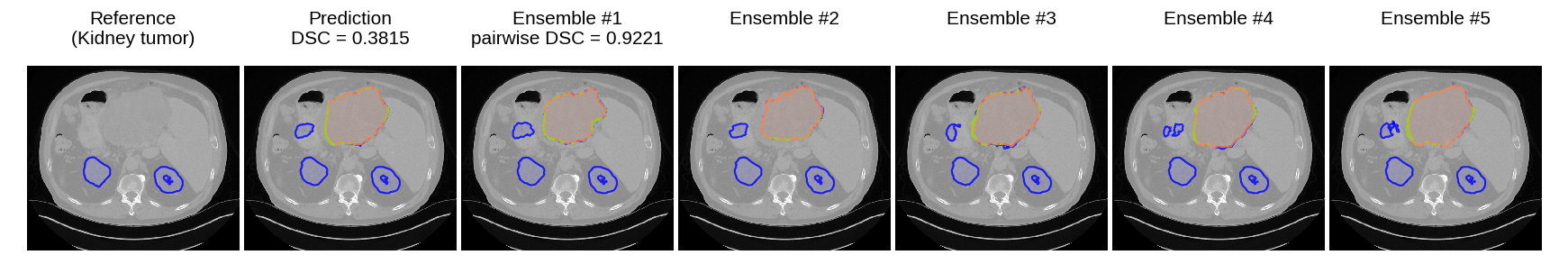}
    \end{subfigure}
    \hfill
    \begin{subfigure}{\textwidth}
        \centering
        \includegraphics[width=\textwidth]{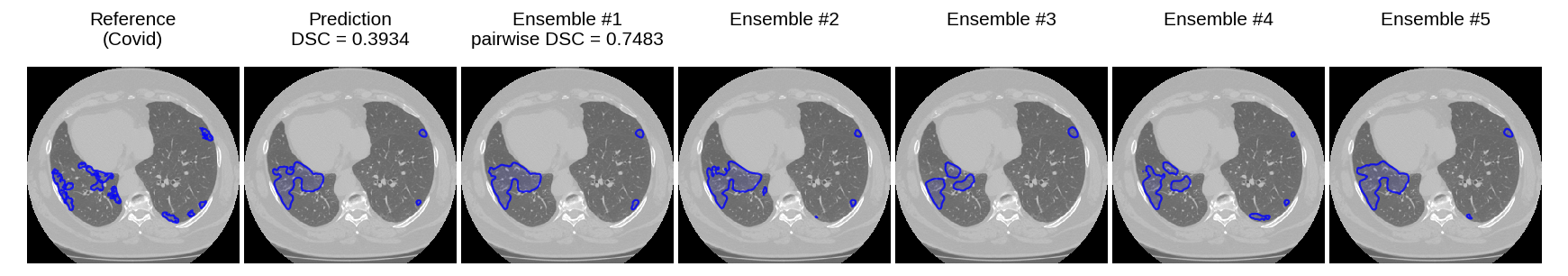}
    \end{subfigure}
    \caption{
        Qualitative analysis of ensemble predictions on all datasets.
        For each dataset, an interesting failure case and the corresponding ensemble predictions are displayed.
        True mean DSC is reported alongside the pairwise DSC scores.
        The ensemble predictions often disagree about test cases for which segmentation errors occur, which leads to low pairwise Dice and can be considered a detected failure (rows 1--4). However, there are also cases where the ensemble is confident about a faulty segment, which could result in a silent failure (last two rows).
    }
    \label{fig:qualitative_ensemble}
\end{figure*}

To get an intuition for the strengths and weaknesses of the ensemble + pairwise DSC method, we show examples of failure cases from all datasets along with their risks, confidence scores, and ensemble predictions in \cref{fig:qualitative_ensemble}.
Note that we deliberately picked faulty predictions here to illustrate the model behavior in failure cases.

For the Brain tumor (2D) dataset, all ensemble members seem to rely heavily on intensity, which makes them fail in the presence of bias field artifacts (global intensity gradients) present in this example. However, disagreements in the ensemble predictions can in cases such as the one shown here be used to detect these failures.

In the more complex, high-resolution 3D Brain tumor dataset, errors often stem from ambiguity in the non-enhancing tumor core region (orange), which is reflected in the ensemble predictions.

Errors in the Heart dataset occur even though the anatomical structures are clearly visible in the images from different scanners. The ensemble predictions often show large variations in regions where segmentation errors occur. In this example, some of the wrong segmentations (orange region on the left) could likely be avoided by excluding all but the largest connected component for each class. Post-processing steps like this can applied before pairwise DSC is computed, highlighting its flexibility.

The Prostate dataset features shifts in acquisition techniques, most notably the presence of endorectal coils, which were absent in the training data. These lead to a large variation in the appearance of images and catastrophic segmentation failures. Ensemble predictions are often rather unstable so that failures can still be detected.

On the Kidney tumor dataset, the ensemble makes some obvious errors where masses beyond the kidneys and an additional kidney are segmented. In contrast to the previous examples, however, the ensemble is overconfident in this case, as the ensemble members agree on the wrongly segmented region. This results in such failures not being detected, i.e. silent.

The example from the Covid dataset shows that there might be a slight annotation shift between the training cases and parts of the test set (MosMed subset), which leads to relatively low DSC scores although most lesions are detected. The ensemble disagreements appear to capture some of the annotation ambiguities, but the absolute value of pairwise DSC differs substantially from the true DSC. Note that this is not necessarily problematic for failure detection, because the latter requires only confidence \emph{ranking}, i.e. samples with lower true DSC have lower pairwise DSC. Still, \emph{calibration} of the pairwise DSC scores is a desirable secondary goal.

\section{Discussion}
\label{sec:discussion}

We revisited the task definition of segmentation failure detection and formulated fundamental requirements for its evaluation (\cref{sec:metric_pitfalls}).
Based on these, we recommend performing risk-coverage analyses and adding AURC to the standard evaluation metrics for segmentation failure detection, as it is a simple and interpretable metric that enables holistic evaluation of a failure detection system.
Under this evaluation protocol, we designed a benchmark with a diverse set of datasets, including realistic distribution shifts at test time, and compared a variety of failure detection methods.
We found that an ensemble with pairwise DSC confidence score performed consistently best across datasets. It is hence a strong baseline for failure detection and should be reported in future work, which is not standard so far. MC-Dropout + pairwise DSC is a good alternative if training resources are limited.
Incorporating multiple datasets in the benchmark turned out to be important, as all other methods showed considerable performance differences between datasets. For example, while quality regression networks performed well on three out of five 3D datasets, the gap towards the best methods was large for the remaining two datasets (Covid and Prostate).

Below we put our methods and results in perspective to the existing literature.
Apart from the metrics examined in \cref{tab:metric_pitfalls}, a few related works proposed a similar analysis to the risk-coverage curves and AURC.
\citet{malinin_shifts_2022} propose ``error-retention'' curves, which replace rejected predictions with oracle predictions for all possible confidence thresholds. At low coverage, the average risk is hence dominated by these oracle predictions and high-confidence-high-risk predictions have relatively little impact on the AUC.
We prefer the risk-coverage curve (and AURC) in our benchmark as it is an established and well-studied measure \cite{el-yaniv_foundations_2010,jaeger_call_2022}.
Although named differently, the analysis in \citet{ng_estimating_2023} is equivalent to risk-coverage curves, and it is based on a binary risk by thresholding a segmentation metric. We avoid binarization as argued in the pitfalls to requirement R3 (\cref{sec:metric_pitfalls}).
For specific applications, there can be other suitable ways to summarize the risk-coverage curve than AURC. For example, if there are well-defined constraints on the target risk or coverage (e.g. the mean Dice score of all accepted samples should be above 0.9), another option is to define target regions (e.g. selective risk $\leq$ 0.1) in the risk-coverage space and compare individual operating points on the risk-coverage curves, similar to the SAC metric in \citet{galil_what_2023}.

While ensembles of neural networks are an established method for uncertainty quantification \citep{mehrtash_confidence_2020}, the combination with pairwise DSC has not been studied extensively, to the best of our knowledge. The original pairwise DSC publication focused on MC-dropout \citep{roy_bayesian_2019}, but others have also applied the method to ensembles \citep{hoebel_exploration_2020,hoebel_i_2022}, with a slight modification in \citet{ng_estimating_2023}. The latter study's results obtained on cardiac MRI data agree with ours in that deep ensembles perform best. Pearson correlation coefficients between uncertainty and true DSC from \citet{hoebel_exploration_2020,hoebel_i_2022} are not in line with ours (\cref{fig:pearson_mean_dice_all}), as MC-Dropout outperformed ensembles in their experiments
This discrepancy (also in terms of the Pearson correlation coefficient, which was used in their study) is surprising given that our Brain tumor dataset is similar (but not identical) to one of the two datasets in \citet{hoebel_i_2022}. Apart from test set differences, another possible explanation is the slightly different segmentation and MC-Dropout setups. As we make our code and benchmarking framework publicly available, we enable a fair and reproducible comparison in the future.
In summary, there have been mixed results for the ensemble + pairwise DSC method in the past. Our results provide evidence across multiple datasets that this method is in fact beneficial for failure detection.

Regarding pixel confidence aggregation, we are aware of two other studies that investigate this topic in depth \citep{jungo_analyzing_2020,kahl_values_2024}. In agreement with the results from \citet{jungo_analyzing_2020} on a brain tumor dataset, aggregation methods played an important role in our benchmark. However, their best-performing method (regression forest with radiomics confidence features) was outperformed by a simpler method we introduced (regression forest with simple features). Furthermore, in our multi-dataset evaluation, we found that the ranking of aggregation methods changed between datasets. This agrees with findings from \citet{kahl_values_2024}, who also examined mean and patch-based aggregation on one medical and one non-medical dataset.

Related works on quality regression networks have so far focused on a single anatomy, such as cardiac segmentation \citep{robinson_real-time_2018,li_towards_2022} and brain tumors \citep{greenspan_qcresunet_2023}.
A direct comparison to our results is not possible, because of the different dataset setup and the fact that these references balance the training distribution of target Dice scores.
Such a balancing could potentially improve upon our quality regression baseline, but we did not implement it in our benchmark due to the lack of reference code for these papers.
Despite this potential shortcoming, quality regression also performs well in our benchmark, which encourages further research in this field. However, we also observed performance degradations on two datasets that may be due to few training samples (Prostate) and strong distribution shift (Prostate, Covid), which suggests potential weaknesses of this method.

Our results clarify the discrepancy between OOD detection and failure detection. Compared to its original publication \citep{gonzalez_distance-based_2022}, the Mahalanobis method ranks worse in our failure detection benchmark. We explain this by noting that \citet{gonzalez_distance-based_2022} mainly use OOD detection metrics for evaluation, which do not measure actual failure detection performance. In fact, when we evaluated using OOD detection metrics (\cref{fig:ood-auc}), their method was superior in our experiments, too. These results consolidate empirical evidence from recent work \citep{greenspan_segmentation_2023} but in a large-scale evaluation.

Although our failure detection benchmark is unprecedented in terms of the diversity of methods and datasets, it has some limitations.
Since many recent works do not have public code (\ref{sec:code_availability}), it was infeasible to include all recently published methods in this study. Our goal was to re-implement a wide spectrum of FD approaches while keeping them simple to avoid implementation errors. For the Mahalanobis method, we evaluated our implementation with OOD detection metrics (\cref{fig:ood-auc}) to ensure comparable performance with the original publication \citep{gonzalez_distance-based_2022}. For the other methods, we performed extensive testing and limited tuning on the validation sets, but the exact reproduction of published results was infeasible due to the lack of a standardized dataset setup and reference implementation. By making the benchmarking framework available to the community, we hope that including new methods and existing variations in a fair and reproducible comparison becomes easier.

Some datasets may have annotation shifts between training and test sets, which can affect the results through noisy risk scores. For the brain tumor, heart, and kidney tumor datasets, this should not be an issue, as these datasets originate from international competitions with standardized annotation protocols. The Covid and Prostate datasets, however, are combinations of independently annotated datasets and could hence contain annotation shifts. Since all methods are affected equally by this issue and the rankings are stable for each dataset, we believe this effect is not too strong. Still, future work could further investigate the existence of annotation shifts and find suitable replacement datasets, if necessary. Extending the benchmark with new datasets or shifts is an important future direction also beyond the topic of annotation shifts.

As argued in the introduction, we focused on image-level failure detection, motivated by the practical scenario where a single confidence score is used to decide whether to retain a segmentation for downstream analyses. Of course, uncertainty estimation methods are also useful for other tasks than segmentation failure detection \citep{kahl_values_2024}. Furthermore, other levels of failure detection can be studied. Note, however, that class-level failure detection methods can be evaluated in a similar way to image-level methods by defining risk functions for each class separately, so the requirements formulated in \cref{sec:metric_pitfalls} apply here as well. However, some methods from our benchmark would need to be adapted to output class-level confidence scores. Pixel-level failure detection is equivalent to the task of classification failure detection from \citet{jaeger_call_2022} for each pixel, so the recommendations from this reference should apply.

\section{Conclusion}

In conclusion, our study addresses the pitfalls in existing evaluation protocols for segmentation failure detection by proposing a flexible evaluation pipeline based on a risk-coverage analysis. Using this pipeline, we introduced a benchmark comprising multiple radiological 3D datasets to assess the generalization of many failure detection methods, and found that the pairwise Dice score between ensemble predictions consistently outperforms other methods, serving as a strong baseline for future studies.

\section*{Acknowledgments}
Research reported in this publication was partially funded by the Helmholtz Association (HA) within the project ``Trustworthy Federated Data Analytics'' (TFDA) (funding number ZT-I-OO1 4) and by the German Federal Ministry of Education and Research (BMBF) Network of University Medicine 2.0: “NUM 2.0", Grant No. 01KX2121.
Paul Jaeger was funded by Helmholtz Imaging (HI), a platform of the Helmholtz Incubator on Information and Data Science.

\section*{Declaration of Generative AI and AI-assisted technologies in the writing process}
During the preparation of this work the author(s) used ChatGPT in order to improve readability and language. After using this tool/service, the author(s) reviewed and edited the content as needed and take(s) full responsibility for the content of the publication.




\bibliographystyle{elsarticle-harv} 
\bibliography{main}

\begin{thebibliography}{77}
\expandafter\ifx\csname natexlab\endcsname\relax\def\natexlab#1{#1}\fi
\providecommand{\url}[1]{\texttt{#1}}
\providecommand{\href}[2]{#2}
\providecommand{\path}[1]{#1}
\providecommand{\DOIprefix}{doi:}
\providecommand{\ArXivprefix}{arXiv:}
\providecommand{\URLprefix}{URL: }
\providecommand{\Pubmedprefix}{pmid:}
\providecommand{\doi}[1]{\href{http://dx.doi.org/#1}{\path{#1}}}
\providecommand{\Pubmed}[1]{\href{pmid:#1}{\path{#1}}}
\providecommand{\bibinfo}[2]{#2}
\ifx\xfnm\relax \def\xfnm[#1]{\unskip,\space#1}\fi
\bibitem[{Adams and Elhabian(2023)}]{adams_benchmarking_2023}
\bibinfo{author}{Adams, J.}, \bibinfo{author}{Elhabian, S.Y.}, \bibinfo{year}{2023}.
\newblock \bibinfo{title}{Benchmarking {Scalable} {Epistemic} {Uncertainty} {Quantification} in {Organ} {Segmentation}}.
\newblock \URLprefix \url{http://arxiv.org/abs/2308.07506}, \DOIprefix\doi{10.48550/arXiv.2308.07506}. \bibinfo{note}{arXiv:2308.07506 [cs, eess]}.
\bibitem[{AlBadawy et~al.(2018)AlBadawy, Saha and Mazurowski}]{albadawy_deep_2018}
\bibinfo{author}{AlBadawy, E.A.}, \bibinfo{author}{Saha, A.}, \bibinfo{author}{Mazurowski, M.A.}, \bibinfo{year}{2018}.
\newblock \bibinfo{title}{Deep learning for segmentation of brain tumors: {Impact} of cross-institutional training and testing}.
\newblock \bibinfo{journal}{Medical Physics} \bibinfo{volume}{45}, \bibinfo{pages}{1150--1158}.
\newblock \URLprefix \url{https://aapm.onlinelibrary.wiley.com/doi/abs/10.1002/mp.12752}, \DOIprefix\doi{10.1002/mp.12752}. \bibinfo{note}{\_eprint: https://aapm.onlinelibrary.wiley.com/doi/pdf/10.1002/mp.12752}.
\bibitem[{An et~al.(2020)An, Xu, Harmon, Turkbey, Sanford, Amalou, Kassin, Varble, Blain, Anderson, Patella, Carrafiello, Turkbey and Wood}]{an_ct_2020}
\bibinfo{author}{An, P.}, \bibinfo{author}{Xu, S.}, \bibinfo{author}{Harmon, S.A.}, \bibinfo{author}{Turkbey, E.B.}, \bibinfo{author}{Sanford, T.H.}, \bibinfo{author}{Amalou, A.}, \bibinfo{author}{Kassin, M.}, \bibinfo{author}{Varble, N.}, \bibinfo{author}{Blain, M.}, \bibinfo{author}{Anderson, V.}, \bibinfo{author}{Patella, F.}, \bibinfo{author}{Carrafiello, G.}, \bibinfo{author}{Turkbey, B.T.}, \bibinfo{author}{Wood, B.J.}, \bibinfo{year}{2020}.
\newblock \bibinfo{title}{{CT} {Images} in {COVID}-19}.
\newblock \URLprefix \url{https://www.cancerimagingarchive.net/collection/ct-images-in-covid-19/}, \DOIprefix\doi{10.7937/TCIA.2020.GQRY-NC81}.
\bibitem[{Antonelli et~al.(2022)Antonelli, Reinke, Bakas, Farahani, Kopp-Schneider, Landman, Litjens, Menze, Ronneberger, Summers, van Ginneken, Bilello, Bilic, Christ, Do, Gollub, Heckers, Huisman, Jarnagin, McHugo, Napel, Pernicka, Rhode, Tobon-Gomez, Vorontsov, Meakin, Ourselin, Wiesenfarth, Arbeláez, Bae, Chen, Daza, Feng, He, Isensee, Ji, Jia, Kim, Maier-Hein, Merhof, Pai, Park, Perslev, Rezaiifar, Rippel, Sarasua, Shen, Son, Wachinger, Wang, Wang, Xia, Xu, Xu, Zheng, Simpson, Maier-Hein and Cardoso}]{antonelli_medical_2022}
\bibinfo{author}{Antonelli, M.}, \bibinfo{author}{Reinke, A.}, \bibinfo{author}{Bakas, S.}, \bibinfo{author}{Farahani, K.}, \bibinfo{author}{Kopp-Schneider, A.}, \bibinfo{author}{Landman, B.A.}, \bibinfo{author}{Litjens, G.}, \bibinfo{author}{Menze, B.}, \bibinfo{author}{Ronneberger, O.}, \bibinfo{author}{Summers, R.M.}, \bibinfo{author}{van Ginneken, B.}, \bibinfo{author}{Bilello, M.}, \bibinfo{author}{Bilic, P.}, \bibinfo{author}{Christ, P.F.}, \bibinfo{author}{Do, R.K.G.}, \bibinfo{author}{Gollub, M.J.}, \bibinfo{author}{Heckers, S.H.}, \bibinfo{author}{Huisman, H.}, \bibinfo{author}{Jarnagin, W.R.}, \bibinfo{author}{McHugo, M.K.}, \bibinfo{author}{Napel, S.}, \bibinfo{author}{Pernicka, J.S.G.}, \bibinfo{author}{Rhode, K.}, \bibinfo{author}{Tobon-Gomez, C.}, \bibinfo{author}{Vorontsov, E.}, \bibinfo{author}{Meakin, J.A.}, \bibinfo{author}{Ourselin, S.}, \bibinfo{author}{Wiesenfarth, M.}, \bibinfo{author}{Arbeláez, P.}, \bibinfo{author}{Bae, B.}, \bibinfo{author}{Chen, S.}, \bibinfo{author}{Daza, L.},
  \bibinfo{author}{Feng, J.}, \bibinfo{author}{He, B.}, \bibinfo{author}{Isensee, F.}, \bibinfo{author}{Ji, Y.}, \bibinfo{author}{Jia, F.}, \bibinfo{author}{Kim, I.}, \bibinfo{author}{Maier-Hein, K.}, \bibinfo{author}{Merhof, D.}, \bibinfo{author}{Pai, A.}, \bibinfo{author}{Park, B.}, \bibinfo{author}{Perslev, M.}, \bibinfo{author}{Rezaiifar, R.}, \bibinfo{author}{Rippel, O.}, \bibinfo{author}{Sarasua, I.}, \bibinfo{author}{Shen, W.}, \bibinfo{author}{Son, J.}, \bibinfo{author}{Wachinger, C.}, \bibinfo{author}{Wang, L.}, \bibinfo{author}{Wang, Y.}, \bibinfo{author}{Xia, Y.}, \bibinfo{author}{Xu, D.}, \bibinfo{author}{Xu, Z.}, \bibinfo{author}{Zheng, Y.}, \bibinfo{author}{Simpson, A.L.}, \bibinfo{author}{Maier-Hein, L.}, \bibinfo{author}{Cardoso, M.J.}, \bibinfo{year}{2022}.
\newblock \bibinfo{title}{The {Medical} {Segmentation} {Decathlon}}.
\newblock \bibinfo{journal}{Nature Communications} \bibinfo{volume}{13}, \bibinfo{pages}{4128}.
\newblock \URLprefix \url{https://www.nature.com/articles/s41467-022-30695-9}, \DOIprefix\doi{10.1038/s41467-022-30695-9}. \bibinfo{note}{number: 1 Publisher: Nature Publishing Group}.
\bibitem[{Badgeley et~al.(2019)Badgeley, Zech, Oakden-Rayner, Glicksberg, Liu, Gale, McConnell, Percha, Snyder and Dudley}]{badgeley_deep_2019}
\bibinfo{author}{Badgeley, M.A.}, \bibinfo{author}{Zech, J.R.}, \bibinfo{author}{Oakden-Rayner, L.}, \bibinfo{author}{Glicksberg, B.S.}, \bibinfo{author}{Liu, M.}, \bibinfo{author}{Gale, W.}, \bibinfo{author}{McConnell, M.V.}, \bibinfo{author}{Percha, B.}, \bibinfo{author}{Snyder, T.M.}, \bibinfo{author}{Dudley, J.T.}, \bibinfo{year}{2019}.
\newblock \bibinfo{title}{Deep learning predicts hip fracture using confounding patient and healthcare variables}.
\newblock \bibinfo{journal}{npj Digital Medicine} \bibinfo{volume}{2}, \bibinfo{pages}{1--10}.
\newblock \URLprefix \url{https://www.nature.com/articles/s41746-019-0105-1}, \DOIprefix\doi{10.1038/s41746-019-0105-1}. \bibinfo{note}{number: 1 Publisher: Nature Publishing Group}.
\bibitem[{Bakas et~al.(2017)Bakas, Akbari, Sotiras, Bilello, Rozycki, Kirby, Freymann, Farahani and Davatzikos}]{bakas_advancing_2017}
\bibinfo{author}{Bakas, S.}, \bibinfo{author}{Akbari, H.}, \bibinfo{author}{Sotiras, A.}, \bibinfo{author}{Bilello, M.}, \bibinfo{author}{Rozycki, M.}, \bibinfo{author}{Kirby, J.S.}, \bibinfo{author}{Freymann, J.B.}, \bibinfo{author}{Farahani, K.}, \bibinfo{author}{Davatzikos, C.}, \bibinfo{year}{2017}.
\newblock \bibinfo{title}{Advancing {The} {Cancer} {Genome} {Atlas} glioma {MRI} collections with expert segmentation labels and radiomic features}.
\newblock \bibinfo{journal}{Scientific Data} \bibinfo{volume}{4}, \bibinfo{pages}{170117}.
\newblock \DOIprefix\doi{10.1038/sdata.2017.117}.
\bibitem[{Bakas et~al.(2019)Bakas, Reyes, Jakab, Bauer, Rempfler, Crimi, Shinohara, Berger, Ha, Rozycki, Prastawa, Alberts, Lipkova, Freymann, Kirby, Bilello, Fathallah-Shaykh, Wiest, Kirschke, Wiestler, Colen, Kotrotsou, Lamontagne, Marcus, Milchenko, Nazeri, Weber, Mahajan, Baid, Gerstner, Kwon, Acharya, Agarwal, Alam, Albiol, Albiol, Albiol, Alex, Allinson, Amorim, Amrutkar, Anand, Andermatt, Arbel, Arbelaez, Avery, Azmat, B., Bai, Banerjee, Barth, Batchelder, Batmanghelich, Battistella, Beers, Belyaev, Bendszus, Benson, Bernal, Bharath, Biros, Bisdas, Brown, Cabezas, Cao, Cardoso, Carver, Casamitjana, Castillo, Catà, Cattin, Cerigues, Chagas, Chandra, Chang, Chang, Chang, Chazalon, Chen, Chen, Chen, Chen, Cheng, Choudhury, Chylla, Clérigues, Colleman, Colmeiro, Combalia, Costa, Cui, Dai, Dai, Daza, Deutsch, Ding, Dong, Dong, Dudzik, Eaton-Rosen, Egan, Escudero, Estienne, Everson, Fabrizio, Fan, Fang, Feng, Ferrante, Fidon, Fischer, French, Fridman, Fu, Fuentes, Gao, Gates, Gering, Gholami, Gierke,
  Glocker, Gong, González-Villá, Grosges, Guan, Guo, Gupta, Han, Han, Harmuth, He, Hernández-Sabaté, Herrmann, Himthani, Hsu, Hsu, Hu, Hu, Hu, Hu, Hua, Huang, Huang, Van~Huffel, Huo, HV, Iftekharuddin, Isensee, Islam, Jackson, Jambawalikar, Jesson, Jian, Jin, Jose, Jungo, Kainz, Kamnitsas, Kao, Karnawat, Kellermeier, Kermi, Keutzer, Khadir, Khened, Kickingereder, Kim, King, Knapp, Knecht, Kohli, Kong, Kong, Koppers, Kori, Krishnamurthi, Krivov, Kumar, Kushibar, Lachinov, Lambrou, Lee, Lee, Lee, Lee, Lefkovits, Lefkovits, Levitt, Li, Li, Li, Li, Li, Li, Li, Li, Li, Li, Li, Li, Lin, Lin, Lio, Liu, Liu, Liu, Liu, Liu, Liu, Llado, Lopez, Lorenzo, Lu, Luo, Luo, Ma, Ma, Mackie, Madabushi, Mahmoudi, Maier-Hein, Maji, Mammen, Mang, Manjunath, Marcinkiewicz, McDonagh, McKenna, McKinley, Mehl, Mehta, Mehta, Meier, Meinel, Merhof, Meyer, Miller, Mitra, Moiyadi, Molina-Garcia, Monteiro, Mrukwa, Myronenko, Nalepa, Ngo, Nie, Ning, Niu, Nuechterlein, Oermann, Oliveira, Oliveira, Oliver, Osman, Ou, Ourselin, Paragios,
  Park, Paschke, Pauloski, Pawar, Pawlowski, Pei, Peng, Pereira, Perez-Beteta, Perez-Garcia, Pezold, Pham, Phophalia, Piella, Pillai, Piraud, Pisov, Popli, Pound, Pourreza, Prasanna, Prkovska, Pridmore, Puch, Puybareau, Qian, Qiao, Rajchl, Rane, Rebsamen, Ren, Ren, Revanuru, Rezaei, Rippel, Rivera, Robert, Rosen, Rueckert, Safwan, Salem, Salvi, Sanchez, Sánchez, Santos, Sartor, Schellingerhout, Scheufele, Scott, Scussel, Sedlar, Serrano-Rubio, Shah, Shah, Shaikh, Shankar, Shboul, Shen, Shen, Shen, Shen, Shenoy, Shi, Shin, Shu, Sima, Sinclair, Smedby, Snyder, Soltaninejad, Song, Soni, Stawiaski, Subramanian, Sun, Sun, Sun, Sun, Sun, Sun, Sun, Suter, Szilagyi, Talbar, Tao, Tao, Teng, Thakur, Thakur, Tharakan, Tiwari, Tochon, Tran, Tsai, Tseng, Tuan, Turlapov, Tustison, Vakalopoulou, Valverde, Vanguri, Vasiliev, Ventura, Vera, Vercauteren, Verrastro, Vidyaratne, Vilaplana, Vivekanandan, Wang, Wang, Wang, Wang, Wang, Wang, Wang, Wang, Wang, Wen, Wen, Weninger, Wick, Wu, Wu, Wu, Xia, Xu, Xu, Xu, Yang, Yang, Yang,
  Yang, Yang, Yang, Yao, Ye, Yin, Young-Moxon, Yu, Yue, Zhang, Zhang, Zhang, Zhang, Zhang, Zhang, Zhang, Zhang, Zhang, Zhang, Zhang, Zhao, Zhao, Zhao, Zhao, Zheng, Zhong, Zhou, Zhou, Zhou, Zhu, Zhu, Zhuge, Zong, Kalpathy-Cramer, Farahani, Davatzikos, van Leemput and Menze}]{bakas_identifying_2019}
\bibinfo{author}{Bakas, S.}, \bibinfo{author}{Reyes, M.}, \bibinfo{author}{Jakab, A.}, \bibinfo{author}{Bauer, S.}, \bibinfo{author}{Rempfler, M.}, \bibinfo{author}{Crimi, A.}, \bibinfo{author}{Shinohara, R.T.}, \bibinfo{author}{Berger, C.}, \bibinfo{author}{Ha, S.M.}, \bibinfo{author}{Rozycki, M.}, \bibinfo{author}{Prastawa, M.}, \bibinfo{author}{Alberts, E.}, \bibinfo{author}{Lipkova, J.}, \bibinfo{author}{Freymann, J.}, \bibinfo{author}{Kirby, J.}, \bibinfo{author}{Bilello, M.}, \bibinfo{author}{Fathallah-Shaykh, H.}, \bibinfo{author}{Wiest, R.}, \bibinfo{author}{Kirschke, J.}, \bibinfo{author}{Wiestler, B.}, \bibinfo{author}{Colen, R.}, \bibinfo{author}{Kotrotsou, A.}, \bibinfo{author}{Lamontagne, P.}, \bibinfo{author}{Marcus, D.}, \bibinfo{author}{Milchenko, M.}, \bibinfo{author}{Nazeri, A.}, \bibinfo{author}{Weber, M.A.}, \bibinfo{author}{Mahajan, A.}, \bibinfo{author}{Baid, U.}, \bibinfo{author}{Gerstner, E.}, \bibinfo{author}{Kwon, D.}, \bibinfo{author}{Acharya, G.}, \bibinfo{author}{Agarwal, M.},
  \bibinfo{author}{Alam, M.}, \bibinfo{author}{Albiol, A.}, \bibinfo{author}{Albiol, A.}, \bibinfo{author}{Albiol, F.J.}, \bibinfo{author}{Alex, V.}, \bibinfo{author}{Allinson, N.}, \bibinfo{author}{Amorim, P.H.A.}, \bibinfo{author}{Amrutkar, A.}, \bibinfo{author}{Anand, G.}, \bibinfo{author}{Andermatt, S.}, \bibinfo{author}{Arbel, T.}, \bibinfo{author}{Arbelaez, P.}, \bibinfo{author}{Avery, A.}, \bibinfo{author}{Azmat, M.}, \bibinfo{author}{B., P.}, \bibinfo{author}{Bai, W.}, \bibinfo{author}{Banerjee, S.}, \bibinfo{author}{Barth, B.}, \bibinfo{author}{Batchelder, T.}, \bibinfo{author}{Batmanghelich, K.}, \bibinfo{author}{Battistella, E.}, \bibinfo{author}{Beers, A.}, \bibinfo{author}{Belyaev, M.}, \bibinfo{author}{Bendszus, M.}, \bibinfo{author}{Benson, E.}, \bibinfo{author}{Bernal, J.}, \bibinfo{author}{Bharath, H.N.}, \bibinfo{author}{Biros, G.}, \bibinfo{author}{Bisdas, S.}, \bibinfo{author}{Brown, J.}, \bibinfo{author}{Cabezas, M.}, \bibinfo{author}{Cao, S.}, \bibinfo{author}{Cardoso, J.M.},
  \bibinfo{author}{Carver, E.N.}, \bibinfo{author}{Casamitjana, A.}, \bibinfo{author}{Castillo, L.S.}, \bibinfo{author}{Catà, M.}, \bibinfo{author}{Cattin, P.}, \bibinfo{author}{Cerigues, A.}, \bibinfo{author}{Chagas, V.S.}, \bibinfo{author}{Chandra, S.}, \bibinfo{author}{Chang, Y.J.}, \bibinfo{author}{Chang, S.}, \bibinfo{author}{Chang, K.}, \bibinfo{author}{Chazalon, J.}, \bibinfo{author}{Chen, S.}, \bibinfo{author}{Chen, W.}, \bibinfo{author}{Chen, J.W.}, \bibinfo{author}{Chen, Z.}, \bibinfo{author}{Cheng, K.}, \bibinfo{author}{Choudhury, A.R.}, \bibinfo{author}{Chylla, R.}, \bibinfo{author}{Clérigues, A.}, \bibinfo{author}{Colleman, S.}, \bibinfo{author}{Colmeiro, R.G.R.}, \bibinfo{author}{Combalia, M.}, \bibinfo{author}{Costa, A.}, \bibinfo{author}{Cui, X.}, \bibinfo{author}{Dai, Z.}, \bibinfo{author}{Dai, L.}, \bibinfo{author}{Daza, L.A.}, \bibinfo{author}{Deutsch, E.}, \bibinfo{author}{Ding, C.}, \bibinfo{author}{Dong, C.}, \bibinfo{author}{Dong, S.}, \bibinfo{author}{Dudzik, W.},
  \bibinfo{author}{Eaton-Rosen, Z.}, \bibinfo{author}{Egan, G.}, \bibinfo{author}{Escudero, G.}, \bibinfo{author}{Estienne, T.}, \bibinfo{author}{Everson, R.}, \bibinfo{author}{Fabrizio, J.}, \bibinfo{author}{Fan, Y.}, \bibinfo{author}{Fang, L.}, \bibinfo{author}{Feng, X.}, \bibinfo{author}{Ferrante, E.}, \bibinfo{author}{Fidon, L.}, \bibinfo{author}{Fischer, M.}, \bibinfo{author}{French, A.P.}, \bibinfo{author}{Fridman, N.}, \bibinfo{author}{Fu, H.}, \bibinfo{author}{Fuentes, D.}, \bibinfo{author}{Gao, Y.}, \bibinfo{author}{Gates, E.}, \bibinfo{author}{Gering, D.}, \bibinfo{author}{Gholami, A.}, \bibinfo{author}{Gierke, W.}, \bibinfo{author}{Glocker, B.}, \bibinfo{author}{Gong, M.}, \bibinfo{author}{González-Villá, S.}, \bibinfo{author}{Grosges, T.}, \bibinfo{author}{Guan, Y.}, \bibinfo{author}{Guo, S.}, \bibinfo{author}{Gupta, S.}, \bibinfo{author}{Han, W.S.}, \bibinfo{author}{Han, I.S.}, \bibinfo{author}{Harmuth, K.}, \bibinfo{author}{He, H.}, \bibinfo{author}{Hernández-Sabaté, A.},
  \bibinfo{author}{Herrmann, E.}, \bibinfo{author}{Himthani, N.}, \bibinfo{author}{Hsu, W.}, \bibinfo{author}{Hsu, C.}, \bibinfo{author}{Hu, X.}, \bibinfo{author}{Hu, X.}, \bibinfo{author}{Hu, Y.}, \bibinfo{author}{Hu, Y.}, \bibinfo{author}{Hua, R.}, \bibinfo{author}{Huang, T.Y.}, \bibinfo{author}{Huang, W.}, \bibinfo{author}{Van~Huffel, S.}, \bibinfo{author}{Huo, Q.}, \bibinfo{author}{HV, V.}, \bibinfo{author}{Iftekharuddin, K.M.}, \bibinfo{author}{Isensee, F.}, \bibinfo{author}{Islam, M.}, \bibinfo{author}{Jackson, A.S.}, \bibinfo{author}{Jambawalikar, S.R.}, \bibinfo{author}{Jesson, A.}, \bibinfo{author}{Jian, W.}, \bibinfo{author}{Jin, P.}, \bibinfo{author}{Jose, V.J.M.}, \bibinfo{author}{Jungo, A.}, \bibinfo{author}{Kainz, B.}, \bibinfo{author}{Kamnitsas, K.}, \bibinfo{author}{Kao, P.Y.}, \bibinfo{author}{Karnawat, A.}, \bibinfo{author}{Kellermeier, T.}, \bibinfo{author}{Kermi, A.}, \bibinfo{author}{Keutzer, K.}, \bibinfo{author}{Khadir, M.T.}, \bibinfo{author}{Khened, M.},
  \bibinfo{author}{Kickingereder, P.}, \bibinfo{author}{Kim, G.}, \bibinfo{author}{King, N.}, \bibinfo{author}{Knapp, H.}, \bibinfo{author}{Knecht, U.}, \bibinfo{author}{Kohli, L.}, \bibinfo{author}{Kong, D.}, \bibinfo{author}{Kong, X.}, \bibinfo{author}{Koppers, S.}, \bibinfo{author}{Kori, A.}, \bibinfo{author}{Krishnamurthi, G.}, \bibinfo{author}{Krivov, E.}, \bibinfo{author}{Kumar, P.}, \bibinfo{author}{Kushibar, K.}, \bibinfo{author}{Lachinov, D.}, \bibinfo{author}{Lambrou, T.}, \bibinfo{author}{Lee, J.}, \bibinfo{author}{Lee, C.}, \bibinfo{author}{Lee, Y.}, \bibinfo{author}{Lee, M.}, \bibinfo{author}{Lefkovits, S.}, \bibinfo{author}{Lefkovits, L.}, \bibinfo{author}{Levitt, J.}, \bibinfo{author}{Li, T.}, \bibinfo{author}{Li, H.}, \bibinfo{author}{Li, W.}, \bibinfo{author}{Li, H.}, \bibinfo{author}{Li, X.}, \bibinfo{author}{Li, Y.}, \bibinfo{author}{Li, H.}, \bibinfo{author}{Li, Z.}, \bibinfo{author}{Li, X.}, \bibinfo{author}{Li, Z.}, \bibinfo{author}{Li, X.}, \bibinfo{author}{Li, W.},
  \bibinfo{author}{Lin, Z.S.}, \bibinfo{author}{Lin, F.}, \bibinfo{author}{Lio, P.}, \bibinfo{author}{Liu, C.}, \bibinfo{author}{Liu, B.}, \bibinfo{author}{Liu, X.}, \bibinfo{author}{Liu, M.}, \bibinfo{author}{Liu, J.}, \bibinfo{author}{Liu, L.}, \bibinfo{author}{Llado, X.}, \bibinfo{author}{Lopez, M.M.}, \bibinfo{author}{Lorenzo, P.R.}, \bibinfo{author}{Lu, Z.}, \bibinfo{author}{Luo, L.}, \bibinfo{author}{Luo, Z.}, \bibinfo{author}{Ma, J.}, \bibinfo{author}{Ma, K.}, \bibinfo{author}{Mackie, T.}, \bibinfo{author}{Madabushi, A.}, \bibinfo{author}{Mahmoudi, I.}, \bibinfo{author}{Maier-Hein, K.H.}, \bibinfo{author}{Maji, P.}, \bibinfo{author}{Mammen, C.P.}, \bibinfo{author}{Mang, A.}, \bibinfo{author}{Manjunath, B.S.}, \bibinfo{author}{Marcinkiewicz, M.}, \bibinfo{author}{McDonagh, S.}, \bibinfo{author}{McKenna, S.}, \bibinfo{author}{McKinley, R.}, \bibinfo{author}{Mehl, M.}, \bibinfo{author}{Mehta, S.}, \bibinfo{author}{Mehta, R.}, \bibinfo{author}{Meier, R.}, \bibinfo{author}{Meinel, C.},
  \bibinfo{author}{Merhof, D.}, \bibinfo{author}{Meyer, C.}, \bibinfo{author}{Miller, R.}, \bibinfo{author}{Mitra, S.}, \bibinfo{author}{Moiyadi, A.}, \bibinfo{author}{Molina-Garcia, D.}, \bibinfo{author}{Monteiro, M.A.B.}, \bibinfo{author}{Mrukwa, G.}, \bibinfo{author}{Myronenko, A.}, \bibinfo{author}{Nalepa, J.}, \bibinfo{author}{Ngo, T.}, \bibinfo{author}{Nie, D.}, \bibinfo{author}{Ning, H.}, \bibinfo{author}{Niu, C.}, \bibinfo{author}{Nuechterlein, N.K.}, \bibinfo{author}{Oermann, E.}, \bibinfo{author}{Oliveira, A.}, \bibinfo{author}{Oliveira, D.D.C.}, \bibinfo{author}{Oliver, A.}, \bibinfo{author}{Osman, A.F.I.}, \bibinfo{author}{Ou, Y.N.}, \bibinfo{author}{Ourselin, S.}, \bibinfo{author}{Paragios, N.}, \bibinfo{author}{Park, M.S.}, \bibinfo{author}{Paschke, B.}, \bibinfo{author}{Pauloski, J.G.}, \bibinfo{author}{Pawar, K.}, \bibinfo{author}{Pawlowski, N.}, \bibinfo{author}{Pei, L.}, \bibinfo{author}{Peng, S.}, \bibinfo{author}{Pereira, S.M.}, \bibinfo{author}{Perez-Beteta, J.},
  \bibinfo{author}{Perez-Garcia, V.M.}, \bibinfo{author}{Pezold, S.}, \bibinfo{author}{Pham, B.}, \bibinfo{author}{Phophalia, A.}, \bibinfo{author}{Piella, G.}, \bibinfo{author}{Pillai, G.N.}, \bibinfo{author}{Piraud, M.}, \bibinfo{author}{Pisov, M.}, \bibinfo{author}{Popli, A.}, \bibinfo{author}{Pound, M.P.}, \bibinfo{author}{Pourreza, R.}, \bibinfo{author}{Prasanna, P.}, \bibinfo{author}{Prkovska, V.}, \bibinfo{author}{Pridmore, T.P.}, \bibinfo{author}{Puch, S.}, \bibinfo{author}{Puybareau, E.}, \bibinfo{author}{Qian, B.}, \bibinfo{author}{Qiao, X.}, \bibinfo{author}{Rajchl, M.}, \bibinfo{author}{Rane, S.}, \bibinfo{author}{Rebsamen, M.}, \bibinfo{author}{Ren, H.}, \bibinfo{author}{Ren, X.}, \bibinfo{author}{Revanuru, K.}, \bibinfo{author}{Rezaei, M.}, \bibinfo{author}{Rippel, O.}, \bibinfo{author}{Rivera, L.C.}, \bibinfo{author}{Robert, C.}, \bibinfo{author}{Rosen, B.}, \bibinfo{author}{Rueckert, D.}, \bibinfo{author}{Safwan, M.}, \bibinfo{author}{Salem, M.}, \bibinfo{author}{Salvi, J.},
  \bibinfo{author}{Sanchez, I.}, \bibinfo{author}{Sánchez, I.}, \bibinfo{author}{Santos, H.M.}, \bibinfo{author}{Sartor, E.}, \bibinfo{author}{Schellingerhout, D.}, \bibinfo{author}{Scheufele, K.}, \bibinfo{author}{Scott, M.R.}, \bibinfo{author}{Scussel, A.A.}, \bibinfo{author}{Sedlar, S.}, \bibinfo{author}{Serrano-Rubio, J.P.}, \bibinfo{author}{Shah, N.J.}, \bibinfo{author}{Shah, N.}, \bibinfo{author}{Shaikh, M.}, \bibinfo{author}{Shankar, B.U.}, \bibinfo{author}{Shboul, Z.}, \bibinfo{author}{Shen, H.}, \bibinfo{author}{Shen, D.}, \bibinfo{author}{Shen, L.}, \bibinfo{author}{Shen, H.}, \bibinfo{author}{Shenoy, V.}, \bibinfo{author}{Shi, F.}, \bibinfo{author}{Shin, H.E.}, \bibinfo{author}{Shu, H.}, \bibinfo{author}{Sima, D.}, \bibinfo{author}{Sinclair, M.}, \bibinfo{author}{Smedby, O.}, \bibinfo{author}{Snyder, J.M.}, \bibinfo{author}{Soltaninejad, M.}, \bibinfo{author}{Song, G.}, \bibinfo{author}{Soni, M.}, \bibinfo{author}{Stawiaski, J.}, \bibinfo{author}{Subramanian, S.}, \bibinfo{author}{Sun, L.},
  \bibinfo{author}{Sun, R.}, \bibinfo{author}{Sun, J.}, \bibinfo{author}{Sun, K.}, \bibinfo{author}{Sun, Y.}, \bibinfo{author}{Sun, G.}, \bibinfo{author}{Sun, S.}, \bibinfo{author}{Suter, Y.R.}, \bibinfo{author}{Szilagyi, L.}, \bibinfo{author}{Talbar, S.}, \bibinfo{author}{Tao, D.}, \bibinfo{author}{Tao, D.}, \bibinfo{author}{Teng, Z.}, \bibinfo{author}{Thakur, S.}, \bibinfo{author}{Thakur, M.H.}, \bibinfo{author}{Tharakan, S.}, \bibinfo{author}{Tiwari, P.}, \bibinfo{author}{Tochon, G.}, \bibinfo{author}{Tran, T.}, \bibinfo{author}{Tsai, Y.M.}, \bibinfo{author}{Tseng, K.L.}, \bibinfo{author}{Tuan, T.A.}, \bibinfo{author}{Turlapov, V.}, \bibinfo{author}{Tustison, N.}, \bibinfo{author}{Vakalopoulou, M.}, \bibinfo{author}{Valverde, S.}, \bibinfo{author}{Vanguri, R.}, \bibinfo{author}{Vasiliev, E.}, \bibinfo{author}{Ventura, J.}, \bibinfo{author}{Vera, L.}, \bibinfo{author}{Vercauteren, T.}, \bibinfo{author}{Verrastro, C.A.}, \bibinfo{author}{Vidyaratne, L.}, \bibinfo{author}{Vilaplana, V.},
  \bibinfo{author}{Vivekanandan, A.}, \bibinfo{author}{Wang, G.}, \bibinfo{author}{Wang, Q.}, \bibinfo{author}{Wang, C.J.}, \bibinfo{author}{Wang, W.}, \bibinfo{author}{Wang, D.}, \bibinfo{author}{Wang, R.}, \bibinfo{author}{Wang, Y.}, \bibinfo{author}{Wang, C.}, \bibinfo{author}{Wang, G.}, \bibinfo{author}{Wen, N.}, \bibinfo{author}{Wen, X.}, \bibinfo{author}{Weninger, L.}, \bibinfo{author}{Wick, W.}, \bibinfo{author}{Wu, S.}, \bibinfo{author}{Wu, Q.}, \bibinfo{author}{Wu, Y.}, \bibinfo{author}{Xia, Y.}, \bibinfo{author}{Xu, Y.}, \bibinfo{author}{Xu, X.}, \bibinfo{author}{Xu, P.}, \bibinfo{author}{Yang, T.L.}, \bibinfo{author}{Yang, X.}, \bibinfo{author}{Yang, H.Y.}, \bibinfo{author}{Yang, J.}, \bibinfo{author}{Yang, H.}, \bibinfo{author}{Yang, G.}, \bibinfo{author}{Yao, H.}, \bibinfo{author}{Ye, X.}, \bibinfo{author}{Yin, C.}, \bibinfo{author}{Young-Moxon, B.}, \bibinfo{author}{Yu, J.}, \bibinfo{author}{Yue, X.}, \bibinfo{author}{Zhang, S.}, \bibinfo{author}{Zhang, A.}, \bibinfo{author}{Zhang, K.},
  \bibinfo{author}{Zhang, X.}, \bibinfo{author}{Zhang, L.}, \bibinfo{author}{Zhang, X.}, \bibinfo{author}{Zhang, Y.}, \bibinfo{author}{Zhang, L.}, \bibinfo{author}{Zhang, J.}, \bibinfo{author}{Zhang, X.}, \bibinfo{author}{Zhang, T.}, \bibinfo{author}{Zhao, S.}, \bibinfo{author}{Zhao, Y.}, \bibinfo{author}{Zhao, X.}, \bibinfo{author}{Zhao, L.}, \bibinfo{author}{Zheng, Y.}, \bibinfo{author}{Zhong, L.}, \bibinfo{author}{Zhou, C.}, \bibinfo{author}{Zhou, X.}, \bibinfo{author}{Zhou, F.}, \bibinfo{author}{Zhu, H.}, \bibinfo{author}{Zhu, J.}, \bibinfo{author}{Zhuge, Y.}, \bibinfo{author}{Zong, W.}, \bibinfo{author}{Kalpathy-Cramer, J.}, \bibinfo{author}{Farahani, K.}, \bibinfo{author}{Davatzikos, C.}, \bibinfo{author}{van Leemput, K.}, \bibinfo{author}{Menze, B.}, \bibinfo{year}{2019}.
\newblock \bibinfo{title}{Identifying the {Best} {Machine} {Learning} {Algorithms} for {Brain} {Tumor} {Segmentation}, {Progression} {Assessment}, and {Overall} {Survival} {Prediction} in the {BRATS} {Challenge}}.
\newblock \bibinfo{journal}{arXiv:1811.02629 [cs, stat]} \URLprefix \url{http://arxiv.org/abs/1811.02629}. \bibinfo{note}{arXiv: 1811.02629}.
\bibitem[{Beede et~al.(2020)Beede, Baylor, Hersch, Iurchenko, Wilcox, Ruamviboonsuk and Vardoulakis}]{beede_human-centered_2020}
\bibinfo{author}{Beede, E.}, \bibinfo{author}{Baylor, E.}, \bibinfo{author}{Hersch, F.}, \bibinfo{author}{Iurchenko, A.}, \bibinfo{author}{Wilcox, L.}, \bibinfo{author}{Ruamviboonsuk, P.}, \bibinfo{author}{Vardoulakis, L.M.}, \bibinfo{year}{2020}.
\newblock \bibinfo{title}{A {Human}-{Centered} {Evaluation} of a {Deep} {Learning} {System} {Deployed} in {Clinics} for the {Detection} of {Diabetic} {Retinopathy}}, in: \bibinfo{booktitle}{Proceedings of the 2020 {CHI} {Conference} on {Human} {Factors} in {Computing} {Systems}}, \bibinfo{publisher}{Association for Computing Machinery}, \bibinfo{address}{New York, NY, USA}. pp. \bibinfo{pages}{1--12}.
\newblock \URLprefix \url{https://dl.acm.org/doi/10.1145/3313831.3376718}, \DOIprefix\doi{10.1145/3313831.3376718}.
\bibitem[{Bloch et~al.(2015)Bloch, Madabhushi, Huisman, Freymann, Kirby, Grauer, Enquobahrie, Jaffe, Clarke and Farahani}]{bloch_nci-isbi_2015}
\bibinfo{author}{Bloch, B.N.}, \bibinfo{author}{Madabhushi, A.}, \bibinfo{author}{Huisman, H.}, \bibinfo{author}{Freymann, J.}, \bibinfo{author}{Kirby, J.}, \bibinfo{author}{Grauer, M.}, \bibinfo{author}{Enquobahrie, A.}, \bibinfo{author}{Jaffe, C.}, \bibinfo{author}{Clarke, L.}, \bibinfo{author}{Farahani, K.}, \bibinfo{year}{2015}.
\newblock \bibinfo{title}{{NCI}-{ISBI} 2013 {Challenge}: {Automated} {Segmentation} of {Prostate} {Structures} ({ISBI}-{MR}-{Prostate}-2013)}.
\newblock \URLprefix \url{https://www.cancerimagingarchive.net/analysis-result/isbi-mr-prostate-2013/}, \DOIprefix\doi{10.7937/K9/TCIA.2015.ZF0VLOPV}.
\bibitem[{Campello et~al.(2021)Campello, Gkontra, Izquierdo, Martín-Isla, Sojoudi, Full, Maier-Hein, Zhang, He, Ma, Parreño, Albiol, Kong, Shadden, Acero, Sundaresan, Saber, Elattar, Li, Menze, Khader, Haarburger, Scannell, Veta, Carscadden, Punithakumar, Liu, Tsaftaris, Huang, Yang, Li, Zhuang, Viladés, Descalzo, Guala, Mura, Friedrich, Garg, Lebel, Henriques, Karakas, Çavuş, Petersen, Escalera, Seguí, Rodríguez-Palomares and Lekadir}]{campello_multi-centre_2021}
\bibinfo{author}{Campello, V.M.}, \bibinfo{author}{Gkontra, P.}, \bibinfo{author}{Izquierdo, C.}, \bibinfo{author}{Martín-Isla, C.}, \bibinfo{author}{Sojoudi, A.}, \bibinfo{author}{Full, P.M.}, \bibinfo{author}{Maier-Hein, K.}, \bibinfo{author}{Zhang, Y.}, \bibinfo{author}{He, Z.}, \bibinfo{author}{Ma, J.}, \bibinfo{author}{Parreño, M.}, \bibinfo{author}{Albiol, A.}, \bibinfo{author}{Kong, F.}, \bibinfo{author}{Shadden, S.C.}, \bibinfo{author}{Acero, J.C.}, \bibinfo{author}{Sundaresan, V.}, \bibinfo{author}{Saber, M.}, \bibinfo{author}{Elattar, M.}, \bibinfo{author}{Li, H.}, \bibinfo{author}{Menze, B.}, \bibinfo{author}{Khader, F.}, \bibinfo{author}{Haarburger, C.}, \bibinfo{author}{Scannell, C.M.}, \bibinfo{author}{Veta, M.}, \bibinfo{author}{Carscadden, A.}, \bibinfo{author}{Punithakumar, K.}, \bibinfo{author}{Liu, X.}, \bibinfo{author}{Tsaftaris, S.A.}, \bibinfo{author}{Huang, X.}, \bibinfo{author}{Yang, X.}, \bibinfo{author}{Li, L.}, \bibinfo{author}{Zhuang, X.}, \bibinfo{author}{Viladés, D.},
  \bibinfo{author}{Descalzo, M.L.}, \bibinfo{author}{Guala, A.}, \bibinfo{author}{Mura, L.L.}, \bibinfo{author}{Friedrich, M.G.}, \bibinfo{author}{Garg, R.}, \bibinfo{author}{Lebel, J.}, \bibinfo{author}{Henriques, F.}, \bibinfo{author}{Karakas, M.}, \bibinfo{author}{Çavuş, E.}, \bibinfo{author}{Petersen, S.E.}, \bibinfo{author}{Escalera, S.}, \bibinfo{author}{Seguí, S.}, \bibinfo{author}{Rodríguez-Palomares, J.F.}, \bibinfo{author}{Lekadir, K.}, \bibinfo{year}{2021}.
\newblock \bibinfo{title}{Multi-{Centre}, {Multi}-{Vendor} and {Multi}-{Disease} {Cardiac} {Segmentation}: {The} {M} amp;{Ms} {Challenge}}.
\newblock \bibinfo{journal}{IEEE Transactions on Medical Imaging} \bibinfo{volume}{40}, \bibinfo{pages}{3543--3554}.
\newblock \DOIprefix\doi{10.1109/TMI.2021.3090082}. \bibinfo{note}{conference Name: IEEE Transactions on Medical Imaging}.
\bibitem[{Cardoso et~al.(2022)Cardoso, Li, Brown, Ma, Kerfoot, Wang, Murrey, Myronenko, Zhao, Yang, Nath, He, Xu, Hatamizadeh, Myronenko, Zhu, Liu, Zheng, Tang, Yang, Zephyr, Hashemian, Alle, Darestani, Budd, Modat, Vercauteren, Wang, Li, Hu, Fu, Gorman, Johnson, Genereaux, Erdal, Gupta, Diaz-Pinto, Dourson, Maier-Hein, Jaeger, Baumgartner, Kalpathy-Cramer, Flores, Kirby, Cooper, Roth, Xu, Bericat, Floca, Zhou, Shuaib, Farahani, Maier-Hein, Aylward, Dogra, Ourselin and Feng}]{cardoso_monai_2022}
\bibinfo{author}{Cardoso, M.J.}, \bibinfo{author}{Li, W.}, \bibinfo{author}{Brown, R.}, \bibinfo{author}{Ma, N.}, \bibinfo{author}{Kerfoot, E.}, \bibinfo{author}{Wang, Y.}, \bibinfo{author}{Murrey, B.}, \bibinfo{author}{Myronenko, A.}, \bibinfo{author}{Zhao, C.}, \bibinfo{author}{Yang, D.}, \bibinfo{author}{Nath, V.}, \bibinfo{author}{He, Y.}, \bibinfo{author}{Xu, Z.}, \bibinfo{author}{Hatamizadeh, A.}, \bibinfo{author}{Myronenko, A.}, \bibinfo{author}{Zhu, W.}, \bibinfo{author}{Liu, Y.}, \bibinfo{author}{Zheng, M.}, \bibinfo{author}{Tang, Y.}, \bibinfo{author}{Yang, I.}, \bibinfo{author}{Zephyr, M.}, \bibinfo{author}{Hashemian, B.}, \bibinfo{author}{Alle, S.}, \bibinfo{author}{Darestani, M.Z.}, \bibinfo{author}{Budd, C.}, \bibinfo{author}{Modat, M.}, \bibinfo{author}{Vercauteren, T.}, \bibinfo{author}{Wang, G.}, \bibinfo{author}{Li, Y.}, \bibinfo{author}{Hu, Y.}, \bibinfo{author}{Fu, Y.}, \bibinfo{author}{Gorman, B.}, \bibinfo{author}{Johnson, H.}, \bibinfo{author}{Genereaux, B.}, \bibinfo{author}{Erdal,
  B.S.}, \bibinfo{author}{Gupta, V.}, \bibinfo{author}{Diaz-Pinto, A.}, \bibinfo{author}{Dourson, A.}, \bibinfo{author}{Maier-Hein, L.}, \bibinfo{author}{Jaeger, P.F.}, \bibinfo{author}{Baumgartner, M.}, \bibinfo{author}{Kalpathy-Cramer, J.}, \bibinfo{author}{Flores, M.}, \bibinfo{author}{Kirby, J.}, \bibinfo{author}{Cooper, L.A.D.}, \bibinfo{author}{Roth, H.R.}, \bibinfo{author}{Xu, D.}, \bibinfo{author}{Bericat, D.}, \bibinfo{author}{Floca, R.}, \bibinfo{author}{Zhou, S.K.}, \bibinfo{author}{Shuaib, H.}, \bibinfo{author}{Farahani, K.}, \bibinfo{author}{Maier-Hein, K.H.}, \bibinfo{author}{Aylward, S.}, \bibinfo{author}{Dogra, P.}, \bibinfo{author}{Ourselin, S.}, \bibinfo{author}{Feng, A.}, \bibinfo{year}{2022}.
\newblock \bibinfo{title}{{MONAI}: {An} open-source framework for deep learning in healthcare}.
\newblock \URLprefix \url{http://arxiv.org/abs/2211.02701}, \DOIprefix\doi{10.48550/arXiv.2211.02701}. \bibinfo{note}{arXiv:2211.02701 [cs]}.
\bibitem[{Chen et~al.(2020)Chen, Men, Chen, Tang, Zhang, Wang, Li and Dai}]{chen_cnn-based_2020}
\bibinfo{author}{Chen, X.}, \bibinfo{author}{Men, K.}, \bibinfo{author}{Chen, B.}, \bibinfo{author}{Tang, Y.}, \bibinfo{author}{Zhang, T.}, \bibinfo{author}{Wang, S.}, \bibinfo{author}{Li, Y.}, \bibinfo{author}{Dai, J.}, \bibinfo{year}{2020}.
\newblock \bibinfo{title}{{CNN}-{Based} {Quality} {Assurance} for {Automatic} {Segmentation} of {Breast} {Cancer} in {Radiotherapy}}.
\newblock \bibinfo{journal}{Frontiers in Oncology} \bibinfo{volume}{10}.
\newblock \URLprefix \url{https://www.frontiersin.org/article/10.3389/fonc.2020.00524}.
\bibitem[{Clark et~al.(2013)Clark, Vendt, Smith, Freymann, Kirby, Koppel, Moore, Phillips, Maffitt, Pringle, Tarbox and Prior}]{clark_cancer_2013}
\bibinfo{author}{Clark, K.}, \bibinfo{author}{Vendt, B.}, \bibinfo{author}{Smith, K.}, \bibinfo{author}{Freymann, J.}, \bibinfo{author}{Kirby, J.}, \bibinfo{author}{Koppel, P.}, \bibinfo{author}{Moore, S.}, \bibinfo{author}{Phillips, S.}, \bibinfo{author}{Maffitt, D.}, \bibinfo{author}{Pringle, M.}, \bibinfo{author}{Tarbox, L.}, \bibinfo{author}{Prior, F.}, \bibinfo{year}{2013}.
\newblock \bibinfo{title}{The {Cancer} {Imaging} {Archive} ({TCIA}): {Maintaining} and {Operating} a {Public} {Information} {Repository}}.
\newblock \bibinfo{journal}{Journal of Digital Imaging} \bibinfo{volume}{26}, \bibinfo{pages}{1045--1057}.
\newblock \URLprefix \url{https://doi.org/10.1007/s10278-013-9622-7}, \DOIprefix\doi{10.1007/s10278-013-9622-7}.
\bibitem[{Crum et~al.(2006)Crum, Camara and Hill}]{crum_generalized_2006}
\bibinfo{author}{Crum, W.}, \bibinfo{author}{Camara, O.}, \bibinfo{author}{Hill, D.}, \bibinfo{year}{2006}.
\newblock \bibinfo{title}{Generalized {Overlap} {Measures} for {Evaluation} and {Validation} in {Medical} {Image} {Analysis}}.
\newblock \bibinfo{journal}{IEEE Transactions on Medical Imaging} \bibinfo{volume}{25}, \bibinfo{pages}{1451--1461}.
\newblock \DOIprefix\doi{10.1109/TMI.2006.880587}. \bibinfo{note}{conference Name: IEEE Transactions on Medical Imaging}.
\bibitem[{DeVries and Taylor(2018)}]{devries_leveraging_2018}
\bibinfo{author}{DeVries, T.}, \bibinfo{author}{Taylor, G.W.}, \bibinfo{year}{2018}.
\newblock \bibinfo{title}{Leveraging {Uncertainty} {Estimates} for {Predicting} {Segmentation} {Quality}}.
\newblock \bibinfo{journal}{arXiv:1807.00502 [cs]} \URLprefix \url{http://arxiv.org/abs/1807.00502}. \bibinfo{note}{arXiv: 1807.00502}.
\bibitem[{El-Yaniv and Wiener(2010)}]{el-yaniv_foundations_2010}
\bibinfo{author}{El-Yaniv, R.}, \bibinfo{author}{Wiener, Y.}, \bibinfo{year}{2010}.
\newblock \bibinfo{title}{On the {Foundations} of {Noise}-free {Selective} {Classification}}.
\newblock \bibinfo{journal}{The Journal of Machine Learning Research} \bibinfo{volume}{11}, \bibinfo{pages}{1605--1641}.
\bibitem[{Esser et~al.(2021)Esser, Rombach and Ommer}]{esser_taming_2021}
\bibinfo{author}{Esser, P.}, \bibinfo{author}{Rombach, R.}, \bibinfo{author}{Ommer, B.}, \bibinfo{year}{2021}.
\newblock \bibinfo{title}{Taming {Transformers} for {High}-{Resolution} {Image} {Synthesis}}.
\newblock \URLprefix \url{http://arxiv.org/abs/2012.09841}, \DOIprefix\doi{10.48550/arXiv.2012.09841}. \bibinfo{note}{arXiv:2012.09841 [cs]}.
\bibitem[{Full et~al.(2020)Full, Isensee, Jäger and Maier-Hein}]{full_studying_2020}
\bibinfo{author}{Full, P.M.}, \bibinfo{author}{Isensee, F.}, \bibinfo{author}{Jäger, P.F.}, \bibinfo{author}{Maier-Hein, K.}, \bibinfo{year}{2020}.
\newblock \bibinfo{title}{Studying {Robustness} of {Semantic} {Segmentation} under {Domain} {Shift} in cardiac {MRI}}.
\newblock \bibinfo{journal}{arXiv:2011.07592 [cs, eess]} \URLprefix \url{http://arxiv.org/abs/2011.07592}. \bibinfo{note}{arXiv: 2011.07592}.
\bibitem[{Gal and Ghahramani(2016)}]{gal_dropout_2016}
\bibinfo{author}{Gal, Y.}, \bibinfo{author}{Ghahramani, Z.}, \bibinfo{year}{2016}.
\newblock \bibinfo{title}{Dropout as a {Bayesian} {Approximation}: {Representing} {Model} {Uncertainty} in {Deep} {Learning}} , \bibinfo{pages}{10}.
\bibitem[{Galil et~al.(2023)Galil, Dabbah and El-Yaniv}]{galil_what_2023}
\bibinfo{author}{Galil, I.}, \bibinfo{author}{Dabbah, M.}, \bibinfo{author}{El-Yaniv, R.}, \bibinfo{year}{2023}.
\newblock \bibinfo{title}{What {Can} {We} {Learn} {From} {The} {Selective} {Prediction} {And} {Uncertainty} {Estimation} {Performance} {Of} 523 {Imagenet} {Classifiers}}.
\newblock \URLprefix \url{http://arxiv.org/abs/2302.11874}, \DOIprefix\doi{10.48550/arXiv.2302.11874}. \bibinfo{note}{arXiv:2302.11874 [cs]}.
\bibitem[{Geifman and El-Yaniv(2017)}]{geifman_selective_2017}
\bibinfo{author}{Geifman, Y.}, \bibinfo{author}{El-Yaniv, R.}, \bibinfo{year}{2017}.
\newblock \bibinfo{title}{Selective {Classification} for {Deep} {Neural} {Networks}} \URLprefix \url{https://arxiv.org/abs/1705.08500v2}.
\bibitem[{González et~al.(2022)González, Gotkowski, Fuchs, Bucher, Dadras, Fischbach, Kaltenborn and Mukhopadhyay}]{gonzalez_distance-based_2022}
\bibinfo{author}{González, C.}, \bibinfo{author}{Gotkowski, K.}, \bibinfo{author}{Fuchs, M.}, \bibinfo{author}{Bucher, A.}, \bibinfo{author}{Dadras, A.}, \bibinfo{author}{Fischbach, R.}, \bibinfo{author}{Kaltenborn, I.J.}, \bibinfo{author}{Mukhopadhyay, A.}, \bibinfo{year}{2022}.
\newblock \bibinfo{title}{Distance-based detection of out-of-distribution silent failures for {Covid}-19 lung lesion segmentation}.
\newblock \bibinfo{journal}{Medical Image Analysis} \bibinfo{volume}{82}, \bibinfo{pages}{102596}.
\newblock \URLprefix \url{https://linkinghub.elsevier.com/retrieve/pii/S1361841522002298}, \DOIprefix\doi{10.1016/j.media.2022.102596}.
\bibitem[{Graham et~al.(2022)Graham, Tudosiu, Wright, Pinaya, U-King-Im, Mah, Teo, Jäger, Werring, Nachev, Ourselin and Cardoso}]{graham_transformer-based_2022}
\bibinfo{author}{Graham, M.S.}, \bibinfo{author}{Tudosiu, P.D.}, \bibinfo{author}{Wright, P.}, \bibinfo{author}{Pinaya, W.H.L.}, \bibinfo{author}{U-King-Im, J.M.}, \bibinfo{author}{Mah, Y.}, \bibinfo{author}{Teo, J.}, \bibinfo{author}{Jäger, R.H.}, \bibinfo{author}{Werring, D.}, \bibinfo{author}{Nachev, P.}, \bibinfo{author}{Ourselin, S.}, \bibinfo{author}{Cardoso, M.J.}, \bibinfo{year}{2022}.
\newblock \bibinfo{title}{Transformer-based out-of-distribution detection for clinically safe segmentation}.
\newblock \URLprefix \url{https://openreview.net/forum?id=En7660i-CLJ}.
\bibitem[{van Griethuysen et~al.(2017)van Griethuysen, Fedorov, Parmar, Hosny, Aucoin, Narayan, Beets-Tan, Fillion-Robin, Pieper and Aerts}]{van_griethuysen_computational_2017}
\bibinfo{author}{van Griethuysen, J.J.M.}, \bibinfo{author}{Fedorov, A.}, \bibinfo{author}{Parmar, C.}, \bibinfo{author}{Hosny, A.}, \bibinfo{author}{Aucoin, N.}, \bibinfo{author}{Narayan, V.}, \bibinfo{author}{Beets-Tan, R.G.H.}, \bibinfo{author}{Fillion-Robin, J.C.}, \bibinfo{author}{Pieper, S.}, \bibinfo{author}{Aerts, H.J.W.L.}, \bibinfo{year}{2017}.
\newblock \bibinfo{title}{Computational {Radiomics} {System} to {Decode} the {Radiographic} {Phenotype}}.
\newblock \bibinfo{journal}{Cancer Research} \bibinfo{volume}{77}, \bibinfo{pages}{e104--e107}.
\newblock \DOIprefix\doi{10.1158/0008-5472.CAN-17-0339}.
\bibitem[{Heller et~al.(2021)Heller, Isensee, Maier-Hein, Hou, Xie, Li, Nan, Mu, Lin, Han, Yao, Gao, Zhang, Wang, Hou, Yang, Xiong, Tian, Zhong, Ma, Rickman, Dean, Stai, Tejpaul, Oestreich, Blake, Kaluzniak, Raza, Rosenberg, Moore, Walczak, Rengel, Edgerton, Vasdev, Peterson, McSweeney, Peterson, Kalapara, Sathianathen, Papanikolopoulos and Weight}]{heller_state_2021}
\bibinfo{author}{Heller, N.}, \bibinfo{author}{Isensee, F.}, \bibinfo{author}{Maier-Hein, K.H.}, \bibinfo{author}{Hou, X.}, \bibinfo{author}{Xie, C.}, \bibinfo{author}{Li, F.}, \bibinfo{author}{Nan, Y.}, \bibinfo{author}{Mu, G.}, \bibinfo{author}{Lin, Z.}, \bibinfo{author}{Han, M.}, \bibinfo{author}{Yao, G.}, \bibinfo{author}{Gao, Y.}, \bibinfo{author}{Zhang, Y.}, \bibinfo{author}{Wang, Y.}, \bibinfo{author}{Hou, F.}, \bibinfo{author}{Yang, J.}, \bibinfo{author}{Xiong, G.}, \bibinfo{author}{Tian, J.}, \bibinfo{author}{Zhong, C.}, \bibinfo{author}{Ma, J.}, \bibinfo{author}{Rickman, J.}, \bibinfo{author}{Dean, J.}, \bibinfo{author}{Stai, B.}, \bibinfo{author}{Tejpaul, R.}, \bibinfo{author}{Oestreich, M.}, \bibinfo{author}{Blake, P.}, \bibinfo{author}{Kaluzniak, H.}, \bibinfo{author}{Raza, S.}, \bibinfo{author}{Rosenberg, J.}, \bibinfo{author}{Moore, K.}, \bibinfo{author}{Walczak, E.}, \bibinfo{author}{Rengel, Z.}, \bibinfo{author}{Edgerton, Z.}, \bibinfo{author}{Vasdev, R.}, \bibinfo{author}{Peterson, M.},
  \bibinfo{author}{McSweeney, S.}, \bibinfo{author}{Peterson, S.}, \bibinfo{author}{Kalapara, A.}, \bibinfo{author}{Sathianathen, N.}, \bibinfo{author}{Papanikolopoulos, N.}, \bibinfo{author}{Weight, C.}, \bibinfo{year}{2021}.
\newblock \bibinfo{title}{The state of the art in kidney and kidney tumor segmentation in contrast-enhanced {CT} imaging: {Results} of the {KiTS19} challenge}.
\newblock \bibinfo{journal}{Medical Image Analysis} \bibinfo{volume}{67}, \bibinfo{pages}{101821}.
\newblock \URLprefix \url{https://www.sciencedirect.com/science/article/pii/S1361841520301857}, \DOIprefix\doi{10.1016/j.media.2020.101821}.
\bibitem[{Heller et~al.(2023)Heller, Isensee, Trofimova, Tejpaul, Zhao, Chen, Wang, Golts, Khapun, Shats, Shoshan, Gilboa-Solomon, George, Yang, Zhang, Zhang, Xia, Wu, Liu, Walczak, McSweeney, Vasdev, Hornung, Solaiman, Schoephoerster, Abernathy, Wu, Abdulkadir, Byun, Spriggs, Struyk, Austin, Simpson, Hagstrom, Virnig, French, Venkatesh, Chan, Moore, Jacobsen, Austin, Austin, Regmi, Papanikolopoulos and Weight}]{heller_kits21_2023}
\bibinfo{author}{Heller, N.}, \bibinfo{author}{Isensee, F.}, \bibinfo{author}{Trofimova, D.}, \bibinfo{author}{Tejpaul, R.}, \bibinfo{author}{Zhao, Z.}, \bibinfo{author}{Chen, H.}, \bibinfo{author}{Wang, L.}, \bibinfo{author}{Golts, A.}, \bibinfo{author}{Khapun, D.}, \bibinfo{author}{Shats, D.}, \bibinfo{author}{Shoshan, Y.}, \bibinfo{author}{Gilboa-Solomon, F.}, \bibinfo{author}{George, Y.}, \bibinfo{author}{Yang, X.}, \bibinfo{author}{Zhang, J.}, \bibinfo{author}{Zhang, J.}, \bibinfo{author}{Xia, Y.}, \bibinfo{author}{Wu, M.}, \bibinfo{author}{Liu, Z.}, \bibinfo{author}{Walczak, E.}, \bibinfo{author}{McSweeney, S.}, \bibinfo{author}{Vasdev, R.}, \bibinfo{author}{Hornung, C.}, \bibinfo{author}{Solaiman, R.}, \bibinfo{author}{Schoephoerster, J.}, \bibinfo{author}{Abernathy, B.}, \bibinfo{author}{Wu, D.}, \bibinfo{author}{Abdulkadir, S.}, \bibinfo{author}{Byun, B.}, \bibinfo{author}{Spriggs, J.}, \bibinfo{author}{Struyk, G.}, \bibinfo{author}{Austin, A.}, \bibinfo{author}{Simpson, B.},
  \bibinfo{author}{Hagstrom, M.}, \bibinfo{author}{Virnig, S.}, \bibinfo{author}{French, J.}, \bibinfo{author}{Venkatesh, N.}, \bibinfo{author}{Chan, S.}, \bibinfo{author}{Moore, K.}, \bibinfo{author}{Jacobsen, A.}, \bibinfo{author}{Austin, S.}, \bibinfo{author}{Austin, M.}, \bibinfo{author}{Regmi, S.}, \bibinfo{author}{Papanikolopoulos, N.}, \bibinfo{author}{Weight, C.}, \bibinfo{year}{2023}.
\newblock \bibinfo{title}{The {KiTS21} {Challenge}: {Automatic} segmentation of kidneys, renal tumors, and renal cysts in corticomedullary-phase {CT}}.
\newblock \URLprefix \url{http://arxiv.org/abs/2307.01984}, \DOIprefix\doi{10.48550/arXiv.2307.01984}. \bibinfo{note}{arXiv:2307.01984 [cs]}.
\bibitem[{Hoebel et~al.(2020)Hoebel, Andrearczyk, Beers, Patel, Chang, Depeursinge, Müller and Kalpathy-Cramer}]{hoebel_exploration_2020}
\bibinfo{author}{Hoebel, K.}, \bibinfo{author}{Andrearczyk, V.}, \bibinfo{author}{Beers, A.}, \bibinfo{author}{Patel, J.}, \bibinfo{author}{Chang, K.}, \bibinfo{author}{Depeursinge, A.}, \bibinfo{author}{Müller, H.}, \bibinfo{author}{Kalpathy-Cramer, J.}, \bibinfo{year}{2020}.
\newblock \bibinfo{title}{An exploration of uncertainty information for segmentation quality assessment}, in: \bibinfo{booktitle}{Medical {Imaging} 2020: {Image} {Processing}}, \bibinfo{publisher}{SPIE}. pp. \bibinfo{pages}{381--390}.
\newblock \URLprefix \url{https://www.spiedigitallibrary.org/conference-proceedings-of-spie/11313/113131K/An-exploration-of-uncertainty-information-for-segmentation-quality-assessment/10.1117/12.2548722.full}, \DOIprefix\doi{10.1117/12.2548722}.
\bibitem[{Hoebel et~al.(2022)Hoebel, Bridge, Lemay, Chang, Patel, M.d and Kalpathy-Cramer}]{hoebel_i_2022}
\bibinfo{author}{Hoebel, K.}, \bibinfo{author}{Bridge, C.}, \bibinfo{author}{Lemay, A.}, \bibinfo{author}{Chang, K.}, \bibinfo{author}{Patel, J.}, \bibinfo{author}{M.d, B.R.}, \bibinfo{author}{Kalpathy-Cramer, J.}, \bibinfo{year}{2022}.
\newblock \bibinfo{title}{Do {I} know this? segmentation uncertainty under domain shift}, in: \bibinfo{booktitle}{Medical {Imaging} 2022: {Image} {Processing}}, \bibinfo{publisher}{SPIE}. pp. \bibinfo{pages}{261--276}.
\newblock \URLprefix \url{https://www.spiedigitallibrary.org/conference-proceedings-of-spie/12032/1203211/Do-I-know-this-segmentation-uncertainty-under-domain-shift/10.1117/12.2611867.full}, \DOIprefix\doi{10.1117/12.2611867}.
\bibitem[{Isensee et~al.(2021)Isensee, Jaeger, Kohl, Petersen and Maier-Hein}]{isensee_nnu-net_2021}
\bibinfo{author}{Isensee, F.}, \bibinfo{author}{Jaeger, P.F.}, \bibinfo{author}{Kohl, S.A.A.}, \bibinfo{author}{Petersen, J.}, \bibinfo{author}{Maier-Hein, K.H.}, \bibinfo{year}{2021}.
\newblock \bibinfo{title}{{nnU}-{Net}: a self-configuring method for deep learning-based biomedical image segmentation}.
\newblock \bibinfo{journal}{Nature Methods} \bibinfo{volume}{18}, \bibinfo{pages}{203--211}.
\newblock \URLprefix \url{https://www.nature.com/articles/s41592-020-01008-z}, \DOIprefix\doi{10.1038/s41592-020-01008-z}. \bibinfo{note}{number: 2 Publisher: Nature Publishing Group}.
\bibitem[{Jaeger et~al.(2022)Jaeger, Lüth, Klein and Bungert}]{jaeger_call_2022}
\bibinfo{author}{Jaeger, P.F.}, \bibinfo{author}{Lüth, C.T.}, \bibinfo{author}{Klein, L.}, \bibinfo{author}{Bungert, T.J.}, \bibinfo{year}{2022}.
\newblock \bibinfo{title}{A {Call} to {Reflect} on {Evaluation} {Practices} for {Failure} {Detection} in {Image} {Classification}}.
\newblock \URLprefix \url{https://openreview.net/forum?id=YnkGMIh0gvX}.
\bibitem[{Jun et~al.(2020)Jun, Cheng, Yixin, Xingle, Jiantao, Ziqi, Minqing, Xin, Xueyuan, Shucheng, Hao, Sen, Xiaoyu, Ziwei, Chen, Lu, Yuntao, Qiongjie, Guoqiang and Jian}]{jun_covid-19_2020}
\bibinfo{author}{Jun, M.}, \bibinfo{author}{Cheng, G.}, \bibinfo{author}{Yixin, W.}, \bibinfo{author}{Xingle, A.}, \bibinfo{author}{Jiantao, G.}, \bibinfo{author}{Ziqi, Y.}, \bibinfo{author}{Minqing, Z.}, \bibinfo{author}{Xin, L.}, \bibinfo{author}{Xueyuan, D.}, \bibinfo{author}{Shucheng, C.}, \bibinfo{author}{Hao, W.}, \bibinfo{author}{Sen, M.}, \bibinfo{author}{Xiaoyu, Y.}, \bibinfo{author}{Ziwei, N.}, \bibinfo{author}{Chen, L.}, \bibinfo{author}{Lu, T.}, \bibinfo{author}{Yuntao, Z.}, \bibinfo{author}{Qiongjie, Z.}, \bibinfo{author}{Guoqiang, D.}, \bibinfo{author}{Jian, H.}, \bibinfo{year}{2020}.
\newblock \bibinfo{title}{{COVID}-19 {CT} {Lung} and {Infection} {Segmentation} {Dataset}}.
\newblock \URLprefix \url{https://zenodo.org/records/3757476}, \DOIprefix\doi{10.5281/zenodo.3757476}.
\bibitem[{Jungo et~al.(2020)Jungo, Balsiger and Reyes}]{jungo_analyzing_2020}
\bibinfo{author}{Jungo, A.}, \bibinfo{author}{Balsiger, F.}, \bibinfo{author}{Reyes, M.}, \bibinfo{year}{2020}.
\newblock \bibinfo{title}{Analyzing the {Quality} and {Challenges} of {Uncertainty} {Estimations} for {Brain} {Tumor} {Segmentation}}.
\newblock \bibinfo{journal}{Frontiers in Neuroscience} \bibinfo{volume}{14}.
\newblock \URLprefix \url{https://www.readcube.com/articles/10.3389%2Ffnins.2020.00282}, \DOIprefix\doi{10.3389/fnins.2020.00282}.
\bibitem[{Kahl et~al.(2024)Kahl, Lüth, Zenk, Maier-Hein and Jaeger}]{kahl_values_2024}
\bibinfo{author}{Kahl, K.C.}, \bibinfo{author}{Lüth, C.T.}, \bibinfo{author}{Zenk, M.}, \bibinfo{author}{Maier-Hein, K.}, \bibinfo{author}{Jaeger, P.F.}, \bibinfo{year}{2024}.
\newblock \bibinfo{title}{{ValUES}: {A} {Framework} for {Systematic} {Validation} of {Uncertainty} {Estimation} in {Semantic} {Segmentation}}.
\newblock \URLprefix \url{http://arxiv.org/abs/2401.08501}, \DOIprefix\doi{10.48550/arXiv.2401.08501}. \bibinfo{note}{arXiv:2401.08501 [cs]}.
\bibitem[{Kendall et~al.(2016)Kendall, Badrinarayanan and Cipolla}]{kendall_bayesian_2016}
\bibinfo{author}{Kendall, A.}, \bibinfo{author}{Badrinarayanan, V.}, \bibinfo{author}{Cipolla, R.}, \bibinfo{year}{2016}.
\newblock \bibinfo{title}{Bayesian {SegNet}: {Model} {Uncertainty} in {Deep} {Convolutional} {Encoder}-{Decoder} {Architectures} for {Scene} {Understanding}}.
\newblock \URLprefix \url{http://arxiv.org/abs/1511.02680}, \DOIprefix\doi{10.48550/arXiv.1511.02680}. \bibinfo{note}{arXiv:1511.02680 [cs]}.
\bibitem[{Kingma and Ba(2017)}]{kingma_adam_2017}
\bibinfo{author}{Kingma, D.P.}, \bibinfo{author}{Ba, J.}, \bibinfo{year}{2017}.
\newblock \bibinfo{title}{Adam: {A} {Method} for {Stochastic} {Optimization}}.
\newblock \URLprefix \url{http://arxiv.org/abs/1412.6980}, \DOIprefix\doi{10.48550/arXiv.1412.6980}. \bibinfo{note}{arXiv:1412.6980 [cs]}.
\bibitem[{Kingma and Welling(2013)}]{kingma_auto-encoding_2013}
\bibinfo{author}{Kingma, D.P.}, \bibinfo{author}{Welling, M.}, \bibinfo{year}{2013}.
\newblock \bibinfo{title}{Auto-{Encoding} {Variational} {Bayes}}.
\newblock \bibinfo{journal}{arXiv:1312.6114 [cs, stat]} \URLprefix \url{http://arxiv.org/abs/1312.6114}. \bibinfo{note}{arXiv: 1312.6114}.
\bibitem[{Kohl et~al.(2018)Kohl, Romera-Paredes, Meyer, De~Fauw, Ledsam, Maier-Hein, Eslami, Rezende and Ronneberger}]{kohl_probabilistic_2018}
\bibinfo{author}{Kohl, S.A.A.}, \bibinfo{author}{Romera-Paredes, B.}, \bibinfo{author}{Meyer, C.}, \bibinfo{author}{De~Fauw, J.}, \bibinfo{author}{Ledsam, J.R.}, \bibinfo{author}{Maier-Hein, K.H.}, \bibinfo{author}{Eslami, S.M.A.}, \bibinfo{author}{Rezende, D.J.}, \bibinfo{author}{Ronneberger, O.}, \bibinfo{year}{2018}.
\newblock \bibinfo{title}{A {Probabilistic} {U}-{Net} for {Segmentation} of {Ambiguous} {Images}}.
\newblock \bibinfo{journal}{arXiv:1806.05034 [cs, stat]} \URLprefix \url{http://arxiv.org/abs/1806.05034}. \bibinfo{note}{arXiv: 1806.05034}.
\bibitem[{Kohlberger et~al.(2012)Kohlberger, Singh, Alvino, Bahlmann and Grady}]{kohlberger_evaluating_2012}
\bibinfo{author}{Kohlberger, T.}, \bibinfo{author}{Singh, V.}, \bibinfo{author}{Alvino, C.}, \bibinfo{author}{Bahlmann, C.}, \bibinfo{author}{Grady, L.}, \bibinfo{year}{2012}.
\newblock \bibinfo{title}{Evaluating {Segmentation} {Error} without {Ground} {Truth}}, in: \bibinfo{editor}{Ayache, N.}, \bibinfo{editor}{Delingette, H.}, \bibinfo{editor}{Golland, P.}, \bibinfo{editor}{Mori, K.} (Eds.), \bibinfo{booktitle}{Medical {Image} {Computing} and {Computer}-{Assisted} {Intervention} – {MICCAI} 2012}, \bibinfo{publisher}{Springer}, \bibinfo{address}{Berlin, Heidelberg}. pp. \bibinfo{pages}{528--536}.
\newblock \DOIprefix\doi{10.1007/978-3-642-33415-3_65}.
\bibitem[{Kushibar et~al.(2022)Kushibar, Campello, Garrucho, Linardos, Radeva and Lekadir}]{wang_layer_2022}
\bibinfo{author}{Kushibar, K.}, \bibinfo{author}{Campello, V.}, \bibinfo{author}{Garrucho, L.}, \bibinfo{author}{Linardos, A.}, \bibinfo{author}{Radeva, P.}, \bibinfo{author}{Lekadir, K.}, \bibinfo{year}{2022}.
\newblock \bibinfo{title}{Layer {Ensembles}: {A} {Single}-{Pass} {Uncertainty} {Estimation} in {Deep} {Learning} for {Segmentation}}, in: \bibinfo{editor}{Wang, L.}, \bibinfo{editor}{Dou, Q.}, \bibinfo{editor}{Fletcher, P.T.}, \bibinfo{editor}{Speidel, S.}, \bibinfo{editor}{Li, S.} (Eds.), \bibinfo{booktitle}{Medical {Image} {Computing} and {Computer} {Assisted} {Intervention} – {MICCAI} 2022}. \bibinfo{publisher}{Springer Nature Switzerland}, \bibinfo{address}{Cham}. volume \bibinfo{volume}{13438}, pp. \bibinfo{pages}{514--524}.
\newblock \URLprefix \url{https://link.springer.com/10.1007/978-3-031-16452-1_49}, \DOIprefix\doi{10.1007/978-3-031-16452-1_49}. \bibinfo{note}{series Title: Lecture Notes in Computer Science}.
\bibitem[{Kwon et~al.(2020)Kwon, Won, Kim and Paik}]{kwon_uncertainty_2020}
\bibinfo{author}{Kwon, Y.}, \bibinfo{author}{Won, J.H.}, \bibinfo{author}{Kim, B.J.}, \bibinfo{author}{Paik, M.C.}, \bibinfo{year}{2020}.
\newblock \bibinfo{title}{Uncertainty quantification using {Bayesian} neural networks in classification: {Application} to biomedical image segmentation}.
\newblock \bibinfo{journal}{Computational Statistics \& Data Analysis} \bibinfo{volume}{142}, \bibinfo{pages}{106816}.
\newblock \URLprefix \url{https://www.sciencedirect.com/science/article/pii/S016794731930163X}, \DOIprefix\doi{10.1016/j.csda.2019.106816}.
\bibitem[{Lakshminarayanan et~al.(2017)Lakshminarayanan, Pritzel and Blundell}]{lakshminarayanan_simple_2017}
\bibinfo{author}{Lakshminarayanan, B.}, \bibinfo{author}{Pritzel, A.}, \bibinfo{author}{Blundell, C.}, \bibinfo{year}{2017}.
\newblock \bibinfo{title}{Simple and {Scalable} {Predictive} {Uncertainty} {Estimation} using {Deep} {Ensembles}}, in: \bibinfo{booktitle}{Advances in {Neural} {Information} {Processing} {Systems}}, \bibinfo{publisher}{Curran Associates, Inc.}
\newblock \URLprefix \url{https://proceedings.neurips.cc/paper_files/paper/2017/hash/9ef2ed4b7fd2c810847ffa5fa85bce38-Abstract.html}.
\bibitem[{Lemaître et~al.(2015)Lemaître, Martí, Freixenet, Vilanova, Walker and Meriaudeau}]{lemaitre_computer-aided_2015}
\bibinfo{author}{Lemaître, G.}, \bibinfo{author}{Martí, R.}, \bibinfo{author}{Freixenet, J.}, \bibinfo{author}{Vilanova, J.C.}, \bibinfo{author}{Walker, P.M.}, \bibinfo{author}{Meriaudeau, F.}, \bibinfo{year}{2015}.
\newblock \bibinfo{title}{Computer-{Aided} {Detection} and diagnosis for prostate cancer based on mono and multi-parametric {MRI}: {A} review}.
\newblock \bibinfo{journal}{Computers in Biology and Medicine} \bibinfo{volume}{60}, \bibinfo{pages}{8--31}.
\newblock \URLprefix \url{https://www.sciencedirect.com/science/article/pii/S001048251500058X}, \DOIprefix\doi{10.1016/j.compbiomed.2015.02.009}.
\bibitem[{Lennartz and Schultz(2023)}]{greenspan_segmentation_2023}
\bibinfo{author}{Lennartz, J.}, \bibinfo{author}{Schultz, T.}, \bibinfo{year}{2023}.
\newblock \bibinfo{title}{Segmentation {Distortion}: {Quantifying} {Segmentation} {Uncertainty} {Under} {Domain} {Shift} via the {Effects} of {Anomalous} {Activations}}, in: \bibinfo{editor}{Greenspan, H.}, \bibinfo{editor}{Madabhushi, A.}, \bibinfo{editor}{Mousavi, P.}, \bibinfo{editor}{Salcudean, S.}, \bibinfo{editor}{Duncan, J.}, \bibinfo{editor}{Syeda-Mahmood, T.}, \bibinfo{editor}{Taylor, R.} (Eds.), \bibinfo{booktitle}{Medical {Image} {Computing} and {Computer} {Assisted} {Intervention} – {MICCAI} 2023}. \bibinfo{publisher}{Springer Nature Switzerland}, \bibinfo{address}{Cham}. volume \bibinfo{volume}{14222}, pp. \bibinfo{pages}{316--325}.
\newblock \URLprefix \url{https://link.springer.com/10.1007/978-3-031-43898-1_31}, \DOIprefix\doi{10.1007/978-3-031-43898-1_31}. \bibinfo{note}{series Title: Lecture Notes in Computer Science}.
\bibitem[{Li et~al.(2022)Li, Yu and Heng}]{li_towards_2022}
\bibinfo{author}{Li, K.}, \bibinfo{author}{Yu, L.}, \bibinfo{author}{Heng, P.A.}, \bibinfo{year}{2022}.
\newblock \bibinfo{title}{Towards reliable cardiac image segmentation: {Assessing} image-level and pixel-level segmentation quality via self-reflective references}.
\newblock \bibinfo{journal}{Medical Image Analysis} \bibinfo{volume}{78}, \bibinfo{pages}{102426}.
\newblock \URLprefix \url{https://www.sciencedirect.com/science/article/pii/S1361841522000779}, \DOIprefix\doi{10.1016/j.media.2022.102426}.
\bibitem[{Lin et~al.(2022)Lin, Chen, Chen and Garibaldi}]{lin_novel_2022}
\bibinfo{author}{Lin, Q.}, \bibinfo{author}{Chen, X.}, \bibinfo{author}{Chen, C.}, \bibinfo{author}{Garibaldi, J.M.}, \bibinfo{year}{2022}.
\newblock \bibinfo{title}{A {Novel} {Quality} {Control} {Algorithm} for {Medical} {Image} {Segmentation} {Based} on {Fuzzy} {Uncertainty}}.
\newblock \bibinfo{journal}{IEEE Transactions on Fuzzy Systems} , \bibinfo{pages}{1--14}\DOIprefix\doi{10.1109/TFUZZ.2022.3228332}. \bibinfo{note}{conference Name: IEEE Transactions on Fuzzy Systems}.
\bibitem[{Litjens et~al.(2023)Litjens, Ginneken, Huisman, Ven, Hoeks, Barratt and Madabhushi}]{litjens_promise12_2023}
\bibinfo{author}{Litjens, G.}, \bibinfo{author}{Ginneken, B.v.}, \bibinfo{author}{Huisman, H.}, \bibinfo{author}{Ven, W.v.d.}, \bibinfo{author}{Hoeks, C.}, \bibinfo{author}{Barratt, D.}, \bibinfo{author}{Madabhushi, A.}, \bibinfo{year}{2023}.
\newblock \bibinfo{title}{{PROMISE12}: {Data} from the {MICCAI} {Grand} {Challenge}: {Prostate} {MR} {Image} {Segmentation} 2012}.
\newblock \URLprefix \url{https://zenodo.org/records/8026660}, \DOIprefix\doi{10.5281/zenodo.8026660}.
\bibitem[{Liu et~al.(2019)Liu, Xia, Yang, Yuille and Xu}]{liu_alarm_2019}
\bibinfo{author}{Liu, F.}, \bibinfo{author}{Xia, Y.}, \bibinfo{author}{Yang, D.}, \bibinfo{author}{Yuille, A.L.}, \bibinfo{author}{Xu, D.}, \bibinfo{year}{2019}.
\newblock \bibinfo{title}{An {Alarm} {System} for {Segmentation} {Algorithm} {Based} on {Shape} {Model}}, pp. \bibinfo{pages}{10652--10661}.
\newblock \URLprefix \url{https://openaccess.thecvf.com/content_ICCV_2019/html/Liu_An_Alarm_System_for_Segmentation_Algorithm_Based_on_Shape_Model_ICCV_2019_paper.html}.
\bibitem[{Liu et~al.(2020)Liu, Dou and Heng}]{liu_shape-aware_2020}
\bibinfo{author}{Liu, Q.}, \bibinfo{author}{Dou, Q.}, \bibinfo{author}{Heng, P.A.}, \bibinfo{year}{2020}.
\newblock \bibinfo{title}{Shape-{Aware} {Meta}-learning for {Generalizing} {Prostate} {MRI} {Segmentation} to {Unseen} {Domains}}, in: \bibinfo{editor}{Martel, A.L.}, \bibinfo{editor}{Abolmaesumi, P.}, \bibinfo{editor}{Stoyanov, D.}, \bibinfo{editor}{Mateus, D.}, \bibinfo{editor}{Zuluaga, M.A.}, \bibinfo{editor}{Zhou, S.K.}, \bibinfo{editor}{Racoceanu, D.}, \bibinfo{editor}{Joskowicz, L.} (Eds.), \bibinfo{booktitle}{Medical {Image} {Computing} and {Computer} {Assisted} {Intervention} – {MICCAI} 2020}, \bibinfo{publisher}{Springer International Publishing}, \bibinfo{address}{Cham}. pp. \bibinfo{pages}{475--485}.
\newblock \DOIprefix\doi{10.1007/978-3-030-59713-9_46}.
\bibitem[{Loshchilov and Hutter(2019)}]{loshchilov_decoupled_2019}
\bibinfo{author}{Loshchilov, I.}, \bibinfo{author}{Hutter, F.}, \bibinfo{year}{2019}.
\newblock \bibinfo{title}{Decoupled {Weight} {Decay} {Regularization}}.
\newblock \URLprefix \url{http://arxiv.org/abs/1711.05101}, \DOIprefix\doi{10.48550/arXiv.1711.05101}. \bibinfo{note}{arXiv:1711.05101 [cs, math]}.
\bibitem[{Malinin et~al.(2022)Malinin, Band, Ganshin, Alexander, Chesnokov, Gal, Gales, Noskov, Ploskonosov, Prokhorenkova, Provilkov, Raina, Raina, Roginskiy, Denis, Shmatova, Tigas and Yangel}]{malinin_shifts_2022}
\bibinfo{author}{Malinin, A.}, \bibinfo{author}{Band, N.}, \bibinfo{author}{Ganshin}, \bibinfo{author}{Alexander}, \bibinfo{author}{Chesnokov, G.}, \bibinfo{author}{Gal, Y.}, \bibinfo{author}{Gales, M.J.F.}, \bibinfo{author}{Noskov, A.}, \bibinfo{author}{Ploskonosov, A.}, \bibinfo{author}{Prokhorenkova, L.}, \bibinfo{author}{Provilkov, I.}, \bibinfo{author}{Raina, V.}, \bibinfo{author}{Raina, V.}, \bibinfo{author}{Roginskiy}, \bibinfo{author}{Denis}, \bibinfo{author}{Shmatova, M.}, \bibinfo{author}{Tigas, P.}, \bibinfo{author}{Yangel, B.}, \bibinfo{year}{2022}.
\newblock \bibinfo{title}{Shifts: {A} {Dataset} of {Real} {Distributional} {Shift} {Across} {Multiple} {Large}-{Scale} {Tasks}}.
\newblock \URLprefix \url{http://arxiv.org/abs/2107.07455}, \DOIprefix\doi{10.48550/arXiv.2107.07455}. \bibinfo{note}{arXiv:2107.07455 [cs, stat]}.
\bibitem[{Mehrtash et~al.(2020)Mehrtash, Wells, Tempany, Abolmaesumi and Kapur}]{mehrtash_confidence_2020}
\bibinfo{author}{Mehrtash, A.}, \bibinfo{author}{Wells, W.M.}, \bibinfo{author}{Tempany, C.M.}, \bibinfo{author}{Abolmaesumi, P.}, \bibinfo{author}{Kapur, T.}, \bibinfo{year}{2020}.
\newblock \bibinfo{title}{Confidence {Calibration} and {Predictive} {Uncertainty} {Estimation} for {Deep} {Medical} {Image} {Segmentation}}.
\newblock \bibinfo{journal}{IEEE Transactions on Medical Imaging} \bibinfo{volume}{39}, \bibinfo{pages}{3868--3878}.
\newblock \DOIprefix\doi{10.1109/TMI.2020.3006437}. \bibinfo{note}{conference Name: IEEE Transactions on Medical Imaging}.
\bibitem[{Mehta et~al.(2020)Mehta, Filos, Gal and Arbel}]{mehta_uncertainty_2020}
\bibinfo{author}{Mehta, R.}, \bibinfo{author}{Filos, A.}, \bibinfo{author}{Gal, Y.}, \bibinfo{author}{Arbel, T.}, \bibinfo{year}{2020}.
\newblock \bibinfo{title}{Uncertainty {Evaluation} {Metric} for {Brain} {Tumour} {Segmentation}}.
\newblock \bibinfo{journal}{arXiv:2005.14262 [cs, eess]} \URLprefix \url{http://arxiv.org/abs/2005.14262}. \bibinfo{note}{arXiv: 2005.14262}.
\bibitem[{Menze et~al.(2015)Menze, Jakab, Bauer, Kalpathy-Cramer, Farahani, Kirby, Burren, Porz, Slotboom, Wiest, Lanczi, Gerstner, Weber, Arbel, Avants, Ayache, Buendia, Collins, Cordier, Corso, Criminisi, Das, Delingette, Demiralp, Durst, Dojat, Doyle, Festa, Forbes, Geremia, Glocker, Golland, Guo, Hamamci, Iftekharuddin, Jena, John, Konukoglu, Lashkari, Mariz, Meier, Pereira, Precup, Price, Raviv, Reza, Ryan, Sarikaya, Schwartz, Shin, Shotton, Silva, Sousa, Subbanna, Szekely, Taylor, Thomas, Tustison, Unal, Vasseur, Wintermark, Ye, Zhao, Zhao, Zikic, Prastawa, Reyes and Leemput}]{menze_multimodal_2015}
\bibinfo{author}{Menze, B.H.}, \bibinfo{author}{Jakab, A.}, \bibinfo{author}{Bauer, S.}, \bibinfo{author}{Kalpathy-Cramer, J.}, \bibinfo{author}{Farahani, K.}, \bibinfo{author}{Kirby, J.}, \bibinfo{author}{Burren, Y.}, \bibinfo{author}{Porz, N.}, \bibinfo{author}{Slotboom, J.}, \bibinfo{author}{Wiest, R.}, \bibinfo{author}{Lanczi, L.}, \bibinfo{author}{Gerstner, E.}, \bibinfo{author}{Weber, M.A.}, \bibinfo{author}{Arbel, T.}, \bibinfo{author}{Avants, B.B.}, \bibinfo{author}{Ayache, N.}, \bibinfo{author}{Buendia, P.}, \bibinfo{author}{Collins, D.L.}, \bibinfo{author}{Cordier, N.}, \bibinfo{author}{Corso, J.J.}, \bibinfo{author}{Criminisi, A.}, \bibinfo{author}{Das, T.}, \bibinfo{author}{Delingette, H.}, \bibinfo{author}{Demiralp, C.}, \bibinfo{author}{Durst, C.R.}, \bibinfo{author}{Dojat, M.}, \bibinfo{author}{Doyle, S.}, \bibinfo{author}{Festa, J.}, \bibinfo{author}{Forbes, F.}, \bibinfo{author}{Geremia, E.}, \bibinfo{author}{Glocker, B.}, \bibinfo{author}{Golland, P.}, \bibinfo{author}{Guo, X.},
  \bibinfo{author}{Hamamci, A.}, \bibinfo{author}{Iftekharuddin, K.M.}, \bibinfo{author}{Jena, R.}, \bibinfo{author}{John, N.M.}, \bibinfo{author}{Konukoglu, E.}, \bibinfo{author}{Lashkari, D.}, \bibinfo{author}{Mariz, J.A.}, \bibinfo{author}{Meier, R.}, \bibinfo{author}{Pereira, S.}, \bibinfo{author}{Precup, D.}, \bibinfo{author}{Price, S.J.}, \bibinfo{author}{Raviv, T.R.}, \bibinfo{author}{Reza, S.M.S.}, \bibinfo{author}{Ryan, M.}, \bibinfo{author}{Sarikaya, D.}, \bibinfo{author}{Schwartz, L.}, \bibinfo{author}{Shin, H.C.}, \bibinfo{author}{Shotton, J.}, \bibinfo{author}{Silva, C.A.}, \bibinfo{author}{Sousa, N.}, \bibinfo{author}{Subbanna, N.K.}, \bibinfo{author}{Szekely, G.}, \bibinfo{author}{Taylor, T.J.}, \bibinfo{author}{Thomas, O.M.}, \bibinfo{author}{Tustison, N.J.}, \bibinfo{author}{Unal, G.}, \bibinfo{author}{Vasseur, F.}, \bibinfo{author}{Wintermark, M.}, \bibinfo{author}{Ye, D.H.}, \bibinfo{author}{Zhao, L.}, \bibinfo{author}{Zhao, B.}, \bibinfo{author}{Zikic, D.}, \bibinfo{author}{Prastawa, M.},
  \bibinfo{author}{Reyes, M.}, \bibinfo{author}{Leemput, K.V.}, \bibinfo{year}{2015}.
\newblock \bibinfo{title}{The {Multimodal} {Brain} {Tumor} {Image} {Segmentation} {Benchmark} ({BRATS})}.
\newblock \bibinfo{journal}{IEEE Transactions on Medical Imaging} \bibinfo{volume}{34}, \bibinfo{pages}{1993--2024}.
\newblock \DOIprefix\doi{10.1109/TMI.2014.2377694}. \bibinfo{note}{conference Name: IEEE Transactions on Medical Imaging}.
\bibitem[{Monteiro et~al.(2020)Monteiro, Folgoc, de~Castro, Pawlowski, Marques, Kamnitsas, van~der Wilk and Glocker}]{monteiro_stochastic_2020}
\bibinfo{author}{Monteiro, M.}, \bibinfo{author}{Folgoc, L.L.}, \bibinfo{author}{de~Castro, D.C.}, \bibinfo{author}{Pawlowski, N.}, \bibinfo{author}{Marques, B.}, \bibinfo{author}{Kamnitsas, K.}, \bibinfo{author}{van~der Wilk, M.}, \bibinfo{author}{Glocker, B.}, \bibinfo{year}{2020}.
\newblock \bibinfo{title}{Stochastic {Segmentation} {Networks}: {Modelling} {Spatially} {Correlated} {Aleatoric} {Uncertainty}}.
\newblock \bibinfo{journal}{arXiv:2006.06015 [cs]} \URLprefix \url{http://arxiv.org/abs/2006.06015}. \bibinfo{note}{arXiv: 2006.06015}.
\bibitem[{Morozov et~al.(2020)Morozov, Andreychenko, Pavlov, Vladzymyrskyy, Ledikhova, Gombolevskiy, Blokhin, Gelezhe, Gonchar and Chernina}]{morozov_mosmeddata_2020}
\bibinfo{author}{Morozov, S.P.}, \bibinfo{author}{Andreychenko, A.E.}, \bibinfo{author}{Pavlov, N.A.}, \bibinfo{author}{Vladzymyrskyy, A.V.}, \bibinfo{author}{Ledikhova, N.V.}, \bibinfo{author}{Gombolevskiy, V.A.}, \bibinfo{author}{Blokhin, I.A.}, \bibinfo{author}{Gelezhe, P.B.}, \bibinfo{author}{Gonchar, A.V.}, \bibinfo{author}{Chernina, V.Y.}, \bibinfo{year}{2020}.
\newblock \bibinfo{title}{{MosMedData}: {Chest} {CT} {Scans} {With} {COVID}-19 {Related} {Findings} {Dataset}}.
\newblock \URLprefix \url{http://arxiv.org/abs/2005.06465}, \DOIprefix\doi{10.48550/arXiv.2005.06465}. \bibinfo{note}{arXiv:2005.06465 [cs, eess]}.
\bibitem[{Nair et~al.(2020)Nair, Precup, Arnold and Arbel}]{nair_exploring_2020}
\bibinfo{author}{Nair, T.}, \bibinfo{author}{Precup, D.}, \bibinfo{author}{Arnold, D.L.}, \bibinfo{author}{Arbel, T.}, \bibinfo{year}{2020}.
\newblock \bibinfo{title}{Exploring uncertainty measures in deep networks for {Multiple} sclerosis lesion detection and segmentation}.
\newblock \bibinfo{journal}{Medical Image Analysis} \bibinfo{volume}{59}, \bibinfo{pages}{101557}.
\newblock \URLprefix \url{https://www.sciencedirect.com/science/article/pii/S1361841519300994}, \DOIprefix\doi{10.1016/j.media.2019.101557}.
\bibitem[{Ng et~al.(2023)Ng, Guo, Biswas, Petersen, Piechnik, Neubauer and Wright}]{ng_estimating_2023}
\bibinfo{author}{Ng, M.}, \bibinfo{author}{Guo, F.}, \bibinfo{author}{Biswas, L.}, \bibinfo{author}{Petersen, S.E.}, \bibinfo{author}{Piechnik, S.K.}, \bibinfo{author}{Neubauer, S.}, \bibinfo{author}{Wright, G.}, \bibinfo{year}{2023}.
\newblock \bibinfo{title}{Estimating {Uncertainty} in {Neural} {Networks} for {Cardiac} {MRI} {Segmentation}: {A} {Benchmark} {Study}}.
\newblock \bibinfo{journal}{IEEE Transactions on Biomedical Engineering} \bibinfo{volume}{70}, \bibinfo{pages}{1955--1966}.
\newblock \DOIprefix\doi{10.1109/TBME.2022.3232730}. \bibinfo{note}{conference Name: IEEE Transactions on Biomedical Engineering}.
\bibitem[{Nikolov et~al.(2020)Nikolov, Blackwell, Zverovitch, Mendes, Livne, De~Fauw, Patel, Meyer, Askham, Romera-Paredes, Kelly, Karthikesalingam, Chu, Carnell, Boon, D'Souza, Moinuddin, Consortium, Montgomery, Rees, Suleyman, Back, Hughes, Ledsam and Ronneberger}]{nikolov_deep_2020}
\bibinfo{author}{Nikolov, S.}, \bibinfo{author}{Blackwell, S.}, \bibinfo{author}{Zverovitch, A.}, \bibinfo{author}{Mendes, R.}, \bibinfo{author}{Livne, M.}, \bibinfo{author}{De~Fauw, J.}, \bibinfo{author}{Patel, Y.}, \bibinfo{author}{Meyer, C.}, \bibinfo{author}{Askham, H.}, \bibinfo{author}{Romera-Paredes, B.}, \bibinfo{author}{Kelly, C.}, \bibinfo{author}{Karthikesalingam, A.}, \bibinfo{author}{Chu, C.}, \bibinfo{author}{Carnell, D.}, \bibinfo{author}{Boon, C.}, \bibinfo{author}{D'Souza, D.}, \bibinfo{author}{Moinuddin, S.A.}, \bibinfo{author}{Consortium, D.R.}, \bibinfo{author}{Montgomery, H.}, \bibinfo{author}{Rees, G.}, \bibinfo{author}{Suleyman, M.}, \bibinfo{author}{Back, T.}, \bibinfo{author}{Hughes, C.}, \bibinfo{author}{Ledsam, J.R.}, \bibinfo{author}{Ronneberger, O.}, \bibinfo{year}{2020}.
\newblock \bibinfo{title}{Deep learning to achieve clinically applicable segmentation of head and neck anatomy for radiotherapy}.
\newblock \bibinfo{journal}{arXiv:1809.04430 [physics, stat]} \URLprefix \url{http://arxiv.org/abs/1809.04430}. \bibinfo{note}{arXiv: 1809.04430}.
\bibitem[{Ouyang et~al.(2022)Ouyang, Wang, Chen, Li, Bai, Kainz and Rueckert}]{ouyang_improved_2022}
\bibinfo{author}{Ouyang, C.}, \bibinfo{author}{Wang, S.}, \bibinfo{author}{Chen, C.}, \bibinfo{author}{Li, Z.}, \bibinfo{author}{Bai, W.}, \bibinfo{author}{Kainz, B.}, \bibinfo{author}{Rueckert, D.}, \bibinfo{year}{2022}.
\newblock \bibinfo{title}{Improved post-hoc probability calibration for out-of-domain {MRI} segmentation}.
\newblock \URLprefix \url{http://arxiv.org/abs/2208.02870}, \DOIprefix\doi{10.48550/arXiv.2208.02870}. \bibinfo{note}{arXiv:2208.02870 [cs]}.
\bibitem[{Paszke et~al.(2019)Paszke, Gross, Massa, Lerer, Bradbury, Chanan, Killeen, Lin, Gimelshein, Antiga, Desmaison, Köpf, Yang, DeVito, Raison, Tejani, Chilamkurthy, Steiner, Fang, Bai and Chintala}]{paszke_pytorch_2019}
\bibinfo{author}{Paszke, A.}, \bibinfo{author}{Gross, S.}, \bibinfo{author}{Massa, F.}, \bibinfo{author}{Lerer, A.}, \bibinfo{author}{Bradbury, J.}, \bibinfo{author}{Chanan, G.}, \bibinfo{author}{Killeen, T.}, \bibinfo{author}{Lin, Z.}, \bibinfo{author}{Gimelshein, N.}, \bibinfo{author}{Antiga, L.}, \bibinfo{author}{Desmaison, A.}, \bibinfo{author}{Köpf, A.}, \bibinfo{author}{Yang, E.}, \bibinfo{author}{DeVito, Z.}, \bibinfo{author}{Raison, M.}, \bibinfo{author}{Tejani, A.}, \bibinfo{author}{Chilamkurthy, S.}, \bibinfo{author}{Steiner, B.}, \bibinfo{author}{Fang, L.}, \bibinfo{author}{Bai, J.}, \bibinfo{author}{Chintala, S.}, \bibinfo{year}{2019}.
\newblock \bibinfo{title}{{PyTorch}: {An} {Imperative} {Style}, {High}-{Performance} {Deep} {Learning} {Library}}.
\newblock \URLprefix \url{http://arxiv.org/abs/1912.01703}, \DOIprefix\doi{10.48550/arXiv.1912.01703}. \bibinfo{note}{arXiv:1912.01703 [cs, stat]}.
\bibitem[{Pati et~al.(2021)Pati, Baid, Zenk, Edwards, Sheller, Reina, Foley, Gruzdev, Martin, Albarqouni, Chen, Shinohara, Reinke, Zimmerer, Freymann, Kirby, Davatzikos, Colen, Kotrotsou, Marcus, Milchenko, Nazeri, Fathallah-Shaykh, Wiest, Jakab, Weber, Mahajan, Maier-Hein, Kleesiek, Menze, Maier-Hein and Bakas}]{pati_federated_2021}
\bibinfo{author}{Pati, S.}, \bibinfo{author}{Baid, U.}, \bibinfo{author}{Zenk, M.}, \bibinfo{author}{Edwards, B.}, \bibinfo{author}{Sheller, M.}, \bibinfo{author}{Reina, G.A.}, \bibinfo{author}{Foley, P.}, \bibinfo{author}{Gruzdev, A.}, \bibinfo{author}{Martin, J.}, \bibinfo{author}{Albarqouni, S.}, \bibinfo{author}{Chen, Y.}, \bibinfo{author}{Shinohara, R.T.}, \bibinfo{author}{Reinke, A.}, \bibinfo{author}{Zimmerer, D.}, \bibinfo{author}{Freymann, J.B.}, \bibinfo{author}{Kirby, J.S.}, \bibinfo{author}{Davatzikos, C.}, \bibinfo{author}{Colen, R.R.}, \bibinfo{author}{Kotrotsou, A.}, \bibinfo{author}{Marcus, D.}, \bibinfo{author}{Milchenko, M.}, \bibinfo{author}{Nazeri, A.}, \bibinfo{author}{Fathallah-Shaykh, H.}, \bibinfo{author}{Wiest, R.}, \bibinfo{author}{Jakab, A.}, \bibinfo{author}{Weber, M.A.}, \bibinfo{author}{Mahajan, A.}, \bibinfo{author}{Maier-Hein, L.}, \bibinfo{author}{Kleesiek, J.}, \bibinfo{author}{Menze, B.}, \bibinfo{author}{Maier-Hein, K.}, \bibinfo{author}{Bakas, S.}, \bibinfo{year}{2021}.
\newblock \bibinfo{title}{The {Federated} {Tumor} {Segmentation} ({FeTS}) {Challenge}}.
\newblock \bibinfo{journal}{arXiv:2105.05874 [cs, eess]} \URLprefix \url{http://arxiv.org/abs/2105.05874}. \bibinfo{note}{arXiv: 2105.05874}.
\bibitem[{Pedregosa et~al.(2011)Pedregosa, Varoquaux, Gramfort, Michel, Thirion, Grisel, Blondel, Prettenhofer, Weiss, Dubourg, Vanderplas, Passos, Cournapeau, Brucher, Perrot and Duchesnay}]{pedregosa_scikit-learn_2011}
\bibinfo{author}{Pedregosa, F.}, \bibinfo{author}{Varoquaux, G.}, \bibinfo{author}{Gramfort, A.}, \bibinfo{author}{Michel, V.}, \bibinfo{author}{Thirion, B.}, \bibinfo{author}{Grisel, O.}, \bibinfo{author}{Blondel, M.}, \bibinfo{author}{Prettenhofer, P.}, \bibinfo{author}{Weiss, R.}, \bibinfo{author}{Dubourg, V.}, \bibinfo{author}{Vanderplas, J.}, \bibinfo{author}{Passos, A.}, \bibinfo{author}{Cournapeau, D.}, \bibinfo{author}{Brucher, M.}, \bibinfo{author}{Perrot, M.}, \bibinfo{author}{Duchesnay, E.}, \bibinfo{year}{2011}.
\newblock \bibinfo{title}{Scikit-learn: {Machine} {Learning} in {Python}}.
\newblock \bibinfo{journal}{Journal of Machine Learning Research} \bibinfo{volume}{12}, \bibinfo{pages}{2825--2830}.
\newblock \URLprefix \url{http://jmlr.org/papers/v12/pedregosa11a.html}.
\bibitem[{Pérez-García et~al.(2021)Pérez-García, Sparks and Ourselin}]{perez-garcia_torchio_2021}
\bibinfo{author}{Pérez-García, F.}, \bibinfo{author}{Sparks, R.}, \bibinfo{author}{Ourselin, S.}, \bibinfo{year}{2021}.
\newblock \bibinfo{title}{{TorchIO}: {A} {Python} library for efficient loading, preprocessing, augmentation and patch-based sampling of medical images in deep learning}.
\newblock \bibinfo{journal}{Computer Methods and Programs in Biomedicine} \bibinfo{volume}{208}, \bibinfo{pages}{106236}.
\newblock \URLprefix \url{https://www.sciencedirect.com/science/article/pii/S0169260721003102}, \DOIprefix\doi{10.1016/j.cmpb.2021.106236}.
\bibitem[{Qiu et~al.(2023)Qiu, Chakrabarty, Nguyen, Ghosh and Sotiras}]{greenspan_qcresunet_2023}
\bibinfo{author}{Qiu, P.}, \bibinfo{author}{Chakrabarty, S.}, \bibinfo{author}{Nguyen, P.}, \bibinfo{author}{Ghosh, S.S.}, \bibinfo{author}{Sotiras, A.}, \bibinfo{year}{2023}.
\newblock \bibinfo{title}{{QCResUNet}: {Joint} {Subject}-{Level} and {Voxel}-{Level} {Prediction} of {Segmentation} {Quality}}, in: \bibinfo{editor}{Greenspan, H.}, \bibinfo{editor}{Madabhushi, A.}, \bibinfo{editor}{Mousavi, P.}, \bibinfo{editor}{Salcudean, S.}, \bibinfo{editor}{Duncan, J.}, \bibinfo{editor}{Syeda-Mahmood, T.}, \bibinfo{editor}{Taylor, R.} (Eds.), \bibinfo{booktitle}{Medical {Image} {Computing} and {Computer} {Assisted} {Intervention} – {MICCAI} 2023}. \bibinfo{publisher}{Springer Nature Switzerland}, \bibinfo{address}{Cham}. volume \bibinfo{volume}{14223}, pp. \bibinfo{pages}{173--182}.
\newblock \URLprefix \url{https://link.springer.com/10.1007/978-3-031-43901-8_17}, \DOIprefix\doi{10.1007/978-3-031-43901-8_17}. \bibinfo{note}{series Title: Lecture Notes in Computer Science}.
\bibitem[{Robinson et~al.(2018)Robinson, Oktay, Bai, Valindria, Sanghvi, Aung, Paiva, Zemrak, Fung, Lukaschuk, Lee, Carapella, Kim, Kainz, Piechnik, Neubauer, Petersen, Page, Rueckert and Glocker}]{robinson_real-time_2018}
\bibinfo{author}{Robinson, R.}, \bibinfo{author}{Oktay, O.}, \bibinfo{author}{Bai, W.}, \bibinfo{author}{Valindria, V.}, \bibinfo{author}{Sanghvi, M.}, \bibinfo{author}{Aung, N.}, \bibinfo{author}{Paiva, J.}, \bibinfo{author}{Zemrak, F.}, \bibinfo{author}{Fung, K.}, \bibinfo{author}{Lukaschuk, E.}, \bibinfo{author}{Lee, A.}, \bibinfo{author}{Carapella, V.}, \bibinfo{author}{Kim, Y.J.}, \bibinfo{author}{Kainz, B.}, \bibinfo{author}{Piechnik, S.}, \bibinfo{author}{Neubauer, S.}, \bibinfo{author}{Petersen, S.}, \bibinfo{author}{Page, C.}, \bibinfo{author}{Rueckert, D.}, \bibinfo{author}{Glocker, B.}, \bibinfo{year}{2018}.
\newblock \bibinfo{title}{Real-time {Prediction} of {Segmentation} {Quality}}.
\newblock \URLprefix \url{http://arxiv.org/abs/1806.06244}, \DOIprefix\doi{10.48550/arXiv.1806.06244}. \bibinfo{note}{arXiv:1806.06244 [cs]}.
\bibitem[{Ronneberger et~al.(2015)Ronneberger, Fischer and Brox}]{ronneberger_u-net:_2015}
\bibinfo{author}{Ronneberger, O.}, \bibinfo{author}{Fischer, P.}, \bibinfo{author}{Brox, T.}, \bibinfo{year}{2015}.
\newblock \bibinfo{title}{U-{Net}: {Convolutional} {Networks} for {Biomedical} {Image} {Segmentation}}.
\newblock \bibinfo{journal}{arXiv:1505.04597 [cs]} \URLprefix \url{http://arxiv.org/abs/1505.04597}. \bibinfo{note}{arXiv: 1505.04597}.
\bibitem[{Roth et~al.(2022)Roth, Xu, Tor-Díez, Sanchez~Jacob, Zember, Molto, Li, Xu, Turkbey, Turkbey, Yang, Harouni, Rieke, Hu, Isensee, Tang, Yu, Sölter, Zheng, Liauchuk, Zhou, Moltz, Oliveira, Xia, Maier-Hein, Li, Husch, Zhang, Kovalev, Kang, Hering, Vilaça, Flores, Xu, Wood and Linguraru}]{roth_rapid_2022}
\bibinfo{author}{Roth, H.R.}, \bibinfo{author}{Xu, Z.}, \bibinfo{author}{Tor-Díez, C.}, \bibinfo{author}{Sanchez~Jacob, R.}, \bibinfo{author}{Zember, J.}, \bibinfo{author}{Molto, J.}, \bibinfo{author}{Li, W.}, \bibinfo{author}{Xu, S.}, \bibinfo{author}{Turkbey, B.}, \bibinfo{author}{Turkbey, E.}, \bibinfo{author}{Yang, D.}, \bibinfo{author}{Harouni, A.}, \bibinfo{author}{Rieke, N.}, \bibinfo{author}{Hu, S.}, \bibinfo{author}{Isensee, F.}, \bibinfo{author}{Tang, C.}, \bibinfo{author}{Yu, Q.}, \bibinfo{author}{Sölter, J.}, \bibinfo{author}{Zheng, T.}, \bibinfo{author}{Liauchuk, V.}, \bibinfo{author}{Zhou, Z.}, \bibinfo{author}{Moltz, J.H.}, \bibinfo{author}{Oliveira, B.}, \bibinfo{author}{Xia, Y.}, \bibinfo{author}{Maier-Hein, K.H.}, \bibinfo{author}{Li, Q.}, \bibinfo{author}{Husch, A.}, \bibinfo{author}{Zhang, L.}, \bibinfo{author}{Kovalev, V.}, \bibinfo{author}{Kang, L.}, \bibinfo{author}{Hering, A.}, \bibinfo{author}{Vilaça, J.L.}, \bibinfo{author}{Flores, M.}, \bibinfo{author}{Xu, D.},
  \bibinfo{author}{Wood, B.}, \bibinfo{author}{Linguraru, M.G.}, \bibinfo{year}{2022}.
\newblock \bibinfo{title}{Rapid artificial intelligence solutions in a pandemic—{The} {COVID}-19-20 {Lung} {CT} {Lesion} {Segmentation} {Challenge}}.
\newblock \bibinfo{journal}{Medical Image Analysis} \bibinfo{volume}{82}, \bibinfo{pages}{102605}.
\newblock \URLprefix \url{https://www.sciencedirect.com/science/article/pii/S1361841522002353}, \DOIprefix\doi{10.1016/j.media.2022.102605}.
\bibitem[{Roy et~al.(2019)Roy, Conjeti, Navab and Wachinger}]{roy_bayesian_2019}
\bibinfo{author}{Roy, A.G.}, \bibinfo{author}{Conjeti, S.}, \bibinfo{author}{Navab, N.}, \bibinfo{author}{Wachinger, C.}, \bibinfo{year}{2019}.
\newblock \bibinfo{title}{Bayesian {QuickNAT}: {Model} uncertainty in deep whole-brain segmentation for structure-wise quality control}.
\newblock \bibinfo{journal}{NeuroImage} \bibinfo{volume}{195}, \bibinfo{pages}{11--22}.
\newblock \URLprefix \url{https://www.sciencedirect.com/science/article/pii/S1053811919302319}, \DOIprefix\doi{10.1016/j.neuroimage.2019.03.042}.
\bibitem[{Salehi et~al.(2022)Salehi, Mirzaei, Hendrycks, Li, Rohban and Sabokrou}]{salehi_unified_2022}
\bibinfo{author}{Salehi, M.}, \bibinfo{author}{Mirzaei, H.}, \bibinfo{author}{Hendrycks, D.}, \bibinfo{author}{Li, Y.}, \bibinfo{author}{Rohban, M.H.}, \bibinfo{author}{Sabokrou, M.}, \bibinfo{year}{2022}.
\newblock \bibinfo{title}{A {Unified} {Survey} on {Anomaly}, {Novelty}, {Open}-{Set}, and {Out} of-{Distribution} {Detection}: {Solutions} and {Future} {Challenges}}.
\newblock \bibinfo{journal}{Transactions on Machine Learning Research} \URLprefix \url{https://openreview.net/forum?id=aRtjVZvbpK&referrer=%5BTMLR%5D(%2Fgroup%3Fid%3DTMLR)}.
\bibitem[{Simpson et~al.(2019)Simpson, Antonelli, Bakas, Bilello, Farahani, van Ginneken, Kopp-Schneider, Landman, Litjens, Menze, Ronneberger, Summers, Bilic, Christ, Do, Gollub, Golia-Pernicka, Heckers, Jarnagin, McHugo, Napel, Vorontsov, Maier-Hein and Cardoso}]{simpson_large_2019}
\bibinfo{author}{Simpson, A.L.}, \bibinfo{author}{Antonelli, M.}, \bibinfo{author}{Bakas, S.}, \bibinfo{author}{Bilello, M.}, \bibinfo{author}{Farahani, K.}, \bibinfo{author}{van Ginneken, B.}, \bibinfo{author}{Kopp-Schneider, A.}, \bibinfo{author}{Landman, B.A.}, \bibinfo{author}{Litjens, G.}, \bibinfo{author}{Menze, B.}, \bibinfo{author}{Ronneberger, O.}, \bibinfo{author}{Summers, R.M.}, \bibinfo{author}{Bilic, P.}, \bibinfo{author}{Christ, P.F.}, \bibinfo{author}{Do, R.K.G.}, \bibinfo{author}{Gollub, M.}, \bibinfo{author}{Golia-Pernicka, J.}, \bibinfo{author}{Heckers, S.H.}, \bibinfo{author}{Jarnagin, W.R.}, \bibinfo{author}{McHugo, M.K.}, \bibinfo{author}{Napel, S.}, \bibinfo{author}{Vorontsov, E.}, \bibinfo{author}{Maier-Hein, L.}, \bibinfo{author}{Cardoso, M.J.}, \bibinfo{year}{2019}.
\newblock \bibinfo{title}{A large annotated medical image dataset for the development and evaluation of segmentation algorithms}.
\newblock \bibinfo{journal}{arXiv:1902.09063 [cs, eess]} \URLprefix \url{http://arxiv.org/abs/1902.09063}. \bibinfo{note}{arXiv: 1902.09063}.
\bibitem[{Valindria et~al.(2017)Valindria, Lavdas, Bai, Kamnitsas, Aboagye, Rockall, Rueckert and Glocker}]{valindria_reverse_2017}
\bibinfo{author}{Valindria, V.V.}, \bibinfo{author}{Lavdas, I.}, \bibinfo{author}{Bai, W.}, \bibinfo{author}{Kamnitsas, K.}, \bibinfo{author}{Aboagye, E.O.}, \bibinfo{author}{Rockall, A.G.}, \bibinfo{author}{Rueckert, D.}, \bibinfo{author}{Glocker, B.}, \bibinfo{year}{2017}.
\newblock \bibinfo{title}{Reverse {Classification} {Accuracy}: {Predicting} {Segmentation} {Performance} in the {Absence} of {Ground} {Truth}}.
\newblock \bibinfo{journal}{IEEE Transactions on Medical Imaging} \bibinfo{volume}{36}, \bibinfo{pages}{1597--1606}.
\newblock \DOIprefix\doi{10.1109/TMI.2017.2665165}. \bibinfo{note}{conference Name: IEEE Transactions on Medical Imaging}.
\bibitem[{Vasiliuk et~al.(2023)Vasiliuk, Frolova, Belyaev and Shirokikh}]{vasiliuk_redesigning_2023}
\bibinfo{author}{Vasiliuk, A.}, \bibinfo{author}{Frolova, D.}, \bibinfo{author}{Belyaev, M.}, \bibinfo{author}{Shirokikh, B.}, \bibinfo{year}{2023}.
\newblock \bibinfo{title}{Redesigning {Out}-of-{Distribution} {Detection} on {3D} {Medical} {Images}}.
\newblock \URLprefix \url{http://arxiv.org/abs/2308.07324}, \DOIprefix\doi{10.48550/arXiv.2308.07324}. \bibinfo{note}{arXiv:2308.07324 [cs, eess]}.
\bibitem[{Wang et~al.(2019)Wang, Li, Aertsen, Deprest, Ourselin and Vercauteren}]{wang_aleatoric_2019}
\bibinfo{author}{Wang, G.}, \bibinfo{author}{Li, W.}, \bibinfo{author}{Aertsen, M.}, \bibinfo{author}{Deprest, J.}, \bibinfo{author}{Ourselin, S.}, \bibinfo{author}{Vercauteren, T.}, \bibinfo{year}{2019}.
\newblock \bibinfo{title}{Aleatoric uncertainty estimation with test-time augmentation for medical image segmentation with convolutional neural networks}.
\newblock \bibinfo{journal}{Neurocomputing} \bibinfo{volume}{338}, \bibinfo{pages}{34--45}.
\newblock \URLprefix \url{https://www.sciencedirect.com/science/article/pii/S0925231219301961}, \DOIprefix\doi{10.1016/j.neucom.2019.01.103}.
\bibitem[{Wang et~al.(2020)Wang, Tarroni, Qin, Mo, Dai, Chen, Glocker, Guo, Rueckert and Bai}]{wang_deep_2020}
\bibinfo{author}{Wang, S.}, \bibinfo{author}{Tarroni, G.}, \bibinfo{author}{Qin, C.}, \bibinfo{author}{Mo, Y.}, \bibinfo{author}{Dai, C.}, \bibinfo{author}{Chen, C.}, \bibinfo{author}{Glocker, B.}, \bibinfo{author}{Guo, Y.}, \bibinfo{author}{Rueckert, D.}, \bibinfo{author}{Bai, W.}, \bibinfo{year}{2020}.
\newblock \bibinfo{title}{Deep {Generative} {Model}-based {Quality} {Control} for {Cardiac} {MRI} {Segmentation}}, volume \bibinfo{volume}{12264}, pp. \bibinfo{pages}{88--97}.
\newblock \URLprefix \url{http://arxiv.org/abs/2006.13379}, \DOIprefix\doi{10.1007/978-3-030-59719-1_9}. \bibinfo{note}{arXiv:2006.13379 [cs, eess]}.
\bibitem[{Xia et~al.(2020)Xia, Zhang, Liu, Shen and Yuille}]{xia_synthesize_2020}
\bibinfo{author}{Xia, Y.}, \bibinfo{author}{Zhang, Y.}, \bibinfo{author}{Liu, F.}, \bibinfo{author}{Shen, W.}, \bibinfo{author}{Yuille, A.}, \bibinfo{year}{2020}.
\newblock \bibinfo{title}{Synthesize then {Compare}: {Detecting} {Failures} and {Anomalies} for {Semantic} {Segmentation}}.
\newblock \bibinfo{journal}{arXiv:2003.08440 [cs]} \URLprefix \url{http://arxiv.org/abs/2003.08440}. \bibinfo{note}{arXiv: 2003.08440}.
\bibitem[{Zech et~al.(2018)Zech, Badgeley, Liu, Costa, Titano and Oermann}]{zech_variable_2018}
\bibinfo{author}{Zech, J.R.}, \bibinfo{author}{Badgeley, M.A.}, \bibinfo{author}{Liu, M.}, \bibinfo{author}{Costa, A.B.}, \bibinfo{author}{Titano, J.J.}, \bibinfo{author}{Oermann, E.K.}, \bibinfo{year}{2018}.
\newblock \bibinfo{title}{Variable generalization performance of a deep learning model to detect pneumonia in chest radiographs: {A} cross-sectional study}.
\newblock \bibinfo{journal}{PLOS Medicine} \bibinfo{volume}{15}, \bibinfo{pages}{e1002683}.
\newblock \URLprefix \url{https://journals.plos.org/plosmedicine/article?id=10.1371/journal.pmed.1002683}, \DOIprefix\doi{10.1371/journal.pmed.1002683}. \bibinfo{note}{publisher: Public Library of Science}.
\bibitem[{Zhao et~al.(2022)Zhao, Yang, Schweidtmann and Tao}]{wang_efficient_2022}
\bibinfo{author}{Zhao, Y.}, \bibinfo{author}{Yang, C.}, \bibinfo{author}{Schweidtmann, A.}, \bibinfo{author}{Tao, Q.}, \bibinfo{year}{2022}.
\newblock \bibinfo{title}{Efficient {Bayesian} {Uncertainty} {Estimation} for {nnU}-{Net}}, in: \bibinfo{editor}{Wang, L.}, \bibinfo{editor}{Dou, Q.}, \bibinfo{editor}{Fletcher, P.T.}, \bibinfo{editor}{Speidel, S.}, \bibinfo{editor}{Li, S.} (Eds.), \bibinfo{booktitle}{Medical {Image} {Computing} and {Computer} {Assisted} {Intervention} – {MICCAI} 2022}. \bibinfo{publisher}{Springer Nature Switzerland}, \bibinfo{address}{Cham}. volume \bibinfo{volume}{13438}, pp. \bibinfo{pages}{535--544}.
\newblock \URLprefix \url{https://link.springer.com/10.1007/978-3-031-16452-1_51}, \DOIprefix\doi{10.1007/978-3-031-16452-1_51}. \bibinfo{note}{series Title: Lecture Notes in Computer Science}.

\end{thebibliography}

\section*{Supplementary Material}

\appendix
\section{Test dataset examples}
\label{sec:dataset_details}

\begin{figure*}
    \centering
    \includegraphics[width=\textwidth]{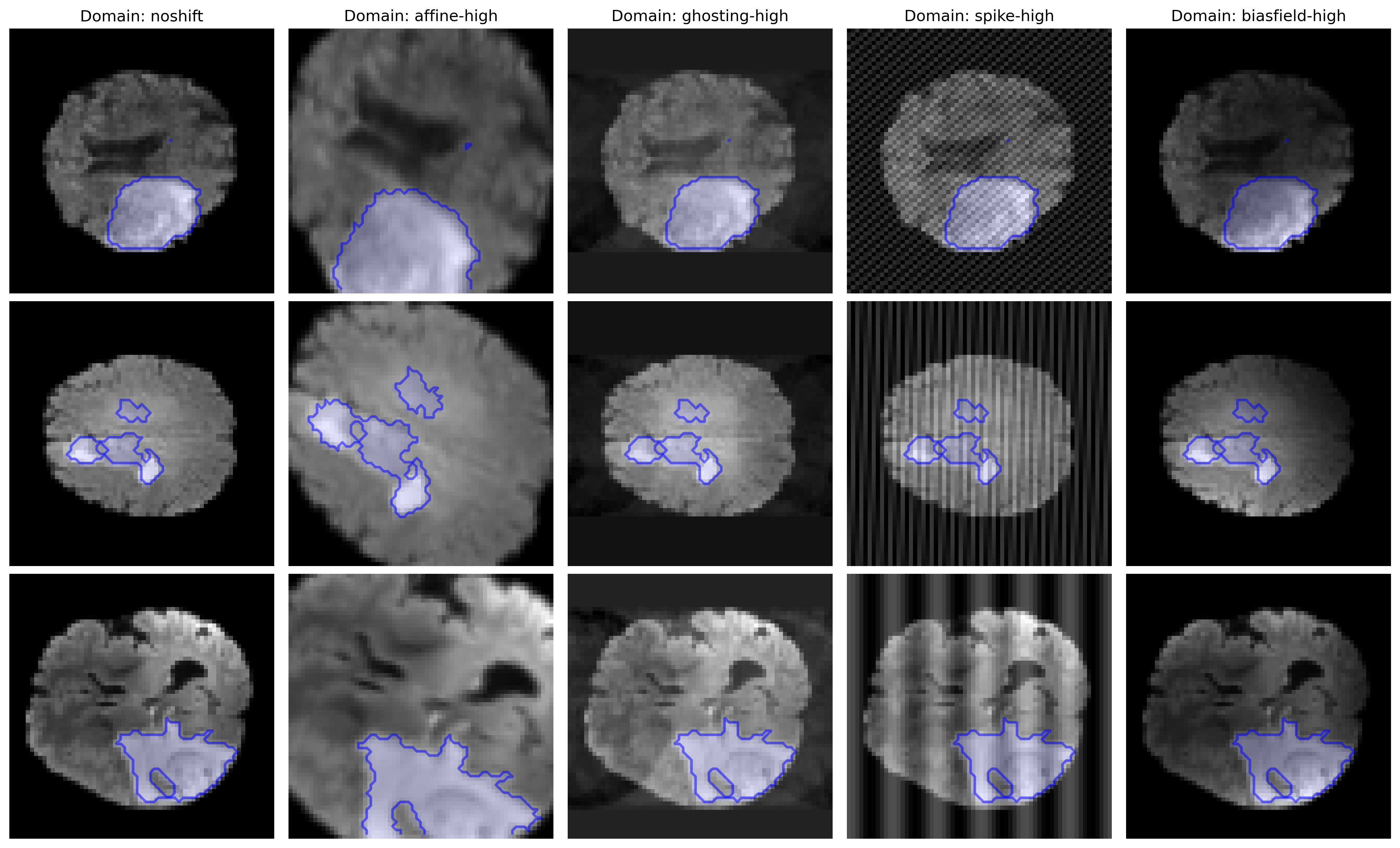}
    \caption{Samples from the test set of the 2D brain (toy) dataset. Each column shows samples from a different ``domain'', which corresponds to an artificial corruption for this dataset.}
    \label{fig:examples_brain2d}
\end{figure*}

\begin{figure*}
    \centering
    \includegraphics[width=\textwidth]{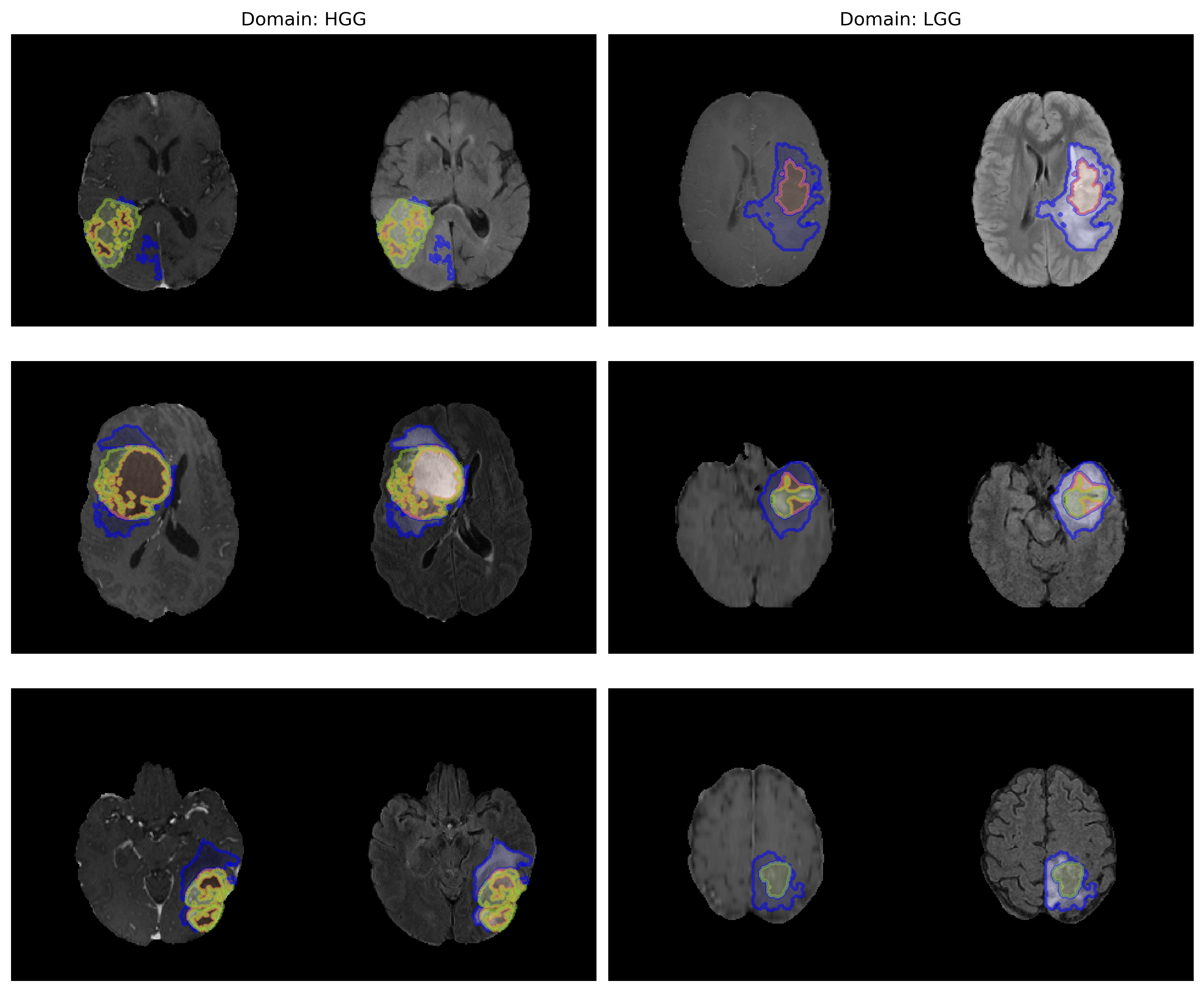}
    \caption{Samples from the test set of the brain tumor dataset. Each column shows samples from a different ``domain'', which corresponds to low-grade and high-grade gliomas.}
    \label{fig:examples_brain3d}
\end{figure*}

\begin{figure*}
    \centering
    \includegraphics[width=\textwidth]{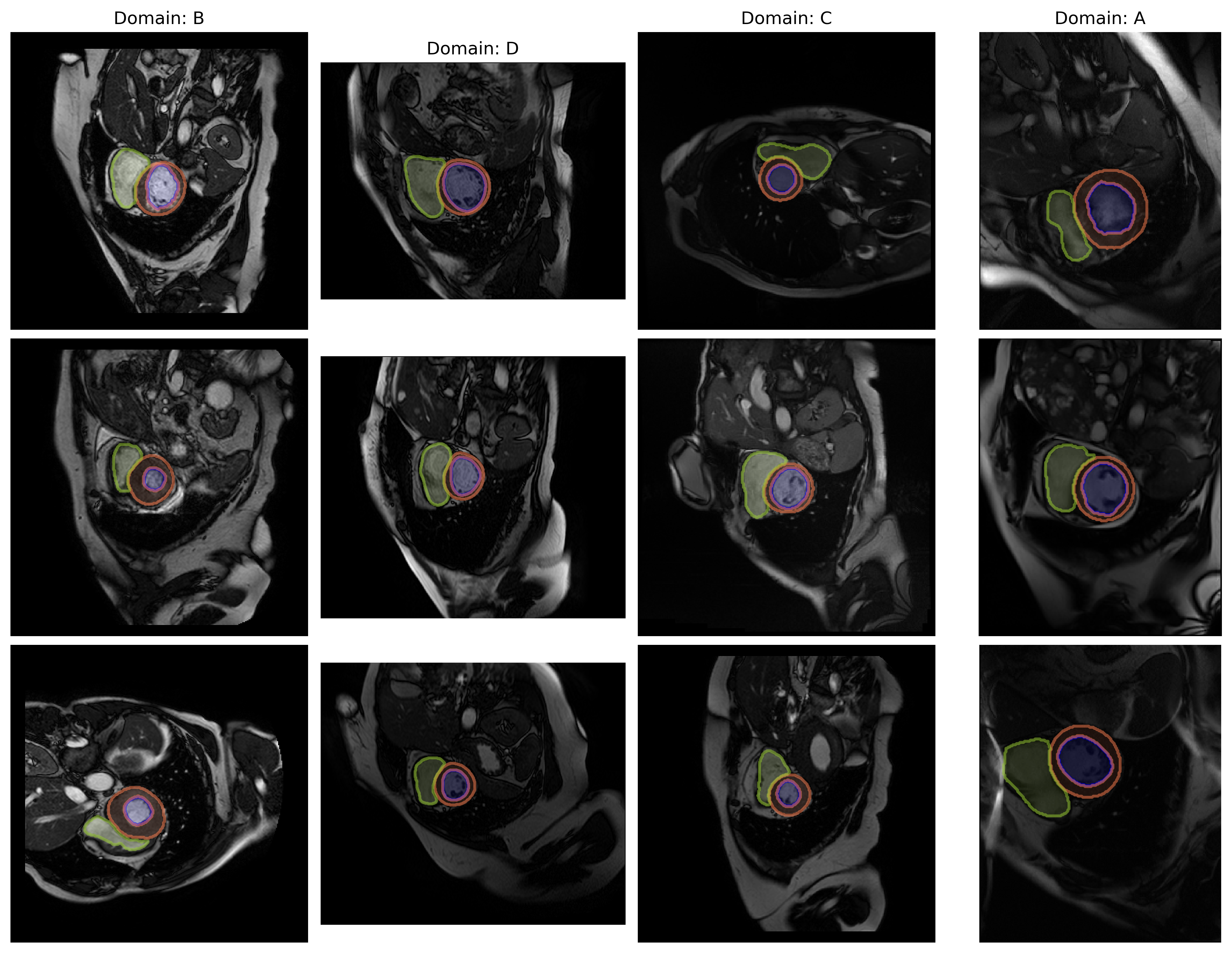}
    \caption{Samples from the test set of the heart dataset. Each column shows samples from a different ``domain'', which corresponds to different MR scanner vendors.}
    \label{fig:examples_heart}
\end{figure*}

\begin{figure*}
    \centering
    \includegraphics[width=\textwidth]{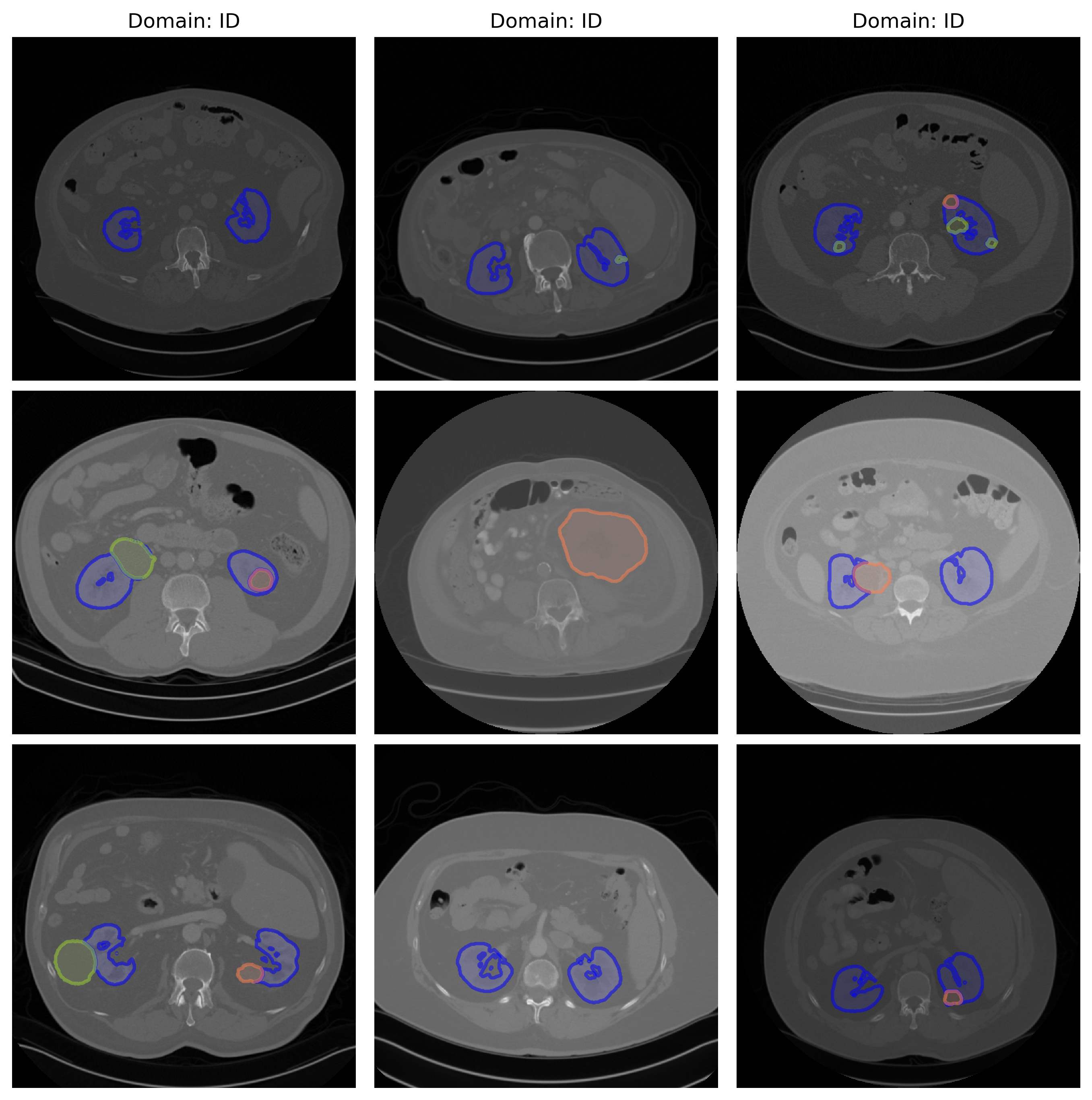}
    \caption{Samples from the test set of the kidney dataset. There is only one ``domain'' for this dataset.}
    \label{fig:examples_kidney}
\end{figure*}

\begin{figure*}
    \centering
    \includegraphics[width=\textwidth]{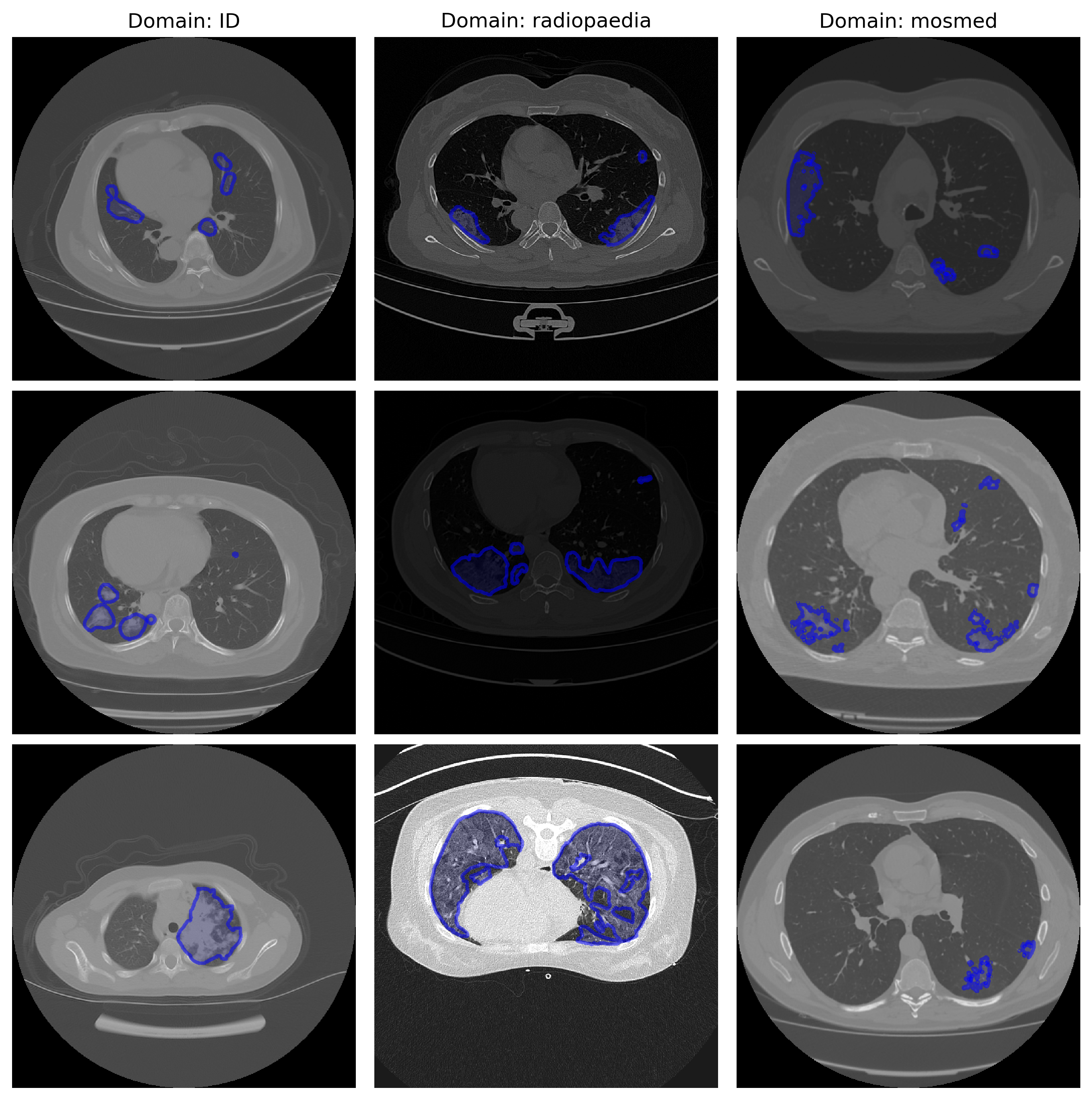}
    \caption{Samples from the test set of the Covid dataset. Each column shows samples from a different ``domain'', which corresponds to a different institution.}
    \label{fig:examples_covid}
\end{figure*}

\begin{figure*}
    \centering
    \includegraphics[width=\textwidth]{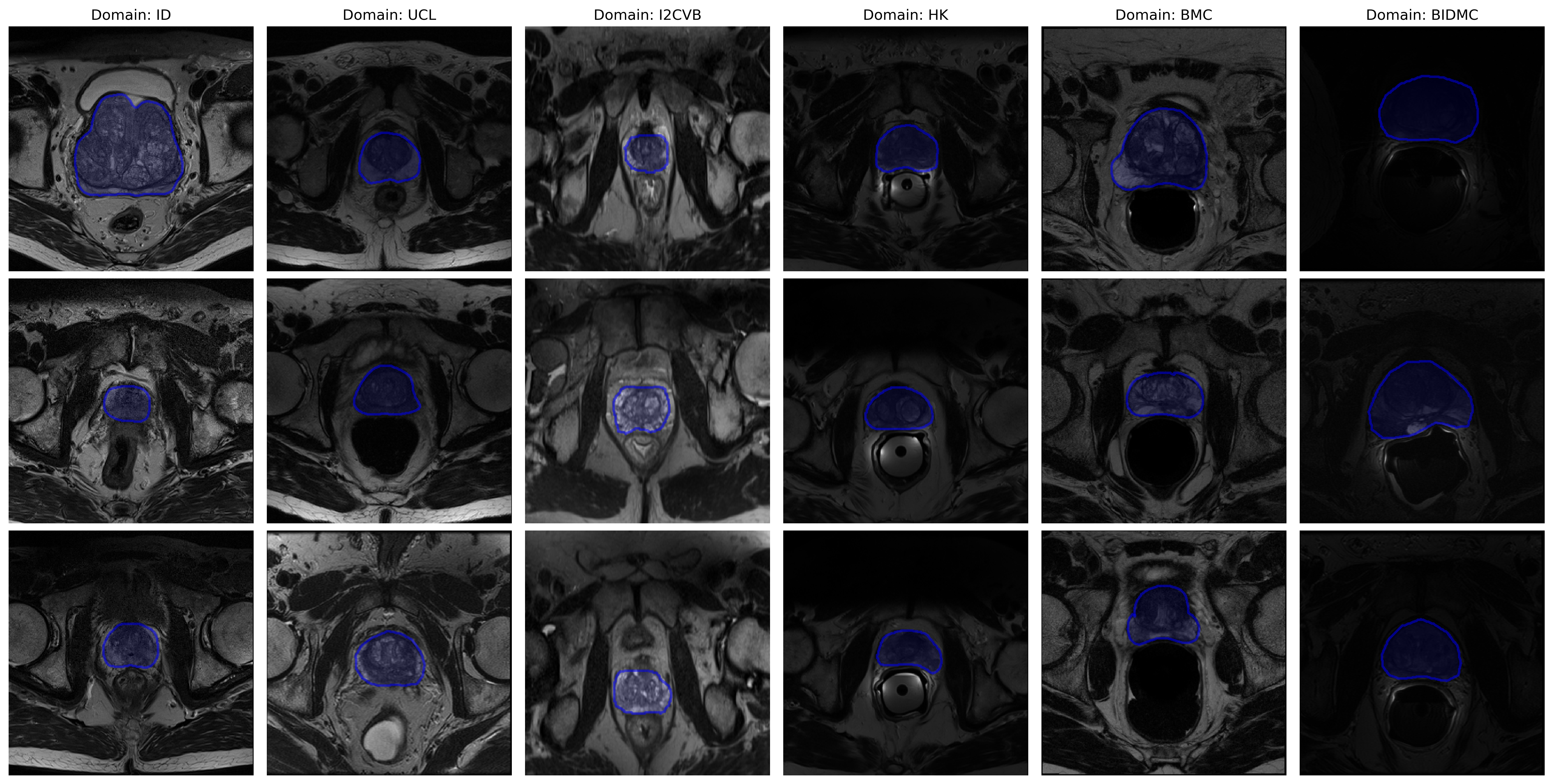}
    \caption{Samples from the test set of the prostate dataset. Each column shows samples from a different ``domain'', which corresponds to a different institution. Note that we did not apply intensity clipping to the images, as this is also not done in the segmentation network training.}
    \label{fig:examples_prostate}
\end{figure*}

In \cref{fig:examples_brain2d,fig:examples_brain3d,fig:examples_heart,fig:examples_kidney,fig:examples_covid,fig:examples_prostate}, we visualize example test cases from each dataset used for the benchmark.
As all datasets except the kidney tumor feature explicit distribution shifts, we show both samples from the training data distribution and from the shifted distributions.

\section{Publicly Available Code of Related Works}
\label{sec:code_availability}

We argue that progress in failure detection research is hampered by reproducibility issues related to the availability of source code for experiments conducted in related work. Here we list example references that range from no published code at all to a full release.
\begin{itemize}
    \item Works that do not provide code:
\citet{mehrtash_confidence_2020,robinson_real-time_2018,li_towards_2022,ng_estimating_2023,liu_alarm_2019,wang_deep_2020}
    \item Works with incomplete code: \citet{jungo_analyzing_2020} only provide an early version of the experiment code from a previous publication\footnote{\url{https://github.com/alainjungo/reliability-challenges-uncertainty}} with fewer aggregation methods. \citet{greenspan_qcresunet_2023} only provide the network architecture\footnote{\url{https://github.com/peijie-chiu/QC-ResUNet/tree/main}} but leave out the dataloading pipeline, which is an important part of their work. \citet{gonzalez_distance-based_2022} published their methods implementation\footnote{\url{https://github.com/MECLabTUDA/Lifelong-nnUNet/tree/dev-ood_detection}} but did not include exact dataset preparation steps.
    \item Works with complete code: 
\citet{greenspan_segmentation_2023,kahl_values_2024}
\end{itemize}
Apart from the availability of source code, other hurdles for reproducibility are the availability of the datasets used and the reporting of hyper-parameters. We did not analyze these aspects above.

\section{Additional Results}
\label{sec:additional_results}

Here we first report a large overview of all mean AURC values measured in our experiments (\cref{tab:all_results_aurc}), which essentially combines the results from \cref{fig:overview_aurc} and \cref{fig:overview_aurc_aggregation} and extends it with a few combinations not reported in the main results for clarity.
Note that we also include the popular \emph{mean foreground PE} baseline used for example in \citet{roy_bayesian_2019,graham_transformer-based_2022}. It averages the pixel confidence map only in the foreground region (excluding the boundary region with a width of 4 pixels).
Additionally, \cref{tab:all_results_aurc_3metrics} compares the results with AURC with the popular metrics Spearman and Pearson correlation.
\begin{table*}
\centering
\caption{Overview of AURC values (multiplied by 100) for all evaluated failure detection methods. The mean AURC is reported across 5 training folds and background color coding is applied per column, such that better scores (lower AURC) are lighter. Standard deviations across the five runs are given in the column next to the mean. The dataset names were abbreviated to save space. PE: predictive entropy, RF: regression forest.}
\label{tab:all_results_aurc}
\resizebox{\textwidth}{!}{
\begin{tabular}{lrlrlrlrlrlrl}
\toprule
Dataset & \multicolumn{2}{c}{Brain} & \multicolumn{2}{c}{Brain 2d} & \multicolumn{2}{c}{Covid} & \multicolumn{2}{c}{Heart} & \multicolumn{2}{c}{Kidney} & \multicolumn{2}{c}{Prostate} \\
 & mean & std & mean & std & mean & std & mean & std & mean & std & mean & std \\
Method &  &  &  &  &  &  &  &  &  &  &  &  \\
\midrule
Single network + mean PE & {\cellcolor[HTML]{792E6F}} \color[HTML]{F1F1F1} 20.1 & \color{gray} 1.3 & {\cellcolor[HTML]{4F2463}} \color[HTML]{F1F1F1} 15.1 & \color{gray} 1.3 & {\cellcolor[HTML]{692B6C}} \color[HTML]{F1F1F1} 36.5 & \color{gray} 1.1 & {\cellcolor[HTML]{D54E5F}} \color[HTML]{F1F1F1} 14.8 & \color{gray} 0.5 & {\cellcolor[HTML]{4B2362}} \color[HTML]{F1F1F1} 16.1 & \color{gray} 0.8 & {\cellcolor[HTML]{E67560}} \color[HTML]{F1F1F1} 27.3 & \color{gray} 2.5 \\
Single network + mean foreground PE & {\cellcolor[HTML]{D54E5F}} \color[HTML]{F1F1F1} 15.4 & \color{gray} 0.6 & {\cellcolor[HTML]{4B2362}} \color[HTML]{F1F1F1} 15.2 & \color{gray} 1.0 & {\cellcolor[HTML]{562666}} \color[HTML]{F1F1F1} 37.7 & \color{gray} 1.2 & {\cellcolor[HTML]{9B3670}} \color[HTML]{F1F1F1} 16.8 & \color{gray} 1.0 & {\cellcolor[HTML]{5C2868}} \color[HTML]{F1F1F1} 15.5 & \color{gray} 1.1 & {\cellcolor[HTML]{D34C60}} \color[HTML]{F1F1F1} 30.5 & \color{gray} 2.9 \\
Single network + non-boundary PE & {\cellcolor[HTML]{D24B60}} \color[HTML]{F1F1F1} 15.6 & \color{gray} 0.5 & {\cellcolor[HTML]{582766}} \color[HTML]{F1F1F1} 14.8 & \color{gray} 0.8 & {\cellcolor[HTML]{822F70}} \color[HTML]{F1F1F1} 35.0 & \color{gray} 1.4 & {\cellcolor[HTML]{E98C6B}} \color[HTML]{F1F1F1} 12.9 & \color{gray} 0.4 & {\cellcolor[HTML]{732C6E}} \color[HTML]{F1F1F1} 14.8 & \color{gray} 0.3 & {\cellcolor[HTML]{E77A62}} \color[HTML]{F1F1F1} 26.9 & \color{gray} 2.3 \\
Single network + patch-based PE & {\cellcolor[HTML]{C54266}} \color[HTML]{F1F1F1} 16.4 & \color{gray} 1.1 & {\cellcolor[HTML]{6D2B6D}} \color[HTML]{F1F1F1} 14.1 & \color{gray} 1.3 & {\cellcolor[HTML]{4B2362}} \color[HTML]{F1F1F1} 38.4 & \color{gray} 1.2 & {\cellcolor[HTML]{DE5C5C}} \color[HTML]{F1F1F1} 14.3 & \color{gray} 0.5 & {\cellcolor[HTML]{8A3171}} \color[HTML]{F1F1F1} 14.0 & \color{gray} 0.8 & {\cellcolor[HTML]{E46F5E}} \color[HTML]{F1F1F1} 27.7 & \color{gray} 2.6 \\
Single network + RF (simple PE-features) & {\cellcolor[HTML]{E98C6B}} \color[HTML]{F1F1F1} 12.3 & \color{gray} 0.3 & {\cellcolor[HTML]{D54E5F}} \color[HTML]{F1F1F1} 10.7 & \color{gray} 0.3 & {\cellcolor[HTML]{E67861}} \color[HTML]{F1F1F1} 27.2 & \color{gray} 1.3 & {\cellcolor[HTML]{E88466}} \color[HTML]{F1F1F1} 13.1 & \color{gray} 0.3 & {\cellcolor[HTML]{D64F5F}} \color[HTML]{F1F1F1} 11.5 & \color{gray} 0.9 & {\cellcolor[HTML]{652A6B}} \color[HTML]{F1F1F1} 38.8 & \color{gray} 4.5 \\
Single network + RF (radiomics PE-features) & {\cellcolor[HTML]{E98667}} \color[HTML]{F1F1F1} 12.5 & \color{gray} 0.3 & {\cellcolor[HTML]{C04068}} \color[HTML]{F1F1F1} 11.5 & \color{gray} 0.6 & {\cellcolor[HTML]{E98868}} \color[HTML]{F1F1F1} 26.3 & \color{gray} 0.8 & {\cellcolor[HTML]{E88265}} \color[HTML]{F1F1F1} 13.2 & \color{gray} 0.6 & {\cellcolor[HTML]{CA4564}} \color[HTML]{F1F1F1} 12.0 & \color{gray} 0.7 & {\cellcolor[HTML]{4B2362}} \color[HTML]{F1F1F1} 40.9 & \color{gray} 5.0 \\
Single network + Quality regression & {\cellcolor[HTML]{EB9C75}} \color[HTML]{000000} 11.5 & \color{gray} 0.2 & {\cellcolor[HTML]{E46C5D}} \color[HTML]{F1F1F1} 9.7 & \color{gray} 0.5 & {\cellcolor[HTML]{CC4663}} \color[HTML]{F1F1F1} 30.5 & \color{gray} 1.9 & {\cellcolor[HTML]{E77B62}} \color[HTML]{F1F1F1} 13.4 & \color{gray} 0.4 & {\cellcolor[HTML]{E88265}} \color[HTML]{F1F1F1} 9.8 & \color{gray} 0.5 & {\cellcolor[HTML]{C34167}} \color[HTML]{F1F1F1} 31.9 & \color{gray} 1.4 \\
Single network + Mahalanobis & {\cellcolor[HTML]{D64F5F}} \color[HTML]{F1F1F1} 15.3 & \color{gray} 1.7 & {\cellcolor[HTML]{873171}} \color[HTML]{F1F1F1} 13.3 & \color{gray} 0.4 & {\cellcolor[HTML]{E0615C}} \color[HTML]{F1F1F1} 28.5 & \color{gray} 0.7 & {\cellcolor[HTML]{D24A61}} \color[HTML]{F1F1F1} 15.0 & \color{gray} 0.6 & {\cellcolor[HTML]{692B6C}} \color[HTML]{F1F1F1} 15.1 & \color{gray} 0.9 & {\cellcolor[HTML]{AB3A6D}} \color[HTML]{F1F1F1} 33.7 & \color{gray} 2.4 \\
MC-Dropout + pairwise DSC & {\cellcolor[HTML]{EC9F76}} \color[HTML]{000000} 11.3 & \color{gray} 0.1 & {\cellcolor[HTML]{ECA278}} \color[HTML]{000000} 8.1 & \color{gray} 0.2 & {\cellcolor[HTML]{EB9C75}} \color[HTML]{000000} 25.1 & \color{gray} 0.9 & {\cellcolor[HTML]{EA8F6C}} \color[HTML]{F1F1F1} 12.8 & \color{gray} 0.4 & {\cellcolor[HTML]{ECA077}} \color[HTML]{000000} 8.9 & \color{gray} 0.4 & {\cellcolor[HTML]{EB9872}} \color[HTML]{000000} 24.8 & \color{gray} 1.2 \\
MC-Dropout + mean PE & {\cellcolor[HTML]{7B2E70}} \color[HTML]{F1F1F1} 20.1 & \color{gray} 1.2 & {\cellcolor[HTML]{E0605C}} \color[HTML]{F1F1F1} 10.1 & \color{gray} 0.6 & {\cellcolor[HTML]{6C2B6D}} \color[HTML]{F1F1F1} 36.3 & \color{gray} 1.1 & {\cellcolor[HTML]{D9525E}} \color[HTML]{F1F1F1} 14.7 & \color{gray} 0.5 & {\cellcolor[HTML]{522564}} \color[HTML]{F1F1F1} 15.8 & \color{gray} 0.9 & {\cellcolor[HTML]{E77E63}} \color[HTML]{F1F1F1} 26.6 & \color{gray} 2.0 \\
MC-Dropout + mean foreground PE & {\cellcolor[HTML]{D64F5F}} \color[HTML]{F1F1F1} 15.3 & \color{gray} 0.6 & {\cellcolor[HTML]{B53D6B}} \color[HTML]{F1F1F1} 11.8 & \color{gray} 0.6 & {\cellcolor[HTML]{562666}} \color[HTML]{F1F1F1} 37.7 & \color{gray} 1.2 & {\cellcolor[HTML]{A6396E}} \color[HTML]{F1F1F1} 16.4 & \color{gray} 0.9 & {\cellcolor[HTML]{542665}} \color[HTML]{F1F1F1} 15.8 & \color{gray} 1.4 & {\cellcolor[HTML]{E05F5C}} \color[HTML]{F1F1F1} 28.8 & \color{gray} 2.2 \\
MC-Dropout + non-boundary PE & {\cellcolor[HTML]{D24B60}} \color[HTML]{F1F1F1} 15.6 & \color{gray} 0.4 & {\cellcolor[HTML]{E46E5E}} \color[HTML]{F1F1F1} 9.6 & \color{gray} 0.2 & {\cellcolor[HTML]{843071}} \color[HTML]{F1F1F1} 34.9 & \color{gray} 1.5 & {\cellcolor[HTML]{E98C6B}} \color[HTML]{F1F1F1} 12.9 & \color{gray} 0.4 & {\cellcolor[HTML]{772D6F}} \color[HTML]{F1F1F1} 14.6 & \color{gray} 0.3 & {\cellcolor[HTML]{E88165}} \color[HTML]{F1F1F1} 26.4 & \color{gray} 2.1 \\
MC-Dropout + patch-based PE & {\cellcolor[HTML]{C44267}} \color[HTML]{F1F1F1} 16.4 & \color{gray} 1.0 & {\cellcolor[HTML]{E87F64}} \color[HTML]{F1F1F1} 9.1 & \color{gray} 0.5 & {\cellcolor[HTML]{4D2463}} \color[HTML]{F1F1F1} 38.2 & \color{gray} 1.2 & {\cellcolor[HTML]{E1645C}} \color[HTML]{F1F1F1} 14.1 & \color{gray} 0.3 & {\cellcolor[HTML]{923371}} \color[HTML]{F1F1F1} 13.8 & \color{gray} 0.6 & {\cellcolor[HTML]{E88165}} \color[HTML]{F1F1F1} 26.4 & \color{gray} 1.9 \\
Ensemble + pairwise DSC & {\cellcolor[HTML]{EDB081}} \color[HTML]{000000} 10.5 & \color{gray} 0.1 & {\cellcolor[HTML]{EDB081}} \color[HTML]{000000} 7.6 & \color{gray} 0.1 & {\cellcolor[HTML]{EDB081}} \color[HTML]{000000} 24.0 & \color{gray} 0.3 & {\cellcolor[HTML]{EDB081}} \color[HTML]{000000} 11.8 & \color{gray} 0.5 & {\cellcolor[HTML]{EDB081}} \color[HTML]{000000} 8.4 & \color{gray} 0.2 & {\cellcolor[HTML]{EDB081}} \color[HTML]{000000} 23.0 & \color{gray} 0.6 \\
Ensemble + mean PE & {\cellcolor[HTML]{C14068}} \color[HTML]{F1F1F1} 16.6 & \color{gray} 0.4 & {\cellcolor[HTML]{E98768}} \color[HTML]{F1F1F1} 8.9 & \color{gray} 0.3 & {\cellcolor[HTML]{A1386F}} \color[HTML]{F1F1F1} 33.1 & \color{gray} 1.5 & {\cellcolor[HTML]{E46D5D}} \color[HTML]{F1F1F1} 13.8 & \color{gray} 0.7 & {\cellcolor[HTML]{822F70}} \color[HTML]{F1F1F1} 14.3 & \color{gray} 0.5 & {\cellcolor[HTML]{EDAB7E}} \color[HTML]{000000} 23.4 & \color{gray} 0.8 \\
Ensemble + mean foreground PE & {\cellcolor[HTML]{E98768}} \color[HTML]{F1F1F1} 12.5 & \color{gray} 0.5 & {\cellcolor[HTML]{CC4663}} \color[HTML]{F1F1F1} 11.1 & \color{gray} 0.3 & {\cellcolor[HTML]{6F2C6D}} \color[HTML]{F1F1F1} 36.1 & \color{gray} 1.6 & {\cellcolor[HTML]{D9525E}} \color[HTML]{F1F1F1} 14.6 & \color{gray} 1.1 & {\cellcolor[HTML]{7C2E70}} \color[HTML]{F1F1F1} 14.5 & \color{gray} 0.8 & {\cellcolor[HTML]{EDAB7E}} \color[HTML]{000000} 23.4 & \color{gray} 1.2 \\
Ensemble + non-boundary PE & {\cellcolor[HTML]{E5705E}} \color[HTML]{F1F1F1} 13.6 & \color{gray} 0.4 & {\cellcolor[HTML]{E88064}} \color[HTML]{F1F1F1} 9.1 & \color{gray} 0.1 & {\cellcolor[HTML]{CF4862}} \color[HTML]{F1F1F1} 30.3 & \color{gray} 1.2 & {\cellcolor[HTML]{EB9A73}} \color[HTML]{000000} 12.5 & \color{gray} 0.5 & {\cellcolor[HTML]{953470}} \color[HTML]{F1F1F1} 13.7 & \color{gray} 0.2 & {\cellcolor[HTML]{EDAC7E}} \color[HTML]{000000} 23.4 & \color{gray} 0.8 \\
Ensemble + patch-based PE & {\cellcolor[HTML]{E3685C}} \color[HTML]{F1F1F1} 14.0 & \color{gray} 0.3 & {\cellcolor[HTML]{EB9872}} \color[HTML]{000000} 8.4 & \color{gray} 0.3 & {\cellcolor[HTML]{7D2E70}} \color[HTML]{F1F1F1} 35.3 & \color{gray} 1.5 & {\cellcolor[HTML]{E87F64}} \color[HTML]{F1F1F1} 13.3 & \color{gray} 0.6 & {\cellcolor[HTML]{A1376F}} \color[HTML]{F1F1F1} 13.3 & \color{gray} 0.6 & {\cellcolor[HTML]{EDAD7F}} \color[HTML]{000000} 23.3 & \color{gray} 1.0 \\
Ensemble + RF (simple PE-features) & {\cellcolor[HTML]{EB9C75}} \color[HTML]{000000} 11.5 & \color{gray} 0.2 & {\cellcolor[HTML]{ECA77B}} \color[HTML]{000000} 7.9 & \color{gray} 0.1 & {\cellcolor[HTML]{EB9872}} \color[HTML]{000000} 25.4 & \color{gray} 0.2 & {\cellcolor[HTML]{EB9973}} \color[HTML]{000000} 12.5 & \color{gray} 0.7 & {\cellcolor[HTML]{E46E5E}} \color[HTML]{F1F1F1} 10.4 & \color{gray} 0.3 & {\cellcolor[HTML]{863071}} \color[HTML]{F1F1F1} 36.5 & \color{gray} 2.2 \\
Ensemble + RF (radiomics PE-features) & {\cellcolor[HTML]{EA936F}} \color[HTML]{F1F1F1} 11.9 & \color{gray} 0.3 & {\cellcolor[HTML]{ECA077}} \color[HTML]{000000} 8.1 & \color{gray} 0.1 & {\cellcolor[HTML]{E98768}} \color[HTML]{F1F1F1} 26.3 & \color{gray} 0.5 & {\cellcolor[HTML]{E88366}} \color[HTML]{F1F1F1} 13.2 & \color{gray} 1.2 & {\cellcolor[HTML]{DD595C}} \color[HTML]{F1F1F1} 11.1 & \color{gray} 0.4 & {\cellcolor[HTML]{762D6F}} \color[HTML]{F1F1F1} 37.6 & \color{gray} 2.6 \\
Ensemble + Quality regression & {\cellcolor[HTML]{ECA67B}} \color[HTML]{000000} 11.0 & \color{gray} 0.2 & {\cellcolor[HTML]{E5705E}} \color[HTML]{F1F1F1} 9.6 & \color{gray} 0.6 & {\cellcolor[HTML]{D44D60}} \color[HTML]{F1F1F1} 29.8 & \color{gray} 1.1 & {\cellcolor[HTML]{E98667}} \color[HTML]{F1F1F1} 13.1 & \color{gray} 0.6 & {\cellcolor[HTML]{EB9973}} \color[HTML]{000000} 9.1 & \color{gray} 0.4 & {\cellcolor[HTML]{D64F5F}} \color[HTML]{F1F1F1} 30.2 & \color{gray} 1.3 \\
Ensemble + VAE (seg) & {\cellcolor[HTML]{4B2362}} \color[HTML]{F1F1F1} 22.5 & \color{gray} 1.2 & {\cellcolor[HTML]{B13C6C}} \color[HTML]{F1F1F1} 12.0 & \color{gray} 0.5 & {\cellcolor[HTML]{5A2767}} \color[HTML]{F1F1F1} 37.4 & \color{gray} 1.1 & {\cellcolor[HTML]{4B2362}} \color[HTML]{F1F1F1} 19.3 & \color{gray} 2.9 & {\cellcolor[HTML]{752D6F}} \color[HTML]{F1F1F1} 14.7 & \color{gray} 0.6 & {\cellcolor[HTML]{E98B6A}} \color[HTML]{F1F1F1} 25.8 & \color{gray} 0.9 \\
\bottomrule
\end{tabular}
}
\end{table*}

\begin{sidewaystable*}
\centering
\caption{Overview of values for AURC, Spearman correlation (SC) and Pearson correlation (PC) for all evaluated methods. Values are averaged across 5 training folds and background color coding is applied per column, such that better (i.e. lower) scores are lighter. SN: Single network. MCD: MC-Dropout. DE: Deep Ensemble. PE: predictive entropy, RF: regression forest.}
\label{tab:all_results_aurc_3metrics}
\resizebox{\textwidth}{!}{
    \begin{tabular}{l|ccc|ccc|ccc|ccc|ccc}
\toprule
dataset & \multicolumn{3}{c}{Brain tumor} & \multicolumn{3}{c}{Covid} & \multicolumn{3}{c}{Heart} & \multicolumn{3}{c}{Kidney tumor} & \multicolumn{3}{c}{Prostate} \\
 & AURC & PC & SC & AURC & PC & SC & AURC & PC & SC & AURC & PC & SC & AURC & PC & SC \\
Method &  &  &  &  &  &  &  &  &  &  &  &  &  &  &  \\
\midrule
SN + mean PE & {\cellcolor[HTML]{792E6F}} \color[HTML]{F1F1F1} 0.201 & {\cellcolor[HTML]{9F376F}} \color[HTML]{F1F1F1} -0.173 & {\cellcolor[HTML]{8F3371}} \color[HTML]{F1F1F1} -0.096 & {\cellcolor[HTML]{692B6C}} \color[HTML]{F1F1F1} 0.365 & {\cellcolor[HTML]{702C6E}} \color[HTML]{F1F1F1} -0.071 & {\cellcolor[HTML]{893171}} \color[HTML]{F1F1F1} -0.150 & {\cellcolor[HTML]{D54E5F}} \color[HTML]{F1F1F1} 0.148 & {\cellcolor[HTML]{903371}} \color[HTML]{F1F1F1} -0.482 & {\cellcolor[HTML]{D24B60}} \color[HTML]{F1F1F1} -0.532 & {\cellcolor[HTML]{4B2362}} \color[HTML]{F1F1F1} 0.161 & {\cellcolor[HTML]{973570}} \color[HTML]{F1F1F1} -0.267 & {\cellcolor[HTML]{572766}} \color[HTML]{F1F1F1} 0.021 & {\cellcolor[HTML]{E67560}} \color[HTML]{F1F1F1} 0.273 & {\cellcolor[HTML]{B23C6C}} \color[HTML]{F1F1F1} -0.289 & {\cellcolor[HTML]{E98969}} \color[HTML]{F1F1F1} -0.702 \\
SN + mean foreground PE & {\cellcolor[HTML]{D54E5F}} \color[HTML]{F1F1F1} 0.154 & {\cellcolor[HTML]{D44D60}} \color[HTML]{F1F1F1} -0.408 & {\cellcolor[HTML]{E0605C}} \color[HTML]{F1F1F1} -0.531 & {\cellcolor[HTML]{562666}} \color[HTML]{F1F1F1} 0.377 & {\cellcolor[HTML]{BE3F69}} \color[HTML]{F1F1F1} -0.300 & {\cellcolor[HTML]{8C3271}} \color[HTML]{F1F1F1} -0.159 & {\cellcolor[HTML]{9B3670}} \color[HTML]{F1F1F1} 0.168 & {\cellcolor[HTML]{552665}} \color[HTML]{F1F1F1} -0.317 & {\cellcolor[HTML]{913371}} \color[HTML]{F1F1F1} -0.338 & {\cellcolor[HTML]{5C2868}} \color[HTML]{F1F1F1} 0.155 & {\cellcolor[HTML]{A93A6E}} \color[HTML]{F1F1F1} -0.317 & {\cellcolor[HTML]{5E2868}} \color[HTML]{F1F1F1} -0.005 & {\cellcolor[HTML]{D34C60}} \color[HTML]{F1F1F1} 0.305 & {\cellcolor[HTML]{7F2F70}} \color[HTML]{F1F1F1} -0.116 & {\cellcolor[HTML]{DD595C}} \color[HTML]{F1F1F1} -0.528 \\
SN + non-boundary PE & {\cellcolor[HTML]{D24B60}} \color[HTML]{F1F1F1} 0.156 & {\cellcolor[HTML]{B93E6A}} \color[HTML]{F1F1F1} -0.281 & {\cellcolor[HTML]{D9525E}} \color[HTML]{F1F1F1} -0.461 & {\cellcolor[HTML]{822F70}} \color[HTML]{F1F1F1} 0.350 & {\cellcolor[HTML]{6D2B6D}} \color[HTML]{F1F1F1} -0.064 & {\cellcolor[HTML]{963570}} \color[HTML]{F1F1F1} -0.193 & {\cellcolor[HTML]{E98C6B}} \color[HTML]{F1F1F1} 0.129 & {\cellcolor[HTML]{E2675C}} \color[HTML]{F1F1F1} -0.767 & {\cellcolor[HTML]{E98868}} \color[HTML]{F1F1F1} -0.709 & {\cellcolor[HTML]{732C6E}} \color[HTML]{F1F1F1} 0.148 & {\cellcolor[HTML]{A4386F}} \color[HTML]{F1F1F1} -0.304 & {\cellcolor[HTML]{6E2C6D}} \color[HTML]{F1F1F1} -0.060 & {\cellcolor[HTML]{E77A62}} \color[HTML]{F1F1F1} 0.269 & {\cellcolor[HTML]{903371}} \color[HTML]{F1F1F1} -0.171 & {\cellcolor[HTML]{E98B6A}} \color[HTML]{F1F1F1} -0.708 \\
SN + patch-based PE & {\cellcolor[HTML]{C54266}} \color[HTML]{F1F1F1} 0.164 & {\cellcolor[HTML]{CB4564}} \color[HTML]{F1F1F1} -0.361 & {\cellcolor[HTML]{D14A61}} \color[HTML]{F1F1F1} -0.409 & {\cellcolor[HTML]{4B2362}} \color[HTML]{F1F1F1} 0.384 & {\cellcolor[HTML]{4E2463}} \color[HTML]{F1F1F1} 0.030 & {\cellcolor[HTML]{4B2362}} \color[HTML]{F1F1F1} 0.050 & {\cellcolor[HTML]{DE5C5C}} \color[HTML]{F1F1F1} 0.143 & {\cellcolor[HTML]{C14068}} \color[HTML]{F1F1F1} -0.621 & {\cellcolor[HTML]{DC585C}} \color[HTML]{F1F1F1} -0.579 & {\cellcolor[HTML]{8A3171}} \color[HTML]{F1F1F1} 0.140 & {\cellcolor[HTML]{983570}} \color[HTML]{F1F1F1} -0.271 & {\cellcolor[HTML]{8A3171}} \color[HTML]{F1F1F1} -0.159 & {\cellcolor[HTML]{E46F5E}} \color[HTML]{F1F1F1} 0.277 & {\cellcolor[HTML]{E88064}} \color[HTML]{F1F1F1} -0.591 & {\cellcolor[HTML]{E88466}} \color[HTML]{F1F1F1} -0.683 \\
SN + RF (simple PE-features) & {\cellcolor[HTML]{E98C6B}} \color[HTML]{F1F1F1} 0.123 & {\cellcolor[HTML]{EA906D}} \color[HTML]{F1F1F1} -0.682 & {\cellcolor[HTML]{EB9A73}} \color[HTML]{000000} -0.786 & {\cellcolor[HTML]{E67861}} \color[HTML]{F1F1F1} 0.272 & {\cellcolor[HTML]{E88466}} \color[HTML]{F1F1F1} -0.539 & {\cellcolor[HTML]{E67660}} \color[HTML]{F1F1F1} -0.533 & {\cellcolor[HTML]{E88466}} \color[HTML]{F1F1F1} 0.131 & {\cellcolor[HTML]{D64F5F}} \color[HTML]{F1F1F1} -0.696 & {\cellcolor[HTML]{E98667}} \color[HTML]{F1F1F1} -0.703 & {\cellcolor[HTML]{D64F5F}} \color[HTML]{F1F1F1} 0.115 & {\cellcolor[HTML]{DF5E5C}} \color[HTML]{F1F1F1} -0.497 & {\cellcolor[HTML]{DD595C}} \color[HTML]{F1F1F1} -0.479 & {\cellcolor[HTML]{652A6B}} \color[HTML]{F1F1F1} 0.388 & {\cellcolor[HTML]{662A6B}} \color[HTML]{F1F1F1} -0.034 & {\cellcolor[HTML]{752D6F}} \color[HTML]{F1F1F1} -0.108 \\
SN + RF (radiomics PE-features) & {\cellcolor[HTML]{E98667}} \color[HTML]{F1F1F1} 0.125 & {\cellcolor[HTML]{E88466}} \color[HTML]{F1F1F1} -0.636 & {\cellcolor[HTML]{EA936F}} \color[HTML]{F1F1F1} -0.751 & {\cellcolor[HTML]{E98868}} \color[HTML]{F1F1F1} 0.263 & {\cellcolor[HTML]{EA8E6C}} \color[HTML]{F1F1F1} -0.566 & {\cellcolor[HTML]{E77D63}} \color[HTML]{F1F1F1} -0.553 & {\cellcolor[HTML]{E88265}} \color[HTML]{F1F1F1} 0.132 & {\cellcolor[HTML]{C94465}} \color[HTML]{F1F1F1} -0.647 & {\cellcolor[HTML]{E98B6A}} \color[HTML]{F1F1F1} -0.717 & {\cellcolor[HTML]{CA4564}} \color[HTML]{F1F1F1} 0.120 & {\cellcolor[HTML]{E0615C}} \color[HTML]{F1F1F1} -0.507 & {\cellcolor[HTML]{D04962}} \color[HTML]{F1F1F1} -0.408 & {\cellcolor[HTML]{4B2362}} \color[HTML]{F1F1F1} 0.409 & {\cellcolor[HTML]{5E2868}} \color[HTML]{F1F1F1} -0.004 & {\cellcolor[HTML]{682A6C}} \color[HTML]{F1F1F1} -0.059 \\
SN + Quality regression & {\cellcolor[HTML]{EB9C75}} \color[HTML]{000000} 0.115 & {\cellcolor[HTML]{ECAA7D}} \color[HTML]{000000} -0.786 & {\cellcolor[HTML]{EDAB7E}} \color[HTML]{000000} -0.862 & {\cellcolor[HTML]{CC4663}} \color[HTML]{F1F1F1} 0.305 & {\cellcolor[HTML]{E0605C}} \color[HTML]{F1F1F1} -0.440 & {\cellcolor[HTML]{CF4862}} \color[HTML]{F1F1F1} -0.374 & {\cellcolor[HTML]{E77B62}} \color[HTML]{F1F1F1} 0.134 & {\cellcolor[HTML]{CD4763}} \color[HTML]{F1F1F1} -0.662 & {\cellcolor[HTML]{E46D5D}} \color[HTML]{F1F1F1} -0.637 & {\cellcolor[HTML]{E88265}} \color[HTML]{F1F1F1} 0.098 & {\cellcolor[HTML]{EDB081}} \color[HTML]{000000} -0.715 & {\cellcolor[HTML]{EA916E}} \color[HTML]{F1F1F1} -0.666 & {\cellcolor[HTML]{C34167}} \color[HTML]{F1F1F1} 0.319 & {\cellcolor[HTML]{DE5C5C}} \color[HTML]{F1F1F1} -0.472 & {\cellcolor[HTML]{D7505E}} \color[HTML]{F1F1F1} -0.491 \\
SN + Mahalanobis & {\cellcolor[HTML]{D64F5F}} \color[HTML]{F1F1F1} 0.153 & {\cellcolor[HTML]{DD5A5C}} \color[HTML]{F1F1F1} -0.470 & {\cellcolor[HTML]{DC585C}} \color[HTML]{F1F1F1} -0.491 & {\cellcolor[HTML]{E0615C}} \color[HTML]{F1F1F1} 0.285 & {\cellcolor[HTML]{D34C60}} \color[HTML]{F1F1F1} -0.370 & {\cellcolor[HTML]{DA545D}} \color[HTML]{F1F1F1} -0.426 & {\cellcolor[HTML]{D24A61}} \color[HTML]{F1F1F1} 0.150 & {\cellcolor[HTML]{63296A}} \color[HTML]{F1F1F1} -0.356 & {\cellcolor[HTML]{B73D6B}} \color[HTML]{F1F1F1} -0.448 & {\cellcolor[HTML]{692B6C}} \color[HTML]{F1F1F1} 0.151 & {\cellcolor[HTML]{4B2362}} \color[HTML]{F1F1F1} -0.055 & {\cellcolor[HTML]{642A6A}} \color[HTML]{F1F1F1} -0.028 & {\cellcolor[HTML]{AB3A6D}} \color[HTML]{F1F1F1} 0.337 & {\cellcolor[HTML]{C34167}} \color[HTML]{F1F1F1} -0.346 & {\cellcolor[HTML]{C14068}} \color[HTML]{F1F1F1} -0.390 \\
MCD + pairwise DSC & {\cellcolor[HTML]{EC9F76}} \color[HTML]{000000} 0.113 & {\cellcolor[HTML]{EA8F6C}} \color[HTML]{F1F1F1} -0.678 & {\cellcolor[HTML]{EDAE7F}} \color[HTML]{000000} -0.872 & {\cellcolor[HTML]{EB9C75}} \color[HTML]{000000} 0.251 & {\cellcolor[HTML]{E88265}} \color[HTML]{F1F1F1} -0.532 & {\cellcolor[HTML]{EDAB7E}} \color[HTML]{000000} -0.693 & {\cellcolor[HTML]{EA8F6C}} \color[HTML]{F1F1F1} 0.128 & {\cellcolor[HTML]{E98868}} \color[HTML]{F1F1F1} -0.853 & {\cellcolor[HTML]{EA9671}} \color[HTML]{F1F1F1} -0.747 & {\cellcolor[HTML]{ECA077}} \color[HTML]{000000} 0.089 & {\cellcolor[HTML]{EB9A73}} \color[HTML]{000000} -0.656 & {\cellcolor[HTML]{EDB081}} \color[HTML]{000000} -0.774 & {\cellcolor[HTML]{EB9872}} \color[HTML]{000000} 0.248 & {\cellcolor[HTML]{ECA479}} \color[HTML]{000000} -0.708 & {\cellcolor[HTML]{EDB081}} \color[HTML]{000000} -0.844 \\
MCD + mean PE & {\cellcolor[HTML]{7B2E70}} \color[HTML]{F1F1F1} 0.201 & {\cellcolor[HTML]{9F376F}} \color[HTML]{F1F1F1} -0.176 & {\cellcolor[HTML]{903371}} \color[HTML]{F1F1F1} -0.101 & {\cellcolor[HTML]{6C2B6D}} \color[HTML]{F1F1F1} 0.363 & {\cellcolor[HTML]{712C6E}} \color[HTML]{F1F1F1} -0.073 & {\cellcolor[HTML]{8B3271}} \color[HTML]{F1F1F1} -0.156 & {\cellcolor[HTML]{D9525E}} \color[HTML]{F1F1F1} 0.147 & {\cellcolor[HTML]{993570}} \color[HTML]{F1F1F1} -0.508 & {\cellcolor[HTML]{D7515E}} \color[HTML]{F1F1F1} -0.555 & {\cellcolor[HTML]{522564}} \color[HTML]{F1F1F1} 0.158 & {\cellcolor[HTML]{953470}} \color[HTML]{F1F1F1} -0.264 & {\cellcolor[HTML]{542665}} \color[HTML]{F1F1F1} 0.032 & {\cellcolor[HTML]{E77E63}} \color[HTML]{F1F1F1} 0.266 & {\cellcolor[HTML]{C44267}} \color[HTML]{F1F1F1} -0.349 & {\cellcolor[HTML]{EA946F}} \color[HTML]{F1F1F1} -0.739 \\
MCD + mean foreground PE & {\cellcolor[HTML]{D64F5F}} \color[HTML]{F1F1F1} 0.153 & {\cellcolor[HTML]{D54E5F}} \color[HTML]{F1F1F1} -0.414 & {\cellcolor[HTML]{E0615C}} \color[HTML]{F1F1F1} -0.534 & {\cellcolor[HTML]{562666}} \color[HTML]{F1F1F1} 0.377 & {\cellcolor[HTML]{C04068}} \color[HTML]{F1F1F1} -0.305 & {\cellcolor[HTML]{8C3271}} \color[HTML]{F1F1F1} -0.160 & {\cellcolor[HTML]{A6396E}} \color[HTML]{F1F1F1} 0.164 & {\cellcolor[HTML]{4B2362}} \color[HTML]{F1F1F1} -0.285 & {\cellcolor[HTML]{9D3670}} \color[HTML]{F1F1F1} -0.371 & {\cellcolor[HTML]{542665}} \color[HTML]{F1F1F1} 0.158 & {\cellcolor[HTML]{A8396E}} \color[HTML]{F1F1F1} -0.313 & {\cellcolor[HTML]{592767}} \color[HTML]{F1F1F1} 0.013 & {\cellcolor[HTML]{E05F5C}} \color[HTML]{F1F1F1} 0.288 & {\cellcolor[HTML]{9C3670}} \color[HTML]{F1F1F1} -0.213 & {\cellcolor[HTML]{E5745F}} \color[HTML]{F1F1F1} -0.627 \\
MCD + non-boundary PE & {\cellcolor[HTML]{D24B60}} \color[HTML]{F1F1F1} 0.156 & {\cellcolor[HTML]{BA3E6A}} \color[HTML]{F1F1F1} -0.284 & {\cellcolor[HTML]{D8525E}} \color[HTML]{F1F1F1} -0.459 & {\cellcolor[HTML]{843071}} \color[HTML]{F1F1F1} 0.349 & {\cellcolor[HTML]{6F2C6D}} \color[HTML]{F1F1F1} -0.069 & {\cellcolor[HTML]{993570}} \color[HTML]{F1F1F1} -0.201 & {\cellcolor[HTML]{E98C6B}} \color[HTML]{F1F1F1} 0.129 & {\cellcolor[HTML]{E46F5E}} \color[HTML]{F1F1F1} -0.788 & {\cellcolor[HTML]{E98969}} \color[HTML]{F1F1F1} -0.712 & {\cellcolor[HTML]{772D6F}} \color[HTML]{F1F1F1} 0.146 & {\cellcolor[HTML]{A5396E}} \color[HTML]{F1F1F1} -0.307 & {\cellcolor[HTML]{672A6B}} \color[HTML]{F1F1F1} -0.038 & {\cellcolor[HTML]{E88165}} \color[HTML]{F1F1F1} 0.264 & {\cellcolor[HTML]{9B3670}} \color[HTML]{F1F1F1} -0.210 & {\cellcolor[HTML]{EA946F}} \color[HTML]{F1F1F1} -0.741 \\
MCD + patch-based PE & {\cellcolor[HTML]{C44267}} \color[HTML]{F1F1F1} 0.164 & {\cellcolor[HTML]{CA4564}} \color[HTML]{F1F1F1} -0.356 & {\cellcolor[HTML]{D04962}} \color[HTML]{F1F1F1} -0.404 & {\cellcolor[HTML]{4D2463}} \color[HTML]{F1F1F1} 0.382 & {\cellcolor[HTML]{502564}} \color[HTML]{F1F1F1} 0.023 & {\cellcolor[HTML]{4D2463}} \color[HTML]{F1F1F1} 0.043 & {\cellcolor[HTML]{E1645C}} \color[HTML]{F1F1F1} 0.141 & {\cellcolor[HTML]{CB4564}} \color[HTML]{F1F1F1} -0.653 & {\cellcolor[HTML]{E2655C}} \color[HTML]{F1F1F1} -0.614 & {\cellcolor[HTML]{923371}} \color[HTML]{F1F1F1} 0.138 & {\cellcolor[HTML]{8F3371}} \color[HTML]{F1F1F1} -0.244 & {\cellcolor[HTML]{883171}} \color[HTML]{F1F1F1} -0.153 & {\cellcolor[HTML]{E88165}} \color[HTML]{F1F1F1} 0.264 & {\cellcolor[HTML]{ECA178}} \color[HTML]{000000} -0.698 & {\cellcolor[HTML]{EA9570}} \color[HTML]{F1F1F1} -0.743 \\
DE + pairwise DSC & {\cellcolor[HTML]{EDB081}} \color[HTML]{000000} 0.105 & {\cellcolor[HTML]{EDB081}} \color[HTML]{000000} -0.812 & {\cellcolor[HTML]{EDB081}} \color[HTML]{000000} -0.884 & {\cellcolor[HTML]{EDB081}} \color[HTML]{000000} 0.240 & {\cellcolor[HTML]{EDB081}} \color[HTML]{000000} -0.664 & {\cellcolor[HTML]{EDB081}} \color[HTML]{000000} -0.708 & {\cellcolor[HTML]{EDB081}} \color[HTML]{000000} 0.118 & {\cellcolor[HTML]{EDB081}} \color[HTML]{000000} -0.964 & {\cellcolor[HTML]{EDB081}} \color[HTML]{000000} -0.819 & {\cellcolor[HTML]{EDB081}} \color[HTML]{000000} 0.084 & {\cellcolor[HTML]{ECA077}} \color[HTML]{000000} -0.672 & {\cellcolor[HTML]{ECAA7D}} \color[HTML]{000000} -0.753 & {\cellcolor[HTML]{EDB081}} \color[HTML]{000000} 0.230 & {\cellcolor[HTML]{EDAD7F}} \color[HTML]{000000} -0.735 & {\cellcolor[HTML]{ECA77B}} \color[HTML]{000000} -0.811 \\
DE + mean PE & {\cellcolor[HTML]{C14068}} \color[HTML]{F1F1F1} 0.166 & {\cellcolor[HTML]{B33C6C}} \color[HTML]{F1F1F1} -0.258 & {\cellcolor[HTML]{BB3F6A}} \color[HTML]{F1F1F1} -0.300 & {\cellcolor[HTML]{A1386F}} \color[HTML]{F1F1F1} 0.331 & {\cellcolor[HTML]{822F70}} \color[HTML]{F1F1F1} -0.124 & {\cellcolor[HTML]{A8396E}} \color[HTML]{F1F1F1} -0.247 & {\cellcolor[HTML]{E46D5D}} \color[HTML]{F1F1F1} 0.138 & {\cellcolor[HTML]{A1386F}} \color[HTML]{F1F1F1} -0.533 & {\cellcolor[HTML]{E2665C}} \color[HTML]{F1F1F1} -0.618 & {\cellcolor[HTML]{822F70}} \color[HTML]{F1F1F1} 0.143 & {\cellcolor[HTML]{963570}} \color[HTML]{F1F1F1} -0.264 & {\cellcolor[HTML]{5D2868}} \color[HTML]{F1F1F1} -0.001 & {\cellcolor[HTML]{EDAB7E}} \color[HTML]{000000} 0.234 & {\cellcolor[HTML]{ECA67B}} \color[HTML]{000000} -0.713 & {\cellcolor[HTML]{ECA278}} \color[HTML]{000000} -0.792 \\
DE + mean foreground PE & {\cellcolor[HTML]{E98768}} \color[HTML]{F1F1F1} 0.125 & {\cellcolor[HTML]{E5745F}} \color[HTML]{F1F1F1} -0.574 & {\cellcolor[HTML]{E88366}} \color[HTML]{F1F1F1} -0.682 & {\cellcolor[HTML]{6F2C6D}} \color[HTML]{F1F1F1} 0.361 & {\cellcolor[HTML]{D24A61}} \color[HTML]{F1F1F1} -0.367 & {\cellcolor[HTML]{953470}} \color[HTML]{F1F1F1} -0.188 & {\cellcolor[HTML]{D9525E}} \color[HTML]{F1F1F1} 0.146 & {\cellcolor[HTML]{953470}} \color[HTML]{F1F1F1} -0.500 & {\cellcolor[HTML]{D64F5F}} \color[HTML]{F1F1F1} -0.545 & {\cellcolor[HTML]{7C2E70}} \color[HTML]{F1F1F1} 0.145 & {\cellcolor[HTML]{9B3670}} \color[HTML]{F1F1F1} -0.277 & {\cellcolor[HTML]{5B2867}} \color[HTML]{F1F1F1} 0.007 & {\cellcolor[HTML]{EDAB7E}} \color[HTML]{000000} 0.234 & {\cellcolor[HTML]{EDB081}} \color[HTML]{000000} -0.746 & {\cellcolor[HTML]{ECA77B}} \color[HTML]{000000} -0.809 \\
DE + non-boundary PE & {\cellcolor[HTML]{E5705E}} \color[HTML]{F1F1F1} 0.136 & {\cellcolor[HTML]{C64366}} \color[HTML]{F1F1F1} -0.336 & {\cellcolor[HTML]{E3695D}} \color[HTML]{F1F1F1} -0.569 & {\cellcolor[HTML]{CF4862}} \color[HTML]{F1F1F1} 0.303 & {\cellcolor[HTML]{8F3371}} \color[HTML]{F1F1F1} -0.163 & {\cellcolor[HTML]{CA4564}} \color[HTML]{F1F1F1} -0.358 & {\cellcolor[HTML]{EB9A73}} \color[HTML]{000000} 0.125 & {\cellcolor[HTML]{E5725F}} \color[HTML]{F1F1F1} -0.795 & {\cellcolor[HTML]{EA8E6C}} \color[HTML]{F1F1F1} -0.725 & {\cellcolor[HTML]{953470}} \color[HTML]{F1F1F1} 0.137 & {\cellcolor[HTML]{A4386F}} \color[HTML]{F1F1F1} -0.305 & {\cellcolor[HTML]{6D2B6D}} \color[HTML]{F1F1F1} -0.056 & {\cellcolor[HTML]{EDAC7E}} \color[HTML]{000000} 0.234 & {\cellcolor[HTML]{ECA278}} \color[HTML]{000000} -0.699 & {\cellcolor[HTML]{ECA077}} \color[HTML]{000000} -0.785 \\
DE + patch-based PE & {\cellcolor[HTML]{E3685C}} \color[HTML]{F1F1F1} 0.140 & {\cellcolor[HTML]{D9535D}} \color[HTML]{F1F1F1} -0.441 & {\cellcolor[HTML]{DC575C}} \color[HTML]{F1F1F1} -0.489 & {\cellcolor[HTML]{7D2E70}} \color[HTML]{F1F1F1} 0.353 & {\cellcolor[HTML]{6B2B6C}} \color[HTML]{F1F1F1} -0.059 & {\cellcolor[HTML]{642A6A}} \color[HTML]{F1F1F1} -0.030 & {\cellcolor[HTML]{E87F64}} \color[HTML]{F1F1F1} 0.133 & {\cellcolor[HTML]{C74366}} \color[HTML]{F1F1F1} -0.638 & {\cellcolor[HTML]{E77D63}} \color[HTML]{F1F1F1} -0.679 & {\cellcolor[HTML]{A1376F}} \color[HTML]{F1F1F1} 0.133 & {\cellcolor[HTML]{802F70}} \color[HTML]{F1F1F1} -0.203 & {\cellcolor[HTML]{843071}} \color[HTML]{F1F1F1} -0.135 & {\cellcolor[HTML]{EDAD7F}} \color[HTML]{000000} 0.233 & {\cellcolor[HTML]{EC9F76}} \color[HTML]{000000} -0.692 & {\cellcolor[HTML]{ECA57A}} \color[HTML]{000000} -0.802 \\
DE + RF (simple PE-features) & {\cellcolor[HTML]{EB9C75}} \color[HTML]{000000} 0.115 & {\cellcolor[HTML]{EB9E76}} \color[HTML]{000000} -0.741 & {\cellcolor[HTML]{EB9C75}} \color[HTML]{000000} -0.797 & {\cellcolor[HTML]{EB9872}} \color[HTML]{000000} 0.254 & {\cellcolor[HTML]{EB9A73}} \color[HTML]{000000} -0.603 & {\cellcolor[HTML]{E98D6B}} \color[HTML]{F1F1F1} -0.600 & {\cellcolor[HTML]{EB9973}} \color[HTML]{000000} 0.125 & {\cellcolor[HTML]{D44D60}} \color[HTML]{F1F1F1} -0.686 & {\cellcolor[HTML]{EC9F76}} \color[HTML]{000000} -0.772 & {\cellcolor[HTML]{E46E5E}} \color[HTML]{F1F1F1} 0.104 & {\cellcolor[HTML]{E46D5D}} \color[HTML]{F1F1F1} -0.538 & {\cellcolor[HTML]{DD5A5C}} \color[HTML]{F1F1F1} -0.484 & {\cellcolor[HTML]{863071}} \color[HTML]{F1F1F1} 0.365 & {\cellcolor[HTML]{712C6E}} \color[HTML]{F1F1F1} -0.066 & {\cellcolor[HTML]{692B6C}} \color[HTML]{F1F1F1} -0.065 \\
DE + RF (radiomics PE-features) & {\cellcolor[HTML]{EA936F}} \color[HTML]{F1F1F1} 0.119 & {\cellcolor[HTML]{E88466}} \color[HTML]{F1F1F1} -0.636 & {\cellcolor[HTML]{EA8E6C}} \color[HTML]{F1F1F1} -0.729 & {\cellcolor[HTML]{E98768}} \color[HTML]{F1F1F1} 0.263 & {\cellcolor[HTML]{E67560}} \color[HTML]{F1F1F1} -0.498 & {\cellcolor[HTML]{E46C5D}} \color[HTML]{F1F1F1} -0.502 & {\cellcolor[HTML]{E88366}} \color[HTML]{F1F1F1} 0.132 & {\cellcolor[HTML]{BE3F69}} \color[HTML]{F1F1F1} -0.612 & {\cellcolor[HTML]{EA8F6C}} \color[HTML]{F1F1F1} -0.728 & {\cellcolor[HTML]{DD595C}} \color[HTML]{F1F1F1} 0.111 & {\cellcolor[HTML]{D44D60}} \color[HTML]{F1F1F1} -0.446 & {\cellcolor[HTML]{CA4564}} \color[HTML]{F1F1F1} -0.386 & {\cellcolor[HTML]{762D6F}} \color[HTML]{F1F1F1} 0.376 & {\cellcolor[HTML]{4B2362}} \color[HTML]{F1F1F1} 0.063 & {\cellcolor[HTML]{4B2362}} \color[HTML]{F1F1F1} 0.054 \\
DE + Quality regression & {\cellcolor[HTML]{ECA67B}} \color[HTML]{000000} 0.110 & {\cellcolor[HTML]{ECA379}} \color[HTML]{000000} -0.759 & {\cellcolor[HTML]{ECA67B}} \color[HTML]{000000} -0.840 & {\cellcolor[HTML]{D44D60}} \color[HTML]{F1F1F1} 0.298 & {\cellcolor[HTML]{DD5A5C}} \color[HTML]{F1F1F1} -0.421 & {\cellcolor[HTML]{C94465}} \color[HTML]{F1F1F1} -0.355 & {\cellcolor[HTML]{E98667}} \color[HTML]{F1F1F1} 0.131 & {\cellcolor[HTML]{CC4663}} \color[HTML]{F1F1F1} -0.659 & {\cellcolor[HTML]{E2675C}} \color[HTML]{F1F1F1} -0.620 & {\cellcolor[HTML]{EB9973}} \color[HTML]{000000} 0.091 & {\cellcolor[HTML]{EDAC7E}} \color[HTML]{000000} -0.703 & {\cellcolor[HTML]{E98B6A}} \color[HTML]{F1F1F1} -0.648 & {\cellcolor[HTML]{D64F5F}} \color[HTML]{F1F1F1} 0.302 & {\cellcolor[HTML]{CE4763}} \color[HTML]{F1F1F1} -0.388 & {\cellcolor[HTML]{CD4763}} \color[HTML]{F1F1F1} -0.443 \\
DE + VAE (seg) & {\cellcolor[HTML]{4B2362}} \color[HTML]{F1F1F1} 0.225 & {\cellcolor[HTML]{4B2362}} \color[HTML]{F1F1F1} 0.174 & {\cellcolor[HTML]{4B2362}} \color[HTML]{F1F1F1} 0.222 & {\cellcolor[HTML]{5A2767}} \color[HTML]{F1F1F1} 0.374 & {\cellcolor[HTML]{4B2362}} \color[HTML]{F1F1F1} 0.040 & {\cellcolor[HTML]{6C2B6D}} \color[HTML]{F1F1F1} -0.058 & {\cellcolor[HTML]{4B2362}} \color[HTML]{F1F1F1} 0.193 & {\cellcolor[HTML]{642A6A}} \color[HTML]{F1F1F1} -0.358 & {\cellcolor[HTML]{4B2362}} \color[HTML]{F1F1F1} -0.137 & {\cellcolor[HTML]{752D6F}} \color[HTML]{F1F1F1} 0.147 & {\cellcolor[HTML]{923371}} \color[HTML]{F1F1F1} -0.252 & {\cellcolor[HTML]{4B2362}} \color[HTML]{F1F1F1} 0.067 & {\cellcolor[HTML]{E98B6A}} \color[HTML]{F1F1F1} 0.258 & {\cellcolor[HTML]{C04068}} \color[HTML]{F1F1F1} -0.333 & {\cellcolor[HTML]{E67961}} \color[HTML]{F1F1F1} -0.646 \\
\bottomrule
\end{tabular}
}
\end{sidewaystable*}

\begin{figure*}[t]
    \centering
    \includegraphics[width=\textwidth]{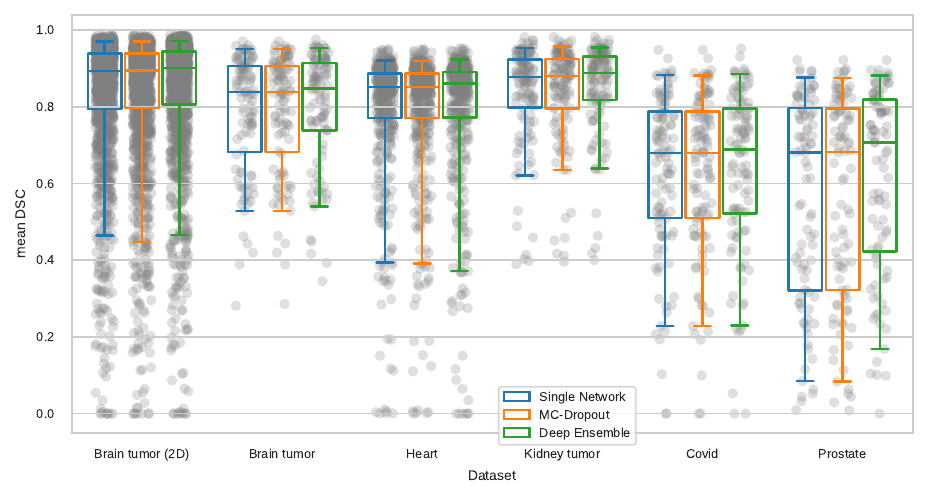}
    \caption{
    Comparison of the segmentation performances of single network, MC-Dropout and ensemble on all test sets.
    Boxes show the median and IQR, while whiskers extend to the 5th and 95th percentiles, respectively.
    The ensemble has consistently higher DSC scores than the other two models. Failure cases with low DSC are, however, present for all models.
    }
    \label{fig:seg_performance_model_comparison}
\end{figure*}

Since AURC is affected by gains in segmentation performance when using MC-Dropout or ensemble instead of standard single-network inference, we illustrate the differences in segmentation performance between these models in \cref{fig:seg_performance_model_comparison}. As expected, the ensemble consistently achieves higher DSC scores, but the difference in the median is small for all datasets.

\begin{figure*}[!t]
    \centering
    \includegraphics[width=\textwidth]{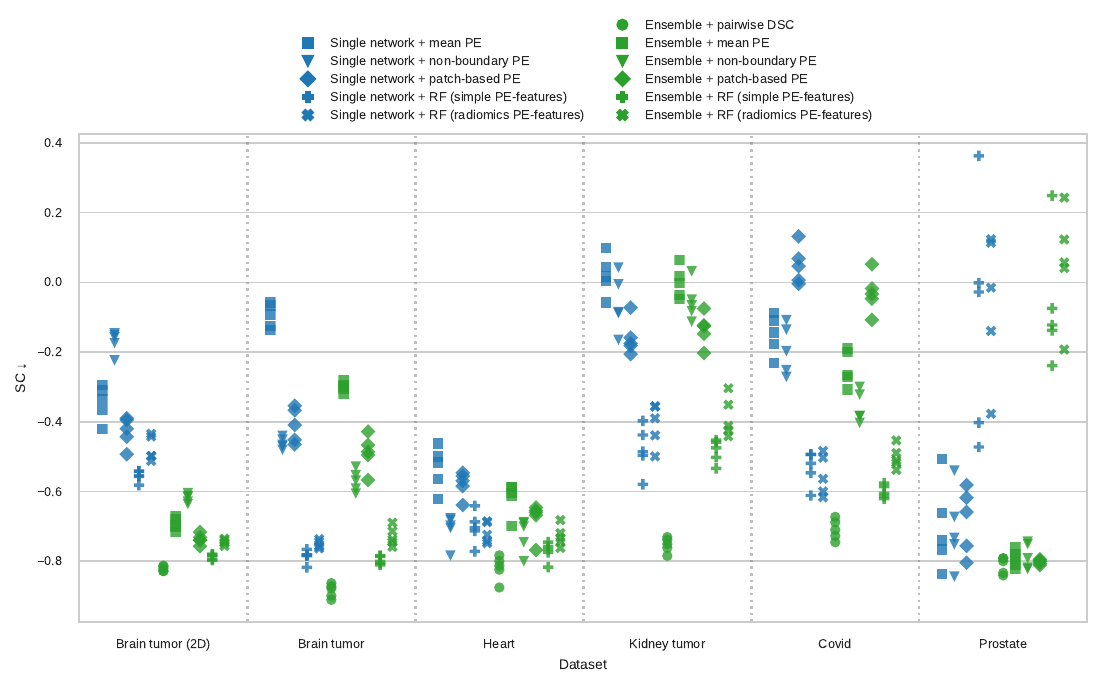}
    \caption{
        Comparison of aggregation methods in terms of Spearman correlation coefficients (SC) for all datasets (negative correlation is better).  The experiments are named as ``prediction model + confidence method'' and each of them was repeated using 5 folds.
        SC neglects segmentation performance so that ensembles do not have an advantage. Nonetheless, comparing the SC of the same aggregation method between single network and ensemble (same marker style, different color), we see that the ensemble is still always better for mean, non-boundary and patch-based aggregation, which might indicate better confidence maps.
        PE: predictive entropy. RF: regression forest.
    }
    \label{fig:overview_agg_spearman}
\end{figure*}

\Cref{fig:overview_agg_spearman} compares the same methods as \cref{fig:overview_aurc_aggregation} but measures performance with the Spearman correlation (SC). This neglects the segmentation performance aspect (which is often not desired, see requirement R2), but here it is helpful to compare the single network and ensemble results only with respect to their confidence ranking capabilities.

\begin{figure*}[!t]
    \centering
    \includegraphics[width=\textwidth]{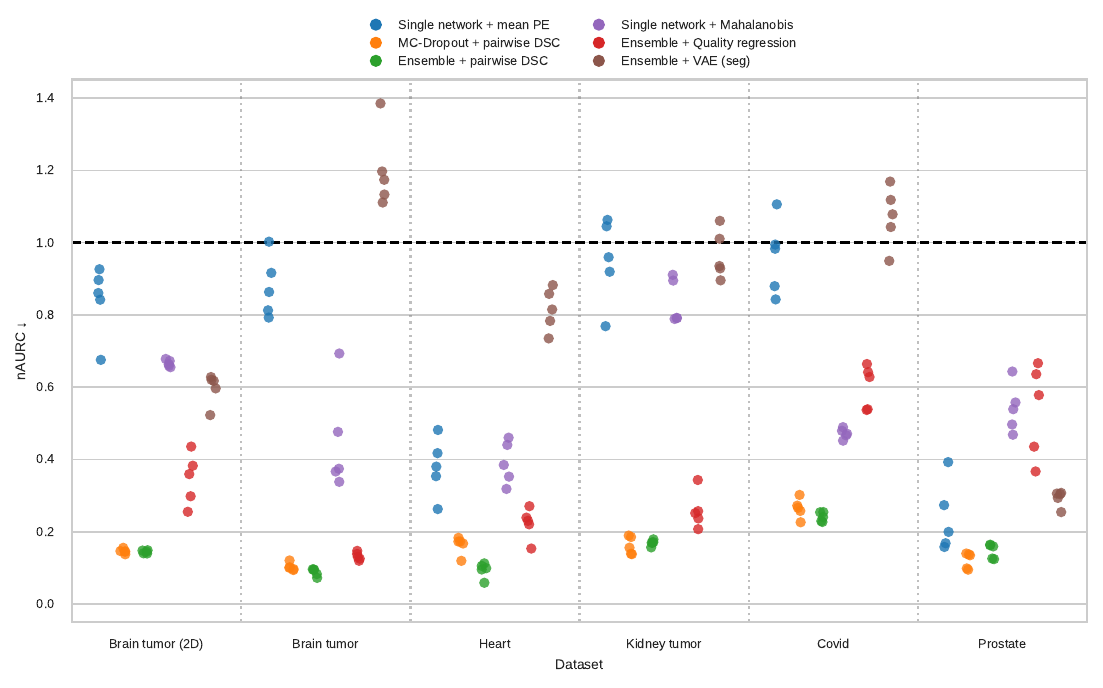}
    \caption{
        Normalized AURC (nAURC) scores for all datasets and methods (lower is better; 1 corresponds to random performance and 0 to optimal). Each experiment was repeated using 5 folds.
        Most of the pixel-confidence aggregation methods were excluded for clarity, as they perform worse than pairwise Dice.
    }
    \label{fig:overview_naurc}
\end{figure*}

\begin{figure*}[!t]
    \centering
    \includegraphics[width=\textwidth]{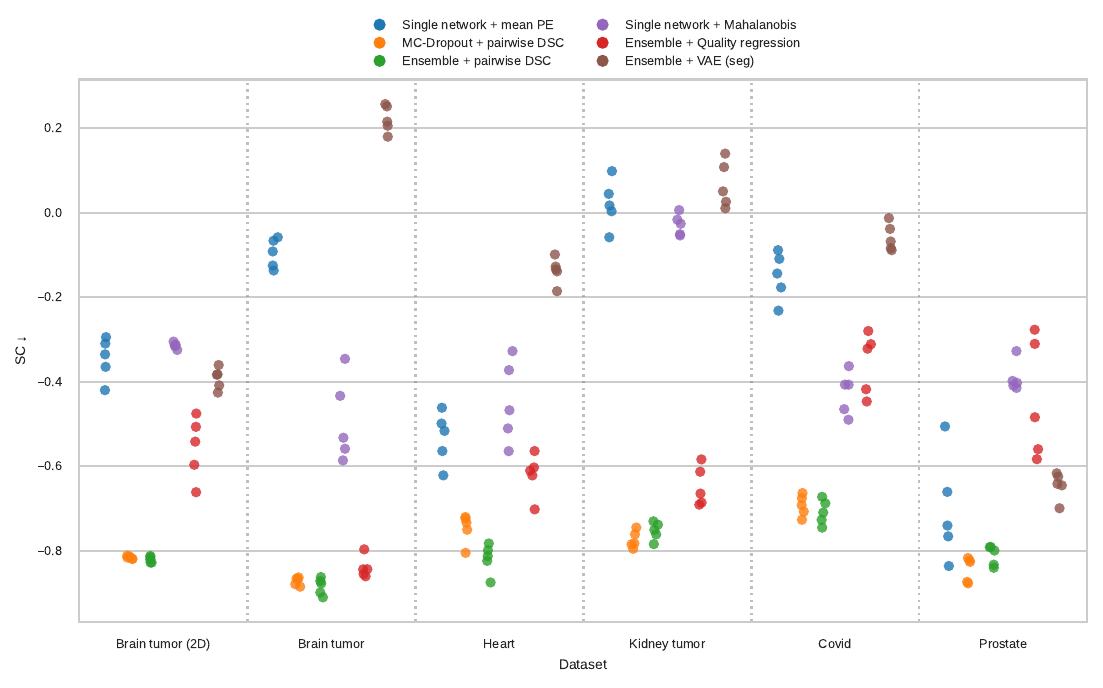}
    \caption{
        Spearman correlation coefficients (SC) for all datasets and methods (negative correlation is better). Each experiment was repeated using 5 folds.
        Most of the pixel-confidence aggregation methods were excluded for clarity, as they perform worse than pairwise Dice.
    }
    \label{fig:overview_spearman}
\end{figure*}

\begin{figure*}[!t]
    \centering
    \includegraphics[width=\textwidth]{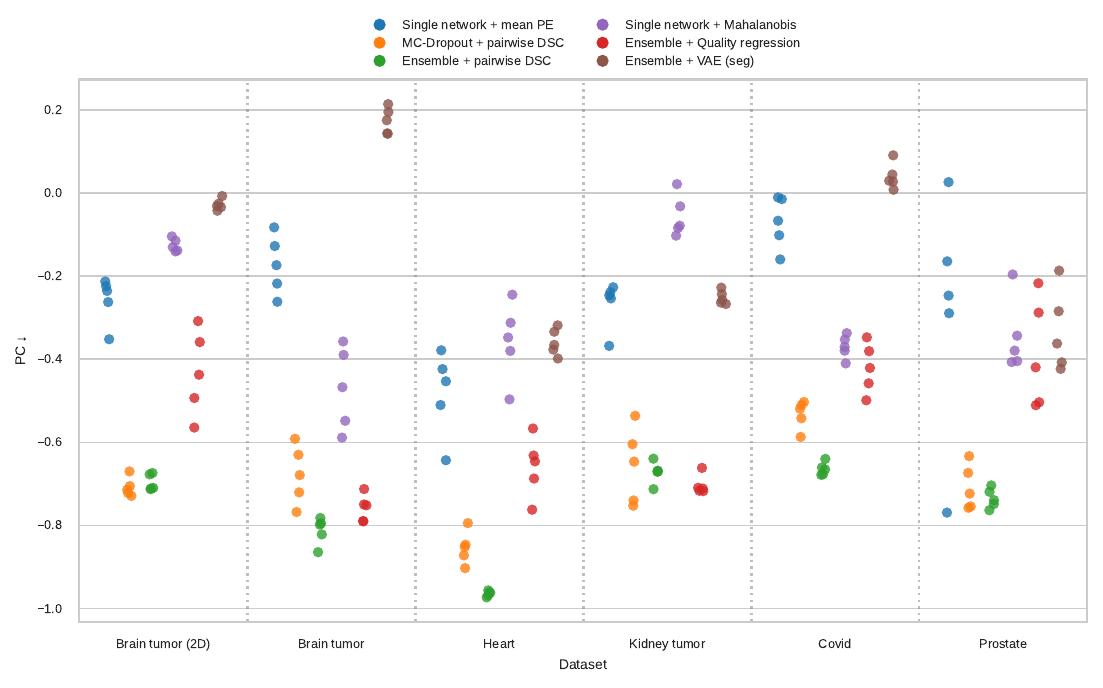}
    \caption{
        Pearson correlation coefficients (PC) for all datasets and methods (negative correlation is better). Each experiment was repeated using 5 folds.
        Most of the pixel-confidence aggregation methods were excluded for clarity, as they perform worse than pairwise Dice.
    }
    \label{fig:pearson_mean_dice_all}
\end{figure*}

\begin{figure*}[!t]
    \centering
    \includegraphics[width=\textwidth]{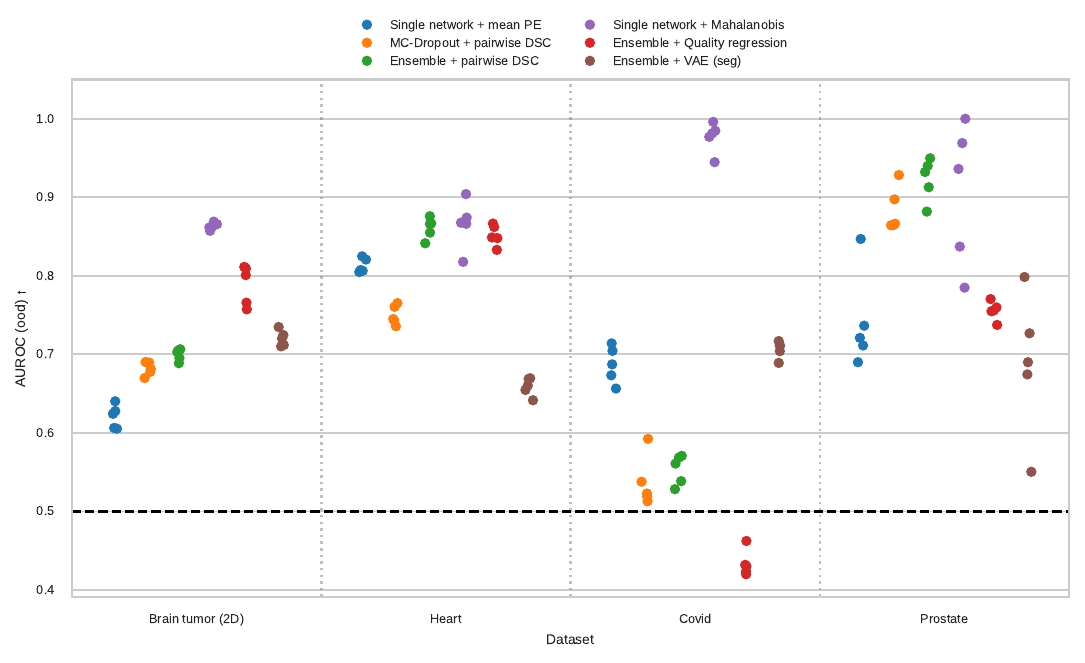}
    \caption{
        OOD-AUROC for all datasets and methods (higher is better). Each experiment was repeated using 5 folds.
        The brain tumor and kidney tumor datasets are not included, as they do not contain OOD samples.
        Most of the pixel-confidence aggregation methods were excluded for clarity, as they perform worse than pairwise Dice.
    }
    \label{fig:ood-auc}
\end{figure*}

\Cref{fig:overview_naurc,fig:overview_spearman,fig:pearson_mean_dice_all} compare the same methods as in \cref{fig:overview_aurc}, but measure performance with the normalized AURC, Spearman and Pearson correlation, respectively.
We include these results to stress that our choice of AURC does not influence the interpretations of the benchmarking experiments significantly, as the method ranking is similar for all metrics.
The nAURC metric is computed from AURC by normalizing it to the range of optimal and random scores (which are different for each experiment): $\mathrm{nAURC} = (\mathrm{AURC} - \mathrm{AURC_{opt}}) / \mathrm{AURC_{rand}} - \mathrm{AURC_{opt}})$, so 0 is the new optimal and 1 corresponds to the AURC of a random confidence score.
In contrast, the OOD-AUROC scores in \cref{fig:ood-auc} show superior performance of the Mahalanobis method, because the OOD detection task evaluated with this metric is very different from failure detection.

\begin{figure*}[!t]
\centering
     \begin{subfigure}[b]{0.48\textwidth}
         \centering
         \includegraphics[width=\textwidth]{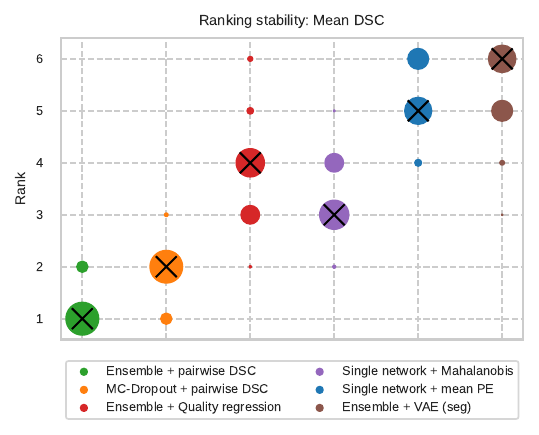}
     \end{subfigure}
     \hfill
     \begin{subfigure}[b]{0.48\textwidth}
         \centering
         \includegraphics[width=\textwidth]{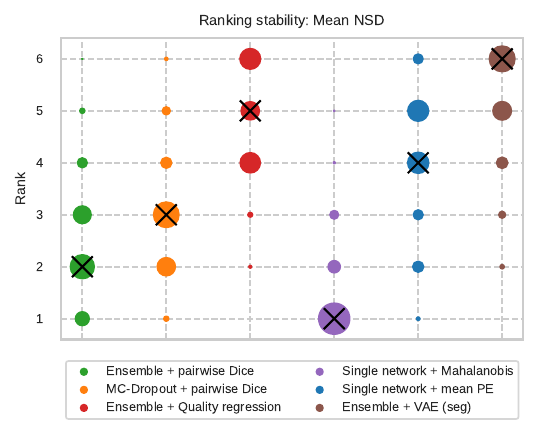}
    \end{subfigure}
    \caption{Impact of the choice of segmentation metric as a risk function on the ranking stability for the Covid dataset. Left: Generalized dice. Right: Mean NSD. Bootstrapping ($N=500$) was used to obtain a distribution of ranks for the results of each fold and the ranking distributions of all folds were combined. All ranks across datasets are combined in this figure, where the circle area is proportional to the rank count and the black x-markers indicate median ranks, which were also used to sort the methods. Compared to \cref{fig:compare_surface_dice}, the Mahalanobis method achieves much better ranking when using NSD as risk.
    }
    \label{fig:compare_surface_dice_520}
\end{figure*}

Finally, in \cref{fig:compare_surface_dice_520}, we show a ranking stability plot for a single dataset (Covid), similar to the combined results from \cref{fig:compare_surface_dice}.
The main observation from this figure is that the ranking stability on individual datasets is higher than what the combined results suggest. This particular dataset is a special case, as the Mahalanobis method can attain the first rank when measuring risks with the NSD metric.

\section{Hyperparameters}
\label{sec:appendix_hparams}

An overview of important hyperparameters is given in \cref{tab:hparams}. Below we describe additional details for each method not covered in the main part.

\textbf{Pixel confidence methods:} For two datasets (brain tumor and kidney tumor) the predicted labels are non-exclusive hierarchical regions. Therefore, we apply a sigmoid nonlinearity at the final layer instead of softmax. To convert this prediction into a confidence map, we get a confidence map for each region first by computing the pixel-wise entropy. As the confidence aggregation methods we consider require a single-channel confidence map, we take the minimum confidence score for each pixel to aggregate region-wise confidence maps.

\textbf{Mahalanobis:} Patch-wise training and inference was performed following \citet{gonzalez_distance-based_2022}. During training, features were extracted from each patch, and a multivariate Gaussian fit with scikit-learn \citep{pedregosa_scikit-learn_2011}. During testing, the Mahalanobis distance (uncertainty) was computed on each patch, up-sampled by repetition to the patch size and aggregated to an uncertainty map using the overlapping-patch-aggregation from the segmentation model. The image-level confidence score is then the mean confidence over the whole image.

\textbf{Regression forest (RF) with simple features:} We used the default parameters for the regression implementation by scikit-learn \citep{pedregosa_scikit-learn_2011}. Features were standardized before model fitting or prediction. Regression targets were the DSC score for each class and the generalized DSC score \citep{crum_generalized_2006}. When evaluating with mean DSC, we computed the confidence score as the mean over the estimated class-wise DSC scores.

\textbf{Regression forest (RF) with radiomics features:} The regression model setup was identical to the RF with simple features. Before training, the confidence threshold for ROI definition was determined as in \citet{jungo_analyzing_2020}: 100 thresholds linearly spaced between [0.05, 0.95] in the normalized confidence score range of the validation set were used to compute the overlap between the resulting uncertain pixels and the factual errors. The threshold with the highest overlap was used for training and evaluation. If some radiomics features were NaN-valued during feature extraction, we replaced them with their mean from the training set.

\textbf{Quality regression:} We used the same regression targets as for the regression forests, i.e. DSC values for each class and generalized DSC. Estimates for mean DSC were obtained by averaging the class-wise DSC predictions. The probability of applying affine misalignment augmentations to the segmentation masks during training was 0.33.

\textbf{VAE:} We use a symmetric encoder-generator architecture that contains convolutions with kernel size 3 and channel sizes [32, 64, 128, 256, 512] (another layer with 16 channels is added in the front for the Covid and Kidney tumor datasets). A fully connected layer projects the bottleneck dimension to 256. Data pre-processing consists of cropping around the foreground prediction, z-normalization, and clipping to 2.
\begin{table*}[t]
\caption{Overview of hyper-parameters for failure detection methods. BCE: binary cross-entropy.}
\label{tab:hparams}
\centering
\begin{tabular}{llll}
\toprule
Method               & Parameter           & 2d dataset  & 3d datasets (if deviating)        \\ \midrule
U-Net                & loss function       & Dice              & Dice + (B)CE               \\
                     & optimizer           & AdamW             & SGD + momentum (0.99)      \\
                     & learning rate       & 0.001             & 0.01                   \\
                     & learning rate decay & -                 & polynomial (exponent 0.9)  \\
                     & weight decay        & 0.00001           & 0.00003                   \\
                     & batch size          & 32                & 2 (heart: 4)               \\
                     & normalization layer & batch             & instance                   \\ \midrule
RF (simple features) & boundary width      & 4                 &                           \\
                     & connectivity for CC & 2                 & 3                          \\ \midrule
Quality regression   & loss function       & \multicolumn{2}{l}{L2}                         \\
                     & optimizer           & \multicolumn{2}{l}{AdamW}                      \\
                     & learning rate       & \multicolumn{2}{l}{0.0002}                   \\
                     & learning rate decay & \multicolumn{2}{l}{cosine}                     \\
                     & weight decay        & \multicolumn{2}{l}{0.0001}                   \\
                     & batch size          & 32                & 2 (heart: 4)               \\ \midrule
Mahalanobis          & Max. feature dim.   & \multicolumn{2}{l}{0.0001}                   \\ \midrule
VAE                  & loss function       & \multicolumn{2}{l}{BCE + $\beta\, \cdot$ KL-div.} \\
                     & $\beta $                & \multicolumn{2}{l}{0.001}                   \\
                     & optimizer           & \multicolumn{2}{l}{Adam}                       \\
                     & learning rate       & \multicolumn{2}{l}{0.0001}                   \\
                     & learning rate decay & \multicolumn{2}{l}{-}                          \\
                     & batch size          & 32                & 6                          \\
 \bottomrule
\end{tabular}

\end{table*}

\end{document}